\pdfoutput=1
\documentclass[11pt]{article}

\usepackage[final]{acl}

\usepackage{times}
\usepackage{latexsym}
\usepackage[T1]{fontenc}
\usepackage[utf8]{inputenc}
\usepackage{microtype}

\usepackage{booktabs}
\usepackage{amsmath,amssymb}
\usepackage{graphicx}
\usepackage{multirow}
\usepackage{tikz}
\usetikzlibrary{arrows.meta,positioning,fit,calc}

\newcommand{\relNConfigs}{8}
\newcommand{\relConfigList}{\texttt{anchored}, \texttt{anchor\_free}, \texttt{full\_corpus}, \texttt{full\_corpus\_anchor\_free}, \texttt{scale}, \texttt{scale\_anchor\_free}, \texttt{anchor\_free\_natural}, \texttt{anchor\_free\_natural\_fallback}}
\newcommand{\nFactKeys}{5,839}
\newcommand{\nSingle}{900}
\newcommand{\nChanging}{892}
\newcommand{\nVal}{424}
\newcommand{\nTest}{468}
\newcommand{\realCosineCur}{0.464}
\newcommand{\realCosineStale}{0.530}

\newcommand{\realBmStale}{0.592}
\newcommand{\realRerankCur}{0.925}

\newcommand{\realBmRerankCur}{0.936}
\newcommand{\realRwCur}{0.784}
\newcommand{\realRwContam}{0.936}
\newcommand{\realBmContam}{0.991}
\newcommand{\realReconCur}{0.859}

\newcommand{\realOracleCur}{0.979}
\newcommand{\realStoreSlots}{995}

\newcommand{\realReconCurPzero}{0.887}
\newcommand{\realOracleCurPzero}{0.998}

\newcommand{\realReconCurPone}{0.829}
\newcommand{\realOracleCurPone}{0.970}

\newcommand{\gapReconOraclePzero}{11.1}
\newcommand{\gapReconOraclePone}{14.1}
\newcommand{\tunedRerankK}{2}
\newcommand{\tunedTau}{0.60}
\newcommand{\realReconCurFive}{0.929}
\newcommand{\realCosineCurFive}{0.985}
\newcommand{\realRerankJump}{0.46}
\newcommand{\realLexCurPzero}{0.991}
\newcommand{\realLexCurPhalf}{0.972}
\newcommand{\realLexCurPone}{0.964}
\newcommand{\realLexCur}{0.972}
\newcommand{\realLexContam}{0.000}
\newcommand{\realLexSlots}{467}

\newcommand{\gapRerankLexPhalf}{4.7}
\newcommand{\stratNTwo}{423}
\newcommand{\stratPctTwo}{90}
\newcommand{\stratNMore}{45}
\newcommand{\stratRerankMore}{0.533}
\newcommand{\stratReconMore}{0.778}
\newcommand{\stratOracleMore}{0.956}
\newcommand{\stratRerankTwo}{0.967}
\newcommand{\stratCosineTwo}{0.482}
\newcommand{\stratCosineMore}{0.289}

\newcommand{\auditNEvents}{995}
\newcommand{\auditOverwrites}{482}
\newcommand{\auditCross}{73}
\newcommand{\auditCrossPct}{7.3}
\newcommand{\auditDup}{71}
\newcommand{\auditZero}{26}
\newcommand{\qpRerankCur}{0.921}
\newcommand{\qpReconCur}{0.857}

\newcommand{\seedReconLoPhalf}{0.823}
\newcommand{\seedReconHiPhalf}{0.859}
\newcommand{\seedRerankLoPhalf}{0.925}
\newcommand{\seedRerankHiPhalf}{0.940}

\newcommand{\mpnetLexCur}{0.940}
\newcommand{\mpnetRwCur}{0.767}
\newcommand{\mpnetCosineCur}{0.468}
\newcommand{\mpnetRerankCur}{0.893}
\newcommand{\mpnetReconCur}{0.780}
\newcommand{\mpnetOracleCur}{0.944}
\newcommand{\mpnetReconLead}{7.5}
\newcommand{\mpnetReconLeadMp}{1.3}
\newcommand{\mpnetReconContam}{0.160}
\newcommand{\mpnetRerankContam}{0.949}
\newcommand{\ctrlSeeds}{20}

\newcommand{\ctrlReconGapHighCur}{1.00}
\newcommand{\ctrlReconJitterTwoCur}{0.80}
\newcommand{\ctrlOracleJitterTwoCur}{0.99}
\newcommand{\ctrlKeyDeficitJitterTwo}{0.20}
\newcommand{\ctrlKeyDeficitJitterFour}{0.16}
\newcommand{\ctrlKeyDeficitJitterEight}{0.05}
\newcommand{\ctrlRwCleanCur}{0.98}
\newcommand{\ctrlRerankBudgetLowCur}{0.47}
\newcommand{\ctrlTuneSeeds}{10}
\newcommand{\ctrlTunedTauGapLow}{0.45}
\newcommand{\ctrlTunedTauGapHigh}{0.3}
\newcommand{\ctrlFixedTauGapHighCur}{0.60}
\newcommand{\ctrlBestTauClean}{0.3}

\newcommand{\ctrlBestTauJitterFour}{0.2}
\newcommand{\ctrlBestTauJitterFourCur}{0.54}
\newcommand{\ctrlTauDefaultJitterFourCur}{0.24}
\newcommand{\ctrlTunedKJitterZero}{20}

\newcommand{\ctrlOracleJitterEightCur}{0.30}
\newcommand{\ctrlOracleJitterFourCur}{0.70}
\newcommand{\ctrlReconJitterFourCur}{0.54}
\newcommand{\ctrlRerankJitterZeroCur}{0.83}

\newcommand{\aliasCoverage}{89}
\newcommand{\aliasNForms}{1,122}
\newcommand{\aliasAmbigPct}{3.5}
\newcommand{\aliasSubjAmbigPct}{17}
\newcommand{\aliasNEventsOne}{472}
\newcommand{\aliasNEventsTotal}{995}

\newcommand{\aliasRerankCurOne}{0.737}

\newcommand{\aliasReconCurZero}{0.859}
\newcommand{\aliasReconCurOne}{0.513}

\newcommand{\aliasReconDefOne}{31.4}
\newcommand{\aliasLexCurZero}{0.972}
\newcommand{\aliasLexCurOne}{0.169}

\newcommand{\aliasLexDefOne}{65.8}

\newcommand{\aliasFuzzyCurOne}{0.714}

\newcommand{\aliasFuzzyDefOne}{11.3}
\newcommand{\aliasEntityCurZero}{0.972}
\newcommand{\aliasEntityCurOne}{0.776}

\newcommand{\aliasEntityDefOne}{5.1}

\newcommand{\aliasFuzzyReadCurOne}{0.831}

\newcommand{\aliasFuzzyRWCurOne}{0.831}

\newcommand{\aliasEntityRWCurZero}{0.994}
\newcommand{\aliasEntityRWCurOne}{0.912}

\newcommand{\aliasOracleCurOne}{0.827}

\newcommand{\aliasFuzzyMissOne}{0.167}

\newcommand{\aliasFuzzyContamOne}{0.032}
\newcommand{\aliasFuzzySlotsOne}{467}

\newcommand{\aliasFuzzyReadMissOne}{0.045}

\newcommand{\aliasFuzzyReadContamOne}{0.983}
\newcommand{\aliasFuzzyReadSlotsOne}{995}

\newcommand{\aliasFuzzyRWStaleOne}{0.122}
\newcommand{\aliasFuzzyRWContamOne}{0.034}
\newcommand{\aliasFuzzyRWSlotsOne}{467}

\newcommand{\aliasEntityContamOne}{0.024}

\newcommand{\aliasOracleMissOne}{0.173}

\newcommand{\symGainFuzzy}{11.8}
\newcommand{\symGainEntity}{13.7}
\newcommand{\resolverGainSym}{8.1}

\newcommand{\aliasPEntityRwFuzzyRwOne}{<10^{-3}}
\newcommand{\aliasPRwEntityOne}{=0.03}
\newcommand{\aliasPRwOracleOne}{=0.93}
\newcommand{\aliasPRwFuzzyOne}{<10^{-3}}
\newcommand{\aliasSkipRecentMaxGap}{2.4}
\newcommand{\llmCacheEntries}{30,818}
\newcommand{\llmCacheLocal}{20,055}
\newcommand{\llmCacheHosted}{10,763}
\newcommand{\blindAnchored}{0.996}

\newcommand{\blindAnchorFree}{0.630}
\newcommand{\blindAnchorFreeContam}{0.641}
\newcommand{\blindFull}{0.970}

\newcommand{\aliasRerankContamOne}{0.882}

\newcommand{\auditThreshSplitOne}{55.3}

\newcommand{\auditLexSplitZero}{0.4}
\newcommand{\auditLexSplitOne}{89.3}

\newcommand{\auditFuzzySplitOne}{13.2}

\newcommand{\auditEntitySplitOne}{6.6}
\newcommand{\auditEntityImpureOne}{25}
\newcommand{\auditEntityImpureZero}{2}
\newcommand{\invSmallN}{468}

\newcommand{\invSmallDef}{11.3}
\newcommand{\invBigN}{2,995}

\newcommand{\invBigDef}{28.6}
\newcommand{\invSmallDefRW}{-0.4}
\newcommand{\invBigDefRW}{19.7}

\newcommand{\invRWAdvantageMin}{9.0}
\newcommand{\invRWAdvantageMax}{12.6}
\newcommand{\substringFalseLink}{42}
\newcommand{\mentionFalseLink}{2.5}
\newcommand{\nDistractStatements}{6,083}
\newcommand{\mechKgSlots}{995}

\newcommand{\mechOwSlots}{467}
\newcommand{\mechKgTauCur}{0.821}
\newcommand{\mechOwTauCur}{0.859}
\newcommand{\fullNEvents}{7,078}
\newcommand{\fullNDistract}{6,083}
\newcommand{\fullNInvOpen}{468}
\newcommand{\fullNInvClosed}{2,995}
\newcommand{\fullCosineOpen}{0.462}

\newcommand{\fullRerankOpen}{0.902}

\newcommand{\fullEntityOpen}{0.902}
\newcommand{\fullEntityClosed}{0.947}

\newcommand{\fullFuzzyDefOne}{23.7}
\newcommand{\fullLexInventoryCost}{0.4}
\newcommand{\realCosineContam}{0.974}
\newcommand{\fullFuzzyRWOpenOne}{0.624}
\newcommand{\fullFuzzyRWDefOne}{6.6}
\newcommand{\fullEntityInventoryCost}{4.5}
\newcommand{\aliasHolmFamily}{9}
\newcommand{\aliasPLexOracleZero}{=1.00}
\newcommand{\aliasPLexOracleRawZero}{=0.26}
\newcommand{\aliasPEntityOracleRawZero}{=0.26}

\newcommand{\aliasPEntityLexZero}{=1.00}
\newcommand{\aliasPLexOracleOne}{<10^{-3}}

\newcommand{\mpAliasLexDefZero}{0.4}

\newcommand{\mpAliasLexDefOne}{52.8}

\newcommand{\mpAliasEntityDefZero}{0.4}

\newcommand{\mpAliasEntityDefOne}{4.3}

\newcommand{\sysMatchedStore}{\texttt{chroma}}

\newcommand{\repeatAdjStore}{\texttt{chroma}}
\newcommand{\repeatAdjCurA}{0.913}
\newcommand{\repeatAdjCurB}{0.913}
\newcommand{\repeatAdjCurDelta}{0.0}
\newcommand{\repeatAdjSlotsA}{238}

\newcommand{\repeatAdjEventsIdentical}{identical}

\newcommand{\probedCurNBackends}{2}
\newcommand{\probedCurSpread}{2.7}
\newcommand{\probedCurLo}{0.180}
\newcommand{\probedCurHi}{0.207}

\newcommand{\probedAdjCurZero}{0.913}
\newcommand{\probedAdjContamZero}{0.120}
\newcommand{\probedAdjSlotsZero}{238}

\newcommand{\textAdjCurZero}{0.573}

\newcommand{\textAdjHedgedZero}{0.000}
\newcommand{\textAdjCurKZero}{0.920}
\newcommand{\textAdjContamZero}{0.593}
\newcommand{\textAdjCurDropZero}{34.0}
\newcommand{\textAdjContamRiseZero}{47.3}
\newcommand{\attrAdjCollapsedZero}{81}

\newcommand{\attrAdjUpdatedStaleZero}{84}
\newcommand{\attrAdjNQueriesZero}{150}

\newcommand{\attrAdjIdWrongPctZero}{1}
\newcommand{\attrAdjNControlsZero}{7}
\newcommand{\attrAdjControlMaxGapZero}{2.0}
\newcommand{\attrAdjControlCollapsedZero}{0}

\newcommand{\probedAdjVersion}{\texttt{0.1.118}}

\newcommand{\probedAdjUpdate}{149}
\newcommand{\probedAdjDelete}{31}
\newcommand{\probedAdjAdd}{269}
\newcommand{\probedAdjRefCosine}{0.460}

\newcommand{\probedAdjRefOracle}{0.987}

\newcommand{\probedAdjSlotsOne}{286}

\newcommand{\textAdjCurOne}{0.207}

\newcommand{\textAdjContamOne}{0.747}

\newcommand{\probedAdjUpdateAF}{114}
\newcommand{\probedCurCurZero}{0.180}
\newcommand{\probedCurContamZero}{0.720}
\newcommand{\probedCurSlotsZero}{857}
\newcommand{\probedCurMissZero}{0.533}
\newcommand{\textCurCurZero}{0.480}

\newcommand{\textCurMissZero}{0.000}

\newcommand{\textCurContamZero}{0.947}

\newcommand{\attrCurIdWrongZero}{421}
\newcommand{\attrCurIdScoredZero}{681}
\newcommand{\attrCurIdWrongPctZero}{62}

\newcommand{\probedCurVersion}{\texttt{2.0.11}}

\newcommand{\ladderSymCurZero}{1.000}
\newcommand{\ladderSymContamZero}{0.000}
\newcommand{\ladderSymSlotsZero}{150}

\newcommand{\ladderSymCurOne}{0.920}
\newcommand{\ladderSymContamOne}{0.027}

\newcommand{\sysTestFamily}{62}
\newcommand{\sysTestSurviving}{46}
\newcommand{\sysTestN}{150}

\newcommand{\probedGapPts}{73}
\newcommand{\sysSubsampleMaxDev}{0.8}
\newcommand{\sysIngestHours}{1.3}
\newcommand{\sysIngestHoursTotal}{4.2}
\newcommand{\sysNFacts}{150}
\newcommand{\sysNStatements}{329}

\newcommand{\sysMemZeroAdds}{329}
\newcommand{\sysMemZeroAddEvents}{857}
\newcommand{\sysMemZeroUpdateEvents}{0}

\newcommand{\sysMemZeroCurFive}{0.820}

\newcommand{\miniAdjVersion}{\texttt{0.1.118}}
\newcommand{\miniNonVersion}{\texttt{2.0.11}}
\newcommand{\miniAdjEvents}{ADD\,1, UPDATE\,1}
\newcommand{\miniNonEvents}{ADD\,2}
\newcommand{\miniAdjSlots}{1}
\newcommand{\miniNonSlots}{2}
\newcommand{\miniNStatements}{2}
\newcommand{\miniAdjRankCurrent}{the current one}
\newcommand{\miniNonRankCurrent}{the stale one}
\newcommand{\sysNGatesProbed}{6}
\newcommand{\sysNGatesPassed}{4}

\newcommand{\sysGateWriteCaveat}{A passing write gate shows only that the system recorded its own decisions, not that it made them well: on a pair whose read gate fails, the adjudicator is choosing against the candidates that backend returned, and its event counts differ from the same release on a passing backend accordingly}
\newcommand{\sysGateFailNote}{we report no retrieval metrics from the rest, because a failing pair still produces plausible-looking numbers --- one returns 2 of the 14 memories it holds and ranks none of them the one the embedder scores first; and one records no decision for any write while still persisting some of them --- that describe our configuration rather than the system}

\newcommand{\scaleNTest}{2,904}

\newcommand{\scaleNMulti}{2,436}
\newcommand{\scaleMeanCurrentValues}{1.23}

\newcommand{\scaleCosineCurZero}{0.470}

\newcommand{\scaleRerankCurZero}{0.721}

\newcommand{\scaleSetRecRwZero}{0.940}
\newcommand{\scaleSetRecMultiRwZero}{0.937}

\newcommand{\scaleReconCurZero}{0.777}

\newcommand{\scaleFuzzyRWDefZero}{2.3}

\newcommand{\scaleSetRecOracleZero}{0.890}
\newcommand{\scaleSetRecMultiOracleZero}{0.871}

\newcommand{\scaleNStatementsOne}{15,809}
\newcommand{\scaleCosineCurOne}{0.232}

\newcommand{\scaleLexCurOne}{0.293}

\newcommand{\scaleFuzzyCurOne}{0.512}

\newcommand{\scaleFuzzyRWCurOne}{0.627}
\newcommand{\scaleFuzzyRWDefOne}{6.3}

\newcommand{\scaleAliasDefOne}{11.4}

\newcommand{\scaleRerankCiLo}{0.705}
\newcommand{\scaleRerankCiHi}{0.737}
\newcommand{\scaleReconCiLo}{0.762}
\newcommand{\scaleReconCiHi}{0.792}
\newcommand{\paraNLevels}{3}
\newcommand{\paraLevels}{0, 0.5, 1}

\newcommand{\paraCosineSpread}{4.3}

\newcommand{\paraFuzzyRWSpread}{0.4}

\newcommand{\paraEntitySpread}{5.1}

\newcommand{\paraOrderStable}{identical at every level}
\newcommand{\paraOrderStableAll}{identical except that the symmetric rung overtakes the write-path oracle once any paraphrasing is present}
\newcommand{\readerNModels}{2}
\newcommand{\readerModelList}{\texttt{qwen2.5:14b}, \texttt{glm-4.6}}

\newcommand{\readerTopK}{5}
\newcommand{\readerCosineCorrect}{0.558}

\newcommand{\readerRerankDates}{0.957}

\newcommand{\readerOracleCorrect}{0.985}

\newcommand{\pairCur}{0.989}
\newcommand{\pairNQueries}{468}
\newcommand{\pairAnchorFreeAgree}{467}
\newcommand{\readerSecCosine}{0.611}

\newcommand{\readerOracleRet}{0.979}
\newcommand{\pairReadContam}{0.998}
\newcommand{\pairRwContam}{0.000}
\newcommand{\pairPriRead}{0.327}
\newcommand{\pairPriRw}{0.983}

\newcommand{\pairPriReadDates}{0.983}
\newcommand{\pairPriRwDates}{0.983}

\newcommand{\pairPriNChangedDates}{10}
\newcommand{\pairPriStaleShare}{64.5}
\newcommand{\pairPriHedgeShare}{1.9}
\newcommand{\pairPriDelta}{65.6}
\newcommand{\pairPriP}{<10^{-3}}
\newcommand{\pairSecRead}{0.684}
\newcommand{\pairSecRw}{0.979}

\newcommand{\pairSecReadDates}{0.957}
\newcommand{\pairSecRwDates}{0.972}
\newcommand{\pairSecDeltaDates}{1.5}

\newcommand{\pairSecStaleShare}{19.2}
\newcommand{\pairSecHedgeShare}{11.1}
\newcommand{\pairSecDelta}{29.5}
\newcommand{\pairSecP}{<10^{-3}}
\newcommand{\readerPriName}{\texttt{qwen2.5:14b}}
\newcommand{\readerPriCosine}{0.558}
\newcommand{\readerPriRerank}{0.329}
\newcommand{\readerPriLex}{0.979}
\newcommand{\readerPriRerankDelta}{-22.9}
\newcommand{\readerPriLexDelta}{+42.1}
\newcommand{\readerSecName}{\texttt{glm-4.6}}

\newcommand{\readerSecRerankDelta}{+6.6}
\newcommand{\readerSecLexDelta}{+36.3}
\newcommand{\readerRetrievalDelta}{+46.2}
\newcommand{\readerSameSetAll}{100}
\newcommand{\readerSameSetReordered}{51}
\newcommand{\revOrderCosine}{0.564}
\newcommand{\revOrderCosineShift}{+0.6}
\newcommand{\revOrderCosineP}{=0.89}
\newcommand{\revOrderRerank}{0.703}
\newcommand{\revOrderRerankShift}{+37.4}
\newcommand{\revOrderRerankP}{<10^{-3}}
\newcommand{\revOrderLex}{0.979}
\newcommand{\revOrderLexShift}{+0.0}
\newcommand{\revOrderLexP}{=1.00}
\newcommand{\revOrderOracle}{0.985}
\newcommand{\revOrderOracleShift}{+0.0}
\newcommand{\revOrderOracleP}{=1.00}
\newcommand{\rerankVsBaseFwd}{-22.9}
\newcommand{\rerankVsBaseRev}{+13.9}
\newcommand{\lexVsBaseFwd}{+42.1}
\newcommand{\lexVsBaseRev}{+41.5}
\newcommand{\concordExactP}{=0.25}
\newcommand{\concordNBlocks}{2}
\newcommand{\concordCurAgree}{27}
\newcommand{\concordCurInvert}{13}

\newcommand{\concordCurTotal}{42}
\newcommand{\concordCurInvertStrict}{3}
\newcommand{\concordCurPairsStrict}{12}
\newcommand{\concordCurTotalStrict}{24}
\newcommand{\concordContamAgree}{32}
\newcommand{\concordContamInvert}{2}
\newcommand{\concordContamTie}{8}

\newcommand{\concordContamInvertStrict}{0}

\newcommand{\concordTieOnCurInvert}{3}

\newcommand{\flipCosine}{213}
\newcommand{\flipRerank}{219}
\newcommand{\flipLex}{0}
\newcommand{\condPriDirty}{0.625}
\newcommand{\condPriClean}{0.913}
\newcommand{\condPriDelta}{+28.8}
\newcommand{\condPriNDirty}{64}
\newcommand{\condPriNClean}{404}
\newcommand{\condSecDirty}{0.672}
\newcommand{\condSecClean}{0.903}
\newcommand{\condSecDelta}{+23.2}

\newcommand{\readerCeiling}{0.994}

\newcommand{\readerNFailed}{0}

\newcommand{\readerPriClosedBook}{0.088}
\newcommand{\readerSecClosedBook}{0.244}
\newcommand{\doseCurrentOnly}{0.991}

\newcommand{\doseSecCurrentOnly}{0.987}
\newcommand{\doseCurrentFirst}{0.314}
\newcommand{\doseCurrentFirstDates}{0.925}

\newcommand{\doseSecCurrentFirst}{0.592}
\newcommand{\doseStaleFirst}{0.850}

\newcommand{\doseSecStaleFirst}{0.637}
\newcommand{\dosePlusTwo}{0.262}

\newcommand{\doseStaleOnly}{0.000}

\newcommand{\doseNModels}{2}
\newcommand{\doseModelList}{\texttt{qwen2.5:14b}, \texttt{glm-4.6}}
\newcommand{\doseDropMin}{39.5}
\newcommand{\doseDropMax}{67.7}
\newcommand{\posPriGain}{53.6}

\newcommand{\posPriP}{<10^{-3}}
\newcommand{\posSecGain}{4.5}
\newcommand{\posSecCiLo}{0.0}
\newcommand{\posSecCiHi}{9.2}
\newcommand{\posSecP}{=0.076}
\newcommand{\posCurrentFirst}{0.314}
\newcommand{\posCurrentLast}{0.850}
\newcommand{\posGain}{53.6}
\newcommand{\posGainLo}{48.7}
\newcommand{\posGainHi}{58.3}
\newcommand{\posP}{<10^{-3}}
\newcommand{\dosePlusTwoN}{42}
\newcommand{\doseWorstDrop}{67.7}
\newcommand{\contamDrop}{14.1}
\newcommand{\contamDropLo}{10.9}
\newcommand{\contamDropHi}{17.3}
\newcommand{\contamDropP}{<10^{-3}}
\newcommand{\contamNPaired}{468}
\newcommand{\llmKeyNModels}{1}

\newcommand{\llmKeyPriName}{\texttt{glm-4.6}}
\newcommand{\llmKeyPriCurZero}{0.846}
\newcommand{\llmKeyPriDefZero}{13.2}
\newcommand{\llmKeyPriSplitZero}{27.6}

\newcommand{\llmKeyPriCurOne}{0.459}

\newcommand{\llmKeyPriSplitOne}{52.4}
\newcommand{\llmKeyPriImpureOne}{10}
\newcommand{\llmKeyPriCalls}{994}
\newcommand{\llmKeyNEvents}{995}
\newcommand{\llmKeyNCand}{4}
\newcommand{\llmKeyRecallZero}{0.970}
\newcommand{\llmKeyRecallOne}{0.861}
\newcommand{\corrSeeds}{3}

\newcommand{\corrArrivalCosine}{0.464}

\newcommand{\corrCleanRerank}{0.925}
\newcommand{\corrClockRerank}{0.484}
\newcommand{\corrArrivalRerank}{0.925}

\newcommand{\corrCleanRw}{0.784}
\newcommand{\corrClockRw}{0.451}

\newcommand{\corrClockRecon}{0.859}

\newcommand{\corrCleanLex}{0.972}
\newcommand{\corrClockLex}{0.972}
\newcommand{\corrArrivalLex}{0.472}

\newcommand{\corrArrivalTkg}{0.472}

\newcommand{\corrClockOracle}{0.979}
\newcommand{\corrArrivalOracle}{0.472}

\newcommand{\corrClockRead}{0.491}
\newcommand{\corrArrivalRead}{0.989}

\newcommand{\corrClockLexNewer}{0.488}
\newcommand{\corrArrivalLexNewer}{0.972}

\newcommand{\corrSeedSpreadMax}{5.8}
\newcommand{\corrSlotsRead}{995}
\newcommand{\corrSlotsSym}{464}

\newcommand{\lateRerankFifty}{0.925}
\newcommand{\lateLexTen}{0.875}
\newcommand{\lateLexTwentyFive}{0.729}
\newcommand{\lateLexFifty}{0.486}
\newcommand{\lateOracleTen}{0.880}
\newcommand{\lateOracleTwentyFive}{0.734}
\newcommand{\lateOracleFifty}{0.491}

\newcommand{\lateLexNewerFifty}{0.972}
\newcommand{\lateLexContamMax}{0.000}
\newcommand{\lateNEligible}{468}

\newcommand{\doseSecCurrentFirstDates}{0.938}

\newcommand{\natNatFbReconDefOne}{22.0}

\newcommand{\natNatFbLexDefOne}{56.4}

\newcommand{\natNatFuzzyRWCurOne}{0.780}

\newcommand{\natNatFbFuzzyRWCurOne}{0.679}

\newcommand{\natNatEntityCurOne}{0.812}
\newcommand{\natNatEntityDefOne}{0.9}
\newcommand{\natNatFbEntityCurOne}{0.701}

\newcommand{\natNatAliasRWCurOne}{0.985}

\newcommand{\natNatFbAliasRWCurOne}{0.959}

\newcommand{\natNatOracleCurOne}{0.821}

\newcommand{\natNatFbOracleCurOne}{0.726}

\newcommand{\natCovSubjects}{270}
\newcommand{\natNSubjects}{467}
\newcommand{\natNatAliasRWAboveOne}{16.5}

\newcommand{\natCovSubjectsPct}{58}
\newcommand{\natUncoveredPct}{42}
\newcommand{\natCovStatementsPct}{59.6}
\newcommand{\natFbCovStatementsPct}{95.1}
\newcommand{\natValCovSubjects}{288}
\newcommand{\natNForms}{1,176}
\newcommand{\natOverlapPct}{6.0}
\newcommand{\natNAmbiguousForms}{0}
\newcommand{\aliasNAmbiguousForms}{55}
\newcommand{\natFbUnlinked}{88}
\newcommand{\natPRwAliasNat}{=0.08}
\newcommand{\natPRwFuzzyNat}{<10^{-3}}
\newcommand{\natPAliasRwAliasNat}{<10^{-3}}
\newcommand{\natPRwAliasNatFb}{=0.40}
\newcommand{\natPRwFuzzyNatFb}{<10^{-3}}

\newcommand{\gphLib}{\texttt{0.29.2}}

\newcommand{\gphNQueries}{150}
\newcommand{\gphCurZero}{0.347}
\newcommand{\gphCurTextZero}{0.347}
\newcommand{\gphContamZero}{0.940}

\newcommand{\gphSlotsZero}{321}
\newcommand{\gphActiveZero}{164}

\newcommand{\gphInvalidationsZero}{157}
\newcommand{\gphNoEdgeZero}{9}
\newcommand{\gphNodesZero}{445}
\newcommand{\gphIngestHoursZero}{3.6}
\newcommand{\gphCallsPerStmtZero}{3.9}
\newcommand{\gphInvalidFirstZero}{96}
\newcommand{\gphVFCurZero}{0.833}

\newcommand{\gphVFContamZero}{0.100}

\newcommand{\gphPVFCurZero}{=0.001}
\newcommand{\gphNSupersessionsZero}{179}
\newcommand{\gphRdAsIsZero}{0.593}
\newcommand{\gphRdVFZero}{0.893}

\newcommand{\gphRdMergedZero}{0.967}
\newcommand{\gphRdOldestZero}{0.033}
\newcommand{\gphCurOne}{0.180}

\newcommand{\gphInvalidationsOne}{137}
\newcommand{\gphNoEdgeOne}{24}
\newcommand{\gphNodesOne}{538}

\newcommand{\gphVFCurOne}{0.640}

\newcommand{\gphNSupersessionsOne}{179}
\newcommand{\gphRdAsIsOne}{0.433}
\newcommand{\gphRdVFOne}{0.773}

\newcommand{\gphIngestHoursTotal}{7.5}

\newcommand{\adjDefaultQwenSilentZero}{3}

\newcommand{\adjDefaultGlmCurZero}{0.067}

\newcommand{\adjDefaultGlmSlotsZero}{20}

\newcommand{\adjDefaultGlmUpdatesZero}{2}

\newcommand{\adjDefaultGlmSilentZero}{307}

\newcommand{\adjTpQwenCurZero}{0.960}
\newcommand{\adjTpQwenCurTextZero}{0.587}

\newcommand{\adjTpQwenContamTextZero}{0.653}

\newcommand{\adjTpQwenUpdatesZero}{155}

\newcommand{\adjTpQwenRdAsIsZero}{0.673}
\newcommand{\adjTpQwenRdMergedZero}{0.927}

\newcommand{\adjTpGlmCurZero}{0.793}
\newcommand{\adjTpGlmCurTextZero}{0.793}
\newcommand{\adjTpGlmContamZero}{0.007}
\newcommand{\adjTpGlmContamTextZero}{0.027}
\newcommand{\adjTpGlmSlotsZero}{138}

\newcommand{\adjTpGlmUpdatesZero}{142}
\newcommand{\adjTpGlmDeletesZero}{19}

\newcommand{\adjTpGlmCurOne}{0.627}
\newcommand{\adjTpGlmCurTextOne}{0.627}

\newcommand{\adjTpGlmRdAsIsZero}{0.807}
\newcommand{\adjTpGlmRdMergedZero}{0.807}

\newcommand{\gphGlmCurZero}{0.347}

\newcommand{\gphGlmSlotsZero}{320}
\newcommand{\gphGlmActiveZero}{158}
\newcommand{\gphGlmInvalidationsZero}{162}

\newcommand{\gphGlmHoursZero}{5.7}
\newcommand{\gphGlmCallsPerStmtZero}{4.0}
\newcommand{\gphGlmVFCurZero}{0.860}

\newcommand{\lgmQwenCurZero}{0.773}
\newcommand{\lgmQwenCurTextZero}{0.773}
\newcommand{\lgmQwenContamZero}{0.140}
\newcommand{\lgmQwenContamTextZero}{0.173}
\newcommand{\lgmQwenMissZero}{0.147}

\newcommand{\lgmQwenSlotsZero}{327}
\newcommand{\lgmQwenInsertsZero}{327}
\newcommand{\lgmQwenUpdatesZero}{375}

\newcommand{\lgmQwenNoopZero}{16}

\newcommand{\lgmQwenHoursZero}{3.1}
\newcommand{\lgmQwenCallsPerStmtZero}{1.3}
\newcommand{\lgmQwenLib}{\texttt{0.0.30}}
\newcommand{\lgmQwenRdAsIsZero}{0.833}

\newcommand{\lgmQwenRdMergedZero}{0.920}

\newcommand{\lgmQwenCurOne}{0.340}
\newcommand{\lgmQwenCurTextOne}{0.413}

\newcommand{\lgmQwenMissOne}{0.553}

\newcommand{\lgmQwenRdAsIsOne}{0.600}

\newcommand{\lgmQwenRdMergedOne}{0.700}

\newcommand{\lgmQwenTextLenZero}{59}

\newcommand{\lgmQwenValuesNamedZero}{0.89}

\newcommand{\lgmQwenPatchFail}{61}

\newcommand{\lgmGlmCurZero}{0.800}
\newcommand{\lgmGlmCurTextZero}{0.147}

\newcommand{\lgmGlmHedgedZero}{0.720}
\newcommand{\lgmGlmSlotsZero}{209}

\newcommand{\lgmGlmRdAsIsZero}{0.613}

\newcommand{\lgmGlmRdMergedZero}{0.627}

\newcommand{\lgmGlmRdStaleZero}{0.033}
\newcommand{\lgmGlmRdOtherZero}{0.367}

\newcommand{\lgmGlmTextLenZero}{332}
\newcommand{\lgmGlmPostWindowZero}{0.253}
\newcommand{\lgmGlmValuesNamedZero}{1.72}

\newcommand{\ersQwenCurZero}{0.853}
\newcommand{\ersQwenCurTextZero}{0.860}
\newcommand{\ersQwenContamZero}{0.020}
\newcommand{\ersQwenContamTextZero}{0.100}
\newcommand{\ersQwenMissZero}{0.147}

\newcommand{\ersQwenHedgedZero}{0.027}
\newcommand{\ersQwenSlotsZero}{182}

\newcommand{\ersQwenKeepsZero}{1395}
\newcommand{\ersQwenRewritesZero}{88}
\newcommand{\ersQwenDeletesZero}{147}
\newcommand{\ersQwenParseFailZero}{0}
\newcommand{\ersQwenHoursZero}{0.5}
\newcommand{\ersQwenRetrieveK}{5}

\newcommand{\ersQwenCallsPerStmtZero}{1.0}
\newcommand{\ersQwenRdAsIsZero}{0.867}

\newcommand{\ersQwenRdMergedZero}{0.873}

\newcommand{\ersQwenRdStaleZero}{0.013}

\newcommand{\ersQwenCurOne}{0.733}
\newcommand{\ersQwenCurTextOne}{0.707}

\newcommand{\ersQwenMissOne}{0.173}

\newcommand{\ersQwenSlotsOne}{194}

\newcommand{\ersQwenRdAsIsOne}{0.807}

\newcommand{\ersQwenGoneZero}{0.133}

\newcommand{\ersGlmCurZero}{0.953}
\newcommand{\ersGlmCurTextZero}{0.953}

\newcommand{\ersGlmContamTextZero}{0.060}

\newcommand{\ersGlmSlotsZero}{194}

\newcommand{\ersGlmRewritesZero}{56}
\newcommand{\ersGlmDeletesZero}{135}

\newcommand{\ersGlmGapToOracleZero}{3.3}
\newcommand{\ersGlmGapQueriesZero}{5}

\newcommand{\ersGlmRdAsIsZero}{0.980}

\newcommand{\ersGlmCurOne}{0.713}

\newcommand{\ersGlmStaleOne}{0.220}

\newcommand{\ersGlmSlotsOne}{250}

\newcommand{\ersGlmDeletesOne}{79}

\newcommand{\ersGlmRdAsIsOne}{0.813}

\newcommand{\ersGlmGoneZero}{0.013}

\newcommand{\sysRecencyValid}{150}
\newcommand{\sysRecencyTotal}{150}
\newcommand{\sysrdAdjReader}{\texttt{qwen2.5:14b}}

\newcommand{\sysrdCosineRagKeyedZero}{0.293}

\newcommand{\sysrdCosineRagKeyedRetrZero}{1.000}

\newcommand{\sysrdCosineRagMergedZero}{1.000}

\newcommand{\sysrdCosineRagOldestZero}{0.007}

\newcommand{\sysrdTimeRerankAsIsZero}{0.300}

\newcommand{\sysrdTimeRerankAsIsRetrZero}{0.920}

\newcommand{\sysrdAdjSysAsIsZero}{0.720}

\newcommand{\sysrdAdjSysKeyedZero}{0.573}

\newcommand{\sysrdAdjSysKeyedRetrZero}{0.973}

\newcommand{\sysrdAdjSysMergedZero}{0.900}

\newcommand{\sysrdAdjSysMergedRetrZero}{0.973}

\newcommand{\sysrdAdjSysOldestZero}{0.353}

\newcommand{\sysrdAdjSysGainZero}{18.0}

\newcommand{\sysrdAdjSysAsIsOne}{0.500}

\newcommand{\sysrdAdjSysMergedOne}{0.853}

\newcommand{\sysrdAdjSysOldestOne}{0.140}

\newcommand{\sysrdAdjSysGainOne}{35.3}
\newcommand{\sysrdAdjFamily}{220}
\newcommand{\sysrdAdjSurviving}{92}
\newcommand{\sysrdPAdjKeyedZero}{$0.039$}
\newcommand{\sysrdPAdjMergedZero}{$0.022$}

\newcommand{\sysrdCurSysAsIsZero}{0.627}

\newcommand{\sysrdCurSysMergedZero}{0.993}

\newcommand{\sysrdCurSysOldestZero}{0.007}

\newcommand{\sysrdCurSysGainZero}{36.7}
\newcommand{\sysrdCurSysAsIsOne}{0.580}

\hypersetup{pdftitle={StaleBench: Isolating Stale Retrieval in Agent Memory, and Pricing the Fix},
  pdfauthor={S M Mahbub Murshed}}
\newcommand{\releaseurl}{\url{https://github.com/thenexuslab/stalebench-release}}
\newcommand{\releasedoi}{\url{https://doi.org/10.5281/zenodo.24237231}}

\title{StaleBench: Isolating Stale Retrieval in Agent Memory,
and Pricing the Fix}

\author{S M Mahbub Murshed \\
  Causal Dynamics Lab \\
  \texttt{murshed@gmail.com}}

\begin{document}

\maketitle

\begin{abstract}
An agent that stores facts over time eventually stores several versions of
the same fact. Asked a \emph{value-agnostic} question --- ``which team does X
play for \emph{now}?'' --- a similarity retriever cannot prefer the current
version, because every version matches the entity the query names.
\textbf{StaleBench} isolates this \emph{stale-retrieval} failure on
\nChanging{} real updating facts, with currency labels from the source
(Wikidata validity years via TempLAMA), never from a retrieval score. Our
central result is that retrieval metrics can rank memory policies backwards:
a retention policy and its retain-everything control return the
\emph{identical} statement at rank 1 for every test query, yet differ by
\pairSecDelta{} and \pairPriDelta{} points in reader accuracy on two readers
when no date is shown. Contamination@5, the metric we add, ranks them
correctly where current@1 and current@5 do not. Withdrawing the lexical
anchor that statements carry costs deployed keying mechanisms tens of points
of freshness; entity resolution on both the write path and the query recovers
most of the loss. Three deployed systems (Mem0, Graphiti, LangMem) and a
reimplemented ERASE each handle supersession differently, and no single
scoring convention describes all four. We release the data, a frozen evaluator, and
reference baselines.
\end{abstract}

\section{Introduction}
\label{sec:intro}

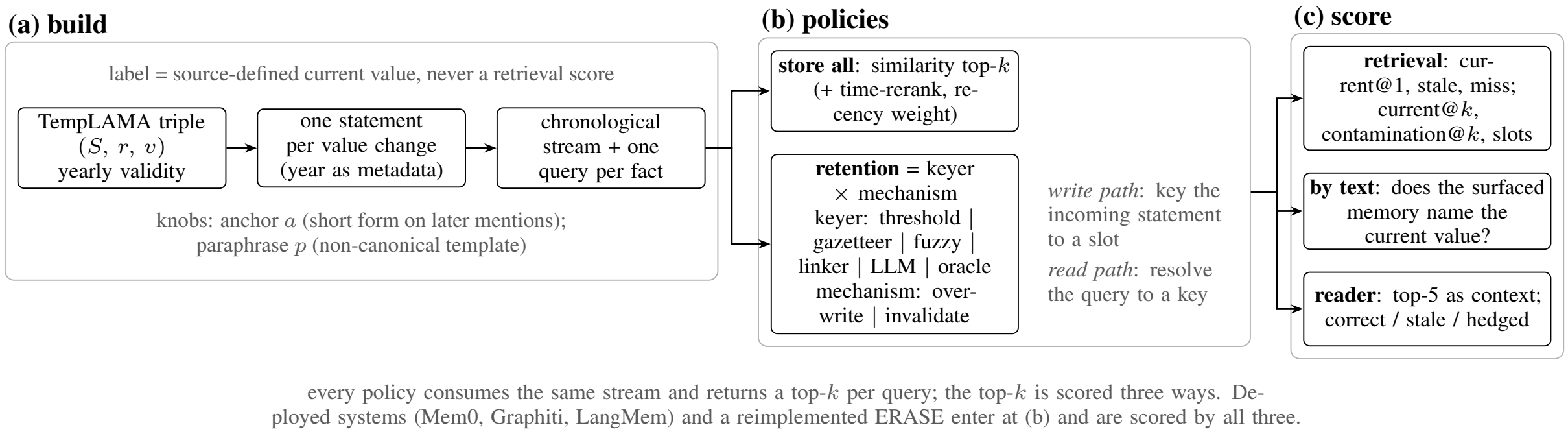
\begin{figure*}[t]
\centering
\resizebox{\textwidth}{!}{\begin{tikzpicture}[
  font=\scriptsize,
  box/.style={draw, rounded corners=2pt, align=center, inner sep=2pt,
              minimum height=2.2em, text width=#1},
  lab/.style={font=\scriptsize, text=black!65, align=center},
  ttl/.style={font=\small\bfseries, anchor=south west, inner sep=1pt},
  arr/.style={-{Stealth[length=4pt]}, semithick},
  pan/.style={draw=black!30, rounded corners=3pt, inner sep=4pt}]

\node[box=2.25cm] (triple) at (0,0) {TempLAMA triple\\$(S,\,r,\,v)$\\yearly validity};
\node[box=2.25cm, right=3.5mm of triple] (runs) {one statement\\per value change\\(year as metadata)};
\node[box=2.25cm, right=3.5mm of runs] (stream) {chronological\\stream + one\\query per fact};
\draw[arr] (triple) -- (runs);
\draw[arr] (runs) -- (stream);
\node[lab, below=1.5mm of runs] (knobs) {knobs: anchor $a$ (short form on later mentions);\\paraphrase $p$ (non-canonical template)};
\node[lab, above=1.5mm of runs] (label) {label = source-defined current value, never a retrieval score};
\node[pan, fit=(triple)(runs)(stream)(knobs)(label)] (pa) {};
\node[ttl] at (pa.north west) {(a) build};

\node[box=2.7cm, anchor=north west] (sa) at ($(pa.north east)+(6mm,-1mm)$)
  {\textbf{store all}: similarity top-$k$\\(+ time-rerank, recency weight)};
\node[box=2.7cm, below=2.5mm of sa] (ret)
  {\textbf{retention} = keyer $\times$ mechanism\\keyer: threshold $|$ gazetteer $|$ fuzzy $|$\\linker $|$ LLM $|$ oracle\\mechanism: overwrite $|$ invalidate};
\node[lab, right=2mm of ret.east, text width=2.05cm, align=left, anchor=west] (paths)
  {\emph{write path}: key the incoming statement to a slot\\[2pt]
   \emph{read path}: resolve the query to a key};
\node[pan, fit=(sa)(ret)(paths)] (pb) {};
\node[ttl] at (pb.north west) {(b) policies};
\draw[arr] (stream.east) -- ++(3mm,0) |- (sa.west);
\draw[arr] (stream.east) -- ++(3mm,0) |- (ret.west);

\node[box=2.7cm, anchor=north west] (retr) at ($(pb.north east)+(6mm,-1mm)$)
  {\textbf{retrieval}: current@1, stale, miss;\\current@$k$, contamination@$k$, slots};
\node[box=2.7cm, below=2.5mm of retr] (text)
  {\textbf{by text}: does the surfaced\\memory name the current value?};
\node[box=2.7cm, below=2.5mm of text] (reader)
  {\textbf{reader}: top-5 as context;\\correct / stale / hedged};
\node[pan, fit=(retr)(text)(reader)] (pc) {};
\node[ttl] at (pc.north west) {(c) score};
\draw[arr] (pb.east) -- ++(3mm,0) |- (retr.west);
\draw[arr] (pb.east) -- ++(3mm,0) |- (text.west);
\draw[arr] (pb.east) -- ++(3mm,0) |- (reader.west);

\node[lab, below=1.5mm of current bounding box.south, text width=15cm]
  {every policy consumes the same stream and returns a top-$k$ per query; the top-$k$ is scored three ways. Deployed systems (Mem0, Graphiti, LangMem) and a reimplemented ERASE enter at (b) and are scored by all three.};
\end{tikzpicture}}
\caption{StaleBench in one picture. (a) Real updating facts become a
chronological stream with one value-agnostic query per fact and a
source-defined current label; two knobs set how hard same-fact detection is.
(b) A policy is store-all retrieval, optionally time-aware, or retention $=$ a
\emph{keyer} (is this a fact already stored?) $\times$ a \emph{mechanism}
(overwrite, or invalidate and keep), run on the write path, the read path, or
both. (c) Each top-$k$ is scored by retrieval metrics, by what its text says,
and by what a reader answers.}
\label{fig:overview}
\end{figure*}

An agent that remembers what it is told will eventually remember several
versions of one fact. Suppose it stored ``Peter Sagan plays for
Bora-Hansgrohe'' and, later, ``Peter Sagan plays for TotalEnergies'', and is
then asked which team Sagan plays for \emph{now}. Both statements name the
entity the question names, so under any similarity function they match about
equally well and the retriever returns whichever one similarity noise
favours: across the real facts of \S\ref{sec:real} that is the stale version
\realCosineStale{} of the time. The cheap fix --- rerank the top hits by
timestamp --- returns the TotalEnergies statement first with the Bora
statement beside it, and a reader handed both does \emph{worse} than one
handed the unfixed retrieval, for one of our two readers
(\S\ref{sec:reader}). A retention policy that recognises the second statement
as a new version of the first and overwrites it returns TotalEnergies alone
--- but only if it can tell that the two statements describe one fact, which
is what most of this paper prices. Figure~\ref{fig:overview} lays out the
benchmark that measures each step.

Existing evaluations do not isolate this failure: freshness benchmarks ask whether a \emph{model} is current with the world, knowledge-conflict studies measure how a reader behaves once conflicting context has been retrieved, long-term-memory benchmarks measure broad recall, where multi-session gold dilutes staleness \citep{maharana2024locomo,wu2025longmemeval}, and production memory systems ship freshness mechanisms without a public instrument that says when the mechanism is needed (\S\ref{sec:related}). Appendix~\ref{sec:systems} shows why citing such a mechanism is not the same as knowing it runs.

StaleBench is that instrument. It contributes:

\textbf{(1) An isolating task with source-defined labels.} Facts stream into a
store in time order, queries are value-agnostic, and the label ``current''
comes from the source (Wikidata validity years via TempLAMA
\citep{dhingra2022templama}, or a generator schedule), never from a retrieval
score; metrics separate \emph{stale} answers from \emph{misses} and add
\emph{contamination@k} and footprint (\S\ref{sec:bench}).

\textbf{(2) A measured price for same-fact detection.} Retention removes
staleness at the source only if it can tell that two statements describe one
fact. We make that decision's difficulty a controlled variable: after the
introducing statement, later mentions may use ordinary short forms
(\emph{Ebrard}, \emph{M. Ebrard}, \emph{Los Angeles}) instead of the full
name, and along that axis \S\ref{sec:anchor} prices a ladder of keyers, from
cosine threshold to a true-key oracle, under two mechanisms.

\textbf{(3) A fix that is free.} Those rungs, and the systems they abstract,
resolve entities when \emph{writing} and answer queries by similarity. Running
the same resolver on the query needs nothing the write path lacked, and the
two paths fix different failures: the read path removes misses, the write
path removes contamination and halves the footprint (\S\ref{sec:anchor}).

\textbf{(4) Evidence that the retrieval metric ranks these policies wrongly.}
We attach readers to the pipeline: current@1 inverts \concordCurInvert{} of
\concordCurTotal{} policy comparisons against reader accuracy,
contamination@5 inverts \concordContamInvert{} (census in
Appendix~\ref{app:readerx}), and a controlled pair --- two keyers that
return the same statement at rank 1 on every one of the \pairNQueries{}
queries and differ only in what fills the rest of the context --- is
\pairPriDelta{} points apart at the reader without surfaced dates
(\S\ref{sec:reader}). Constructed
contexts isolate the mechanism, and a closed-book control bounds what a reader
knows without the store (\readerPriClosedBook{} and \readerSecClosedBook{}).

\textbf{(5) Three deployed memory systems and one reimplemented mechanism
placed on this surface, and the finding that placing one by source
attribution does not work.} Mem0, scored under the same harness, streams and
encoder as the idealized rungs, is reported three ways --- by source
attribution, by what the memories say, by what a reader answers --- and
attribution disagrees with the text only on the deployed system; Graphiti's invalidation fires on the write path and
is ignored by its default read path; LangMem's update fires and keeps what
its model decides; ERASE's edit reaches the stale entry
and, under the smaller model, the neighbouring fact too (\S\ref{sec:sysmain}).

\section{Related work}
\label{sec:related}

Table~\ref{tab:position} (Appendix~\ref{app:position}) positions the benchmark. Four distinctions carry the
novelty. \textbf{Model vs.\ store freshness:} TempLAMA
\citep{dhingra2022templama} probes what a parametric LM believes at a time;
FreshQA \citep{vu2023freshllms}, RealTime~QA \citep{kasai2023realtimeqa},
StreamingQA \citep{liska2022streamingqa}, DyKnow \citep{mousavi2024dyknow},
and GrowOVER \citep{ko2024growover} test whether systems track a changing
world or corpus. StaleBench reuses TempLAMA's temporal triples but asks a
different question one layer down: given that the agent \emph{already stored}
both the old and new value, does its retrieval surface the current one? Temporal QA and temporal-editing datasets --- TimeQA \citep{chen2021timeqa}, SituatedQA \citep{zhang2021situatedqa}, TemporalWiki \citep{jang2022temporalwiki}, MQuAKE \citep{zhong2023mquake} --- likewise ask whether a \emph{model's} answers track time-scoped, evolving, or edited facts; none models an agent-owned store deciding what to retain when both values have already been written to it.
\textbf{Reader vs.\ retrieval:} inter-context conflict studies
\citep{chen2022rich,jin2024tugofwar} and prior-vs-context work
\citep{wu2024clasheval} establish that readers resolve conflicting evidence
unreliably. We take the complementary side --- which versions reach the
context --- and then close the loop (\S\ref{sec:reader}), where the two
literatures turn out to interact: a retrieval fix that looks strong in
isolation can be harmful once a reader consumes its output.
\textbf{Keying vs.\ linking:} the same-fact decision is an entity-resolution
problem with a long literature --- record linkage since
\citet{fellegi1969theory}, neural entity matching \citep{li2020ditto}, entity
linking \citep{wu2020blink}, coreference \citep{joshi2019coref}, and, closest
to our anchor-free axis, temporal entity linking over Wikipedia snapshots
\citep{zaporojets2022tempel}. That work measures matching quality
intrinsically; what is missing is its price in the decision it feeds --- how
many points of freshness a given quality of same-fact detection costs, and how
that moves as the anchor is withdrawn --- which is why our keying ladder is a
cost curve against an oracle rather than another matcher. \textbf{Mechanism vs.\ instrument:} systems ship, and papers propose, freshness machinery ---
MemGPT's self-editing \citep{packer2023memgpt}, Mem0's ADD/UPDATE/DELETE
\citep{chhikara2025mem0}, Zep's bi-temporal invalidation
\citep{rasmussen2025zep}, Memory-R1's RL-learned operations
\citep{yan2025memoryr1}, ERASE's rewrite-on-arrival \citep{li2024erase},
importance-and-recency scoring \citep{park2023generative}, graph-based indexing
\citep{gutierrez2024hipporag} --- but no public benchmark isolates the failure
these mechanisms exist to prevent or prices their hardest component. The
policy ladder evaluates these mechanisms' \emph{principles} under one harness;
\S\ref{sec:sysmain} runs Mem0, Graphiti and LangMem themselves and a
reimplementation of ERASE.

\section{The benchmark}
\label{sec:bench}

\subsection{Task and metrics}

A benchmark instance is a chronological stream of \emph{statements}, each an
(embeddable) natural-language assertion carrying a metadata timestamp, and a
set of \emph{value-agnostic queries}, one per fact, asking for the fact's
current value. A policy consumes the stream (deciding what to store, keep, or
overwrite; optionally under a slot budget $B$) and answers each query with a
ranking of stored items. With ground truth ``the set of stored versions whose
value is current at query time'' (defined by the source; a value that recurs,
A$\to$B$\to$A, is scored by \emph{value}, not emission index), the top-1 is
scored as \textbf{current@1} (right fact, current value), \textbf{stale}
(right fact, outdated value --- the isolated failure), or \textbf{miss} (wrong
fact). Over the top-$k$ we report \textbf{current@k} and
\textbf{contamination@k} --- whether the top-$k$ contains both a current and a
stale version of the queried fact --- and, after the stream, \textbf{slots}
(store size). \S\ref{sec:reader} adds reader-side accuracy on the same
queries, so contamination's cost is measured rather than assumed.

contamination@k is \emph{not} monotone in context quality --- a policy can improve it by deleting the answer --- so we never report it without current@k beside it (Appendix~\ref{app:benchdetails}). Table~\ref{tab:terms} collects the terms used from here on.

\begin{table}[t]
\centering\footnotesize
\setlength{\tabcolsep}{3pt}
\begin{tabular}{@{}p{2.3cm}p{4.6cm}@{}}
\toprule
term & meaning \\
\midrule
stale vs.\ miss & top-1 names the right fact with a superseded value, vs.\ the wrong fact \\
contamination@$k$ & the top-$k$ holds both a current and a stale version of the queried fact \\
hedged & a context or answer names both the current and a superseded value \\
anchor & the subject's full name on a statement; withdrawing it ($a$) leaves a short form \\
keyer & decides whether an incoming statement describes a fact already stored \\
mechanism & what happens then: overwrite the slot, or invalidate and keep it \\
write / read path & where the keyer runs: on incoming statements, on the query, or both \\
controlled pair & two rungs identical at rank 1 on every query, differing only in contamination \\
\bottomrule
\end{tabular}
\caption{Terms used throughout.}
\label{tab:terms}
\end{table}

\subsection{Real data: TempLAMA instantiation}

TempLAMA \citep{dhingra2022templama} provides subject--relation--object
triples with yearly validity (2010--2020) derived from Wikidata qualifiers.
From \nFactKeys{} fact keys we keep those that are single-valued in every year
(crisp top-1 scoring; \nSingle{} keys) and whose value changes at least once
in the window: \textbf{\nChanging{} facts} over nine relations (teams,
positions, employers, parties, coaches, chairs, heads of government, owners,
schools), mean 2.1 (max 6) distinct values. Consecutive years with the same
value collapse to one \emph{run}, verbalized as a statement from
relation-specific templates; the canonical template is the TempLAMA cloze with
the value filled in, and with probability $p$ (\texttt{paraphrase\_p}) a
paraphrase template is used instead, so similarity-based same-key detection
must survive surface variation. Statements carry the year only as metadata.
The query is a value-agnostic current-value question naming the subject
(``Which team does $S$ currently play for?''); gold is the value valid in the
final year --- a label from Wikidata, untouched by any system under test.
Facts split by key hash into val (\nVal, all tuning) and test (\nTest, all
reported numbers). Three larger configurations bound this one: a corpus with
never-queried distractor facts (Appendix~\ref{sec:fullcorpus}), a
\invBigN{}-entry inventory (Appendix~\ref{app:symx}), and a
\scaleNTest{}-fact scale track (Appendix~\ref{sec:scale}).

\subsection{The anchor-free axis}
\label{sec:anchoraxis}

Templated verbalization repeats the subject's surface form on every statement,
so matching that string is a nearly perfect same-fact detector
(\S\ref{sec:real}); real logs introduce an entity once and then refer to it in
short form. We therefore make the anchor a knob: with probability $a$
(\texttt{alias\_p}) a \emph{non-introducing} statement names its subject by an
alias derived mechanically from the full name --- the head token
(\emph{Ebrard}), the leading token (\emph{Marcelo}), an initialed form
(\emph{M.\ Ebrard}), or a drop-last shortening (\emph{Los Angeles} for
\emph{Los Angeles Rams}).

Details of the controlled diagnostic generator, the distractor corpus and inventory configurations, and the tuning guardrails are in Appendix~\ref{app:benchdetails}.

\section{Policies}
\label{sec:policies}

\textbf{Store-all victims.} \emph{BM25} \citep{robertson2009bm25} and
\emph{Cosine} (dense retrieval over sentence embeddings; all-MiniLM-L6-v2
primary, all-mpnet-base-v2 robustness check \citep{reimers2019sbert}) store
every statement and rank by similarity.

\textbf{Time-aware store-all (the cheap fixes).} \emph{Time-rerank$(k)$}
retrieves the top-$k$ by similarity and answers with the most recent
candidate; the candidate \emph{set} is unchanged, so contamination@$j$ for
$j\le k$ is identical to the victim's by construction. \emph{Recency-weighted}
scores $\mathrm{sim} + \lambda\,d^{\,\mathrm{age}}$
\citep{park2023generative}. Both apply to BM25 as well.

\textbf{Retention = a keyer $\times$ a mechanism.} Retention keeps one slot
per fact; we factor it into the two choices it makes (Table~\ref{tab:terms}).
The \emph{keyer} decides whether an incoming statement describes a fact
already stored: \emph{threshold} (nearest stored neighbour within cosine
$\tau$ --- what vector-store memories do), \emph{gazetteer} (longest subject
from a public inventory appearing verbatim), \emph{fuzzy linker} (the canonical inventory and \emph{nothing else}: candidate
mentions are maximal runs of capitalized tokens, linked to the inventory entry
with the highest character-trigram Jaccard similarity above a val-tuned
threshold --- deliberately ordinary machinery, no learning, no curated
aliases), \emph{entity linker} (the same inventory plus every alias derivable from it; its two ambiguity-resolution variants differ by at most \aliasSkipRecentMaxGap{} points), \emph{LLM} (a model is shown the
top-4 candidates and asked which, if any, is the same fact --- it sees only
statement text, never ids or timestamps), and \emph{oracle} (the true fact key). The oracle bounds \emph{keying}, not
current@1: it still ranks by cosine, so it mis-picks a small fraction of facts,
and a policy with a better ranking function can exceed it. The \emph{mechanism} decides what happens
then: \emph{overwrite} replaces the slot destructively, while
\emph{temporal-KG} marks the superseded statement invalid and keeps it, so
retrieval prefers valid edges and false merges stay recoverable --- the
Zep/Graphiti design \citep{rasmussen2025zep}. \emph{Recency (FIFO)} under
budget $B$ is the degenerate retention policy.

\section{Results}
\label{sec:results}

\subsection{Real data: the failure and the cheap fixes}
\label{sec:real}

\begin{table*}[t]
\centering
\caption{TempLAMA test split (\nTest{} facts), paraphrase $p{=}0.5$. Mean over
queries; brackets = bootstrap 95\% CIs; tuned hyperparameters from val
($k{=}\tunedRerankK$ for time-rerank, $\tau{=}\tunedTau$ for the threshold
keyer). Reconsolidation ($\tau$) is the threshold keyer and Lexical key the
gazetteer keyer, both with overwrite. All four pre-specified paired comparisons (sign-flip permutation,
Holm-corrected; the family is hard-coded before test evaluation and its
store-all comparator is selected on val) are significant at $p<10^{-3}$. Two go
against our own policies: time-rerank $>$ cosine, and the best store-all fix
$>$ the threshold keyer on current@1; two go with them: the threshold keyer $<$
the best store-all fix on contamination@5, and oracle $>$ threshold keyer.}
\label{tab:real}
\footnotesize
\setlength{\tabcolsep}{4pt}
\begin{tabular}{lcccccc}
\toprule
policy & current@1 $\uparrow$ & current@5 $\uparrow$ & stale@1 $\downarrow$ & miss@1 $\downarrow$ & contam@5 $\downarrow$ & slots $\downarrow$ \\
\midrule
BM25 (store-all) & $0.400$ {\scriptsize[0.355,0.444]} & $1.000$ & $0.592$ & $0.009$ & $0.991$ & $995$ \\
BM25 + time-rerank & $0.936$ {\scriptsize[0.912,0.957]} & $1.000$ & $0.051$ & $0.013$ & $0.991$ & $995$ \\
Cosine (store-all) & $0.464$ {\scriptsize[0.419,0.506]} & $0.985$ & $0.530$ & $0.006$ & $0.974$ & $995$ \\
Cosine + time-rerank & $0.925$ {\scriptsize[0.900,0.947]} & $0.985$ & $0.049$ & $0.026$ & $0.974$ & $995$ \\
Cosine + recency-weight & $0.784$ {\scriptsize[0.746,0.821]} & $0.991$ & $0.203$ & $0.013$ & $0.936$ & $995$ \\
Reconsolidation ($\tau$) & $0.859$ {\scriptsize[0.827,0.889]} & $0.929$ & $0.073$ & $0.068$ & $0.137$ & $513$ \\
Lexical key (gazetteer) & $0.972$ {\scriptsize[0.957,0.985]} & $0.985$ & $0.004$ & $0.024$ & $0.000$ & $467$ \\
Keep-latest oracle & $0.979$ {\scriptsize[0.966,0.991]} & $0.991$ & $0.000$ & $0.021$ & $0.000$ & $468$ \\
\bottomrule
\end{tabular}

\end{table*}

Table~\ref{tab:real} is the headline. Store-all similarity retrieval fails on
real updating facts exactly as the value-agnostic analysis predicts: cosine
answers with a stale version on \realCosineStale{} of queries (BM25:
\realBmStale), with no adversarial construction --- only facts that changed
and queries that name the entity rather than the answer. The cheap fixes then
do what a practitioner would hope: reranking the top-2 similar hits by
timestamp lifts current@1 to \realRerankCur{} (BM25 variant:
\realBmRerankCur), and recency-weighted scoring reaches \realRwCur. Four
measured boundaries keep this from closing the problem: contamination@5 stays
between \realRwContam{} and \realBmContam{} for every store-all policy,
because reranking demotes stale versions within an unchanged candidate set
rather than removing them; the repair is concentrated on easy facts
(Appendix~\ref{sec:strat}); it assumes trustworthy timestamps, as retention
assumes in-order arrival (Appendix~\ref{sec:tshuffle}); and its gain does
not transfer to a reader (\S\ref{sec:reader}).

On the retention side, the gazetteer keyer reaches \realLexCur{} with
contamination \realLexContam{} at \realLexSlots{} slots, against
\realOracleCur{} for the true-key oracle (a difference we cannot reject:
Holm-corrected sign-flip $p\,\aliasPLexOracleZero$, overlapping CIs), while
the threshold keyer manages \realReconCur{}. Read naively, same-fact detection
is solved by string matching; \S\ref{sec:anchor} shows that reading is an
artifact of templated text.

One property of this configuration must be stated beside its table: \texttt{anchored} is \emph{saturated} on current@1. A rule that recovers the subject from the canonical question, keeps every statement containing it verbatim, and answers with the most recent scores \blindAnchored{} --- at or above the true-key oracle --- while being maximally contaminated (contamination@5 $= 1.0$ at \realStoreSlots{} slots), the arrangement \S\ref{sec:reader} prices as the worst measured at the reader. Saturation is a property of the retrieval metric, not of the task: reader accuracy across policies on this same split spans \pairPriRead{} to \readerOracleCorrect{}, which is why the reader track lives here, and why \texttt{anchor\_free} (\S\ref{sec:anchor}), where the strongest blind rule, a trigram mention linker, scores \blindAnchorFree{}, is the configuration a leaderboard should rank on. Both blind rules ship as reference baselines (see Limitations).

\subsection{Anchor-free: what same-fact detection actually costs}
\label{sec:anchor}

\begin{table*}[t]
\centering
\caption{current@1 as the lexical anchor is withdrawn ($a=$
\texttt{alias\_p}: the probability a non-introducing statement uses a short
form). Test split, $p{=}0.5$, every policy re-tuned on val at every level. The
last column is the \emph{keying deficit} at $a{=}1$: points of current@1 below
the true-key oracle, which isolates the same-fact decision from the retrieval
difficulty that aliasing also creates (and which the oracle pays too). The
oracle is a \emph{write-path} oracle --- perfect same-fact detection, ordinary
similarity retrieval --- so it is an upper bound only within that family, and
the rungs that also resolve the query post a negative deficit against it
(\S\ref{sec:anchor}).}
\label{tab:alias}
\footnotesize
\setlength{\tabcolsep}{4pt}
\begin{tabular}{lccccc|c}
\toprule
policy & $a{=}0$ & $0.25$ & $0.5$ & $0.75$ & $a{=}1$ & keying deficit at $a{=}1$ \\
\midrule
BM25 (store-all) & $0.400$ & $0.321$ & $0.248$ & $0.126$ & $0.062$ & --- \\
BM25 + time-rerank & $0.936$ & $0.891$ & $0.840$ & $0.776$ & $0.726$ & --- \\
Cosine (store-all) & $0.464$ & $0.391$ & $0.282$ & $0.177$ & $0.111$ & --- \\
+ time-rerank & $0.925$ & $0.870$ & $0.831$ & $0.784$ & $0.737$ & --- \\
+ recency-weight & $0.784$ & $0.679$ & $0.585$ & $0.513$ & $0.419$ & --- \\
Overwrite: threshold $\tau$ & $0.859$ & $0.765$ & $0.671$ & $0.605$ & $0.513$ & $31.4$ \\
Overwrite: gazetteer (exact) & $0.972$ & $0.752$ & $0.575$ & $0.333$ & $0.169$ & $65.8$ \\
Overwrite: fuzzy linker (no alias table) & $0.968$ & $0.908$ & $0.846$ & $0.767$ & $0.714$ & $11.3$ \\
Overwrite: linker + alias table & $0.972$ & $0.927$ & $0.861$ & $0.812$ & $0.776$ & $5.1$ \\
Read-path key (no overwrite) & $0.989$ & $0.953$ & $0.915$ & $0.865$ & $0.831$ & $-0.4$ \\
Symmetric key (read+write) & $0.989$ & $0.953$ & $0.915$ & $0.865$ & $0.831$ & $-0.4$ \\
Symmetric key + alias table & $0.994$ & $0.974$ & $0.944$ & $0.936$ & $0.912$ & $-8.5$ \\
Overwrite: true-key oracle & $0.979$ & $0.942$ & $0.895$ & $0.859$ & $0.827$ & --- \\
\bottomrule
\end{tabular}

\end{table*}

Table~\ref{tab:alias} (drawn as Figure~\ref{fig:ladder}, Appendix~\ref{app:symx}) is the paper's central measurement. Withdrawing the
anchor separates rungs that are indistinguishable when it is present.

When anchored, the top three rungs cannot be told apart --- gazetteer vs.\
oracle raw $p\,\aliasPLexOracleRawZero$, entity linker vs.\ oracle raw
$p\,\aliasPEntityOracleRawZero$, the two keyers against each other
$p\,\aliasPEntityLexZero$ (sign-flip permutation; we quote the raw $p$ here
because Holm over this family inflates the first two to
\aliasPLexOracleZero{}, which reads as a stronger statement than the data
makes. These are failures to reject, not demonstrations of equivalence). When it is withdrawn they separate
at $p\,\aliasPLexOracleOne$, and they separate by very different amounts.

The \textbf{gazetteer} keyer falls from \aliasLexCurZero{} to
\aliasLexCurOne{} (deficit \aliasLexDefOne{}); the retrieval-free audit
(Appendix~\ref{sec:keying}) shows why: it can key only the introducing
statement and scatters \auditLexSplitOne\% of facts across slots, so its parity
with the oracle in Table~\ref{tab:real} was a property of templated text, which
any benchmark with a repeated subject will report. The \textbf{threshold}
keyer --- what vector-store memories ship --- degrades from
\aliasReconCurZero{} to \aliasReconCurOne{} (deficit \aliasReconDefOne{}),
because the similarity between ``\emph{Marcelo Ebrard} serves as $X$'' and
``\emph{Ebrard} now occupies the office of $Y$'' is carried by the surface
form the alias removes. Both rungs are what deployed systems use, and both pay
tens of points.

Above both sits the best a write-path keyer does without an alias table. The
\textbf{fuzzy linker} gets the canonical subject inventory and no alias table,
and must resolve short forms itself with character-trigram similarity: it
reaches \aliasFuzzyCurOne{} at $a{=}1$ (deficit \aliasFuzzyDefOne{}), recovering
most of the gazetteer's collapse without being told what any alias is. That is
the number a future write-path keyer should be measured against.

The \textbf{entity linker}, handed the alias inventory for free --- exactly
the resource the other rungs lack --- reaches \aliasEntityCurOne{} at $a{=}1$,
\aliasEntityDefOne{} points from the oracle, splitting \auditEntitySplitOne\%
of facts. Anchor-free same-fact detection is therefore not intractable: it is
an \emph{entity-resolution} problem, solvable to within a few points by
machinery the memory literature has not adopted, and catastrophic without it.

\paragraph{The ladder resolves entities on one path only.} Every rung above
keys on the \emph{write} path and then answers queries by embedding
similarity, exactly as deployed memory systems do. That is a choice, not a
necessity: the query is text naming the same entity, and the same resolver can
run on it. We add two rungs that do, using no resource the fuzzy linker lacks
(same inventory, same $\tau$, no alias table): \textbf{read-path} keying
routes the query through the trigram matcher and ranks the matched key's
slots by recency while retaining everything; \textbf{symmetric} keying does
that on top of the ordinary write-path merge.

The two paths buy different things. Read-path keying fixes top-1: at $a{=}1$
it cuts the fuzzy rung's miss rate from \aliasFuzzyMissOne{} to
\aliasFuzzyReadMissOne{} and lifts current@1 from \aliasFuzzyCurOne{} to
\aliasFuzzyReadCurOne{}, because it stops asking an embedding to match a short
form it never saw. Write-path keying fixes the context: alone it holds
contamination@5 at \aliasFuzzyContamOne{} in \aliasFuzzySlotsOne{} slots,
whereas read-path keying without it retains everything
(\aliasFuzzyReadSlotsOne{} slots, contamination@5 \aliasFuzzyReadContamOne{}:
a store-all context with a good first hit). Only the symmetric rung gets both
(\aliasFuzzyRWCurOne{}, contamination@5 \aliasFuzzyRWContamOne{},
\aliasFuzzyRWSlotsOne{} slots) --- two rungs that tie on current@1 while
differing by \aliasFuzzyReadContamOne{} against \aliasFuzzyRWContamOne{} in
contamination@5 are not interchangeable (\S\ref{sec:reader}), and current@1
alone would call them equal.

Symmetric keying also beats the rung handed a complete alias table on the write
path --- \aliasFuzzyRWCurOne{} against \aliasEntityCurOne{} here
($p\,\aliasPRwEntityOne$), and at the Appendix~\ref{sec:scale} scale, where the
comparison is better powered, deficit \scaleFuzzyRWDefOne{} against
\scaleAliasDefOne{}. \textbf{Ordinary resolution applied symmetrically beats a
complete alias table applied asymmetrically} --- on mechanical short forms.
With the natural short forms of Appendix~\ref{app:natural}, nicknames and
misspellings a trigram matcher cannot link, the alias table wins by two to
three points, and the linker applied symmetrically is best of all.

Three conditions bound this result, each discussed in Appendix~\ref{app:symx}: queries here always name their subject in full, so the read side is the \emph{easy} direction for a resolver; the read-path rungs resolve the query over the same \invSmallN{}-entry inventory as the fuzzy rung, and a deployment whose inventory lacks the queried subject pays on both paths; and \texttt{keep-latest-oracle} is a \emph{write-path} oracle, so the rungs that resolve the query are not bounded by it --- which is why Table~\ref{tab:alias}'s deficit column goes negative for them, and at the Appendix~\ref{sec:scale} scale that ordering reverts (\scaleFuzzyRWDefOne{} points).

\subsection{The reader track: what the retrieval metric does not tell you}
\label{sec:reader}

\begin{table*}[t]
\centering
\caption{Reader accuracy when each policy's own top-\readerTopK{} becomes the
context (test split, anchored configuration $a{=}0$, $p{=}0.5$). Retrieval columns are the same policies'
current@1 and contamination@5. Readers are \readerModelList{}; the last two columns
repeat both readers with the stored year surfaced in the context.
Retrieval hyperparameters are inherited from the val tuning of
\S\ref{sec:real} and are never tuned against reader accuracy. The two
read-path rows are the controlled pair: identical retrieval columns except
contamination@5, because they return the same statement at rank 1 for all
\pairNQueries{} queries of this configuration.}
\label{tab:reader}
\footnotesize
\setlength{\tabcolsep}{4pt}
\begin{tabular}{lcccccc}
\toprule
& \multicolumn{2}{c}{retrieval} & \multicolumn{2}{c}{reader accuracy (no dates)} & \multicolumn{2}{c}{dates} \\
\cmidrule(lr){2-3}\cmidrule(lr){4-5}\cmidrule(lr){6-7}
policy & current@1 & cont.@5 & qwen2.5:14b & glm-4.6 & qwen2.5:14b & glm-4.6 \\
\midrule
Cosine (store-all) & $0.464$ & $0.974$ & $0.558$ & $0.611$ & $0.959$ & $0.957$ \\
+ time-rerank & $0.925$ & $0.974$ & $0.329$ & $0.677$ & $0.957$ & $0.944$ \\
Overwrite: threshold & $0.859$ & $0.137$ & $0.874$ & $0.872$ & $0.929$ & $0.915$ \\
Overwrite: gazetteer & $0.972$ & $0.000$ & $0.979$ & $0.974$ & $0.979$ & $0.970$ \\
Read-path key (no overwrite) & $0.989$ & $0.998$ & $0.327$ & $0.684$ & $0.983$ & $0.957$ \\
Symmetric key (read + write) & $0.989$ & $0.000$ & $0.983$ & $0.979$ & $0.983$ & $0.972$ \\
Overwrite: oracle & $0.979$ & $0.000$ & $0.985$ & $0.981$ & $0.985$ & $0.974$ \\
\bottomrule
\end{tabular}

\end{table*}

Every result so far is a statement about which versions reach the context.
Whether that matters depends on what a reader does with them --- something the
knowledge-conflict literature reports is unreliable \citep{chen2022rich,
jin2024tugofwar,wu2024clasheval} but which our metrics have so far only
assumed. This section measures it.

\textbf{Setup and controls.} This track runs on the \emph{anchored}
configuration ($a{=}0$, $p{=}0.5$; saturated on current@1, \S\ref{sec:real}),
deliberately: retrieval has to be high across policies for a reader
difference to be attributable to context composition rather than to
retrieval failures. For each query a policy's top-\readerTopK{}
becomes a reader's context, and the reader answers the same value-agnostic
question; answers are scored by normalized containment against the
source-defined current value and against the fact's superseded values, giving
the correct/stale decomposition retrieval already has. Two controls matter.
TempLAMA facts are real Wikidata facts, so a reader might simply know them:
the \emph{closed-book} condition (no context) scores \readerPriClosedBook{}
for \readerPriName{} and \readerSecClosedBook{} for \readerSecName{} --- far
enough below the context-conditioned numbers that those are not recall of
pretraining, and different enough that readers should be compared within, not
across. And every condition runs twice, with and without the stored year
surfaced in the context --- the reader-level form of the cheapest objection,
\emph{why not show the model the dates?}

\textbf{Retrieval gains do not transfer} (Table~\ref{tab:reader})\textbf{.}
Timestamp reranking improves
retrieval current@1 by \readerRetrievalDelta{} points over store-all cosine
(\S\ref{sec:real}); at the reader it is worth \readerPriRerankDelta{} points
for \readerPriName{} and \readerSecRerankDelta{} for \readerSecName{}. For one
of the two the sign flips: the policy that nearly doubles retrieval accuracy
leaves the end task \emph{worse} than the baseline it was introduced to
repair. We do not claim to know \emph{why} the readers differ --- they differ
in scale and in how much of this data they already know
(\readerPriClosedBook{} against \readerSecClosedBook{} closed book) --- but
for both the retrieval gain does not arrive. Retention behaves oppositely,
worth \readerPriLexDelta{} and \readerSecLexDelta{} points, so the
\gapRerankLexPhalf-point retrieval difference between reranking and gazetteer
keying (\realRerankCur{} vs \realLexCur{}) opens into a reader gap of tens of
points. \textbf{current@1 compresses a difference the reader amplifies}, and a
benchmark reporting current@1 alone misstates how these policies compare.

\textbf{A controlled pair: identical rank 1, \pairPriDelta{} points of
accuracy.} The comparisons above run across policies that differ in many ways
at once, which is the usual objection to reading a correlation causally. The
read-path and symmetric keying rungs of \S\ref{sec:anchor} remove that
objection, because they are the same policy on the read path: both resolve the
query through the same matcher and rank the same key's slots by recency, so
\emph{they return the same statement at rank 1 for all \pairNQueries{} queries}
--- verified per query and pinned by a regression test, not inferred from
matching aggregates. They are identical on current@1 (\pairCur{} for both,
exactly) and identical on stale@1 (exact in the anchored configuration this
track uses; anchor-free the two agree on \pairAnchorFreeAgree{} of
\pairNQueries{} queries, so the controlled claim is made only where it is
exact). They differ only in whether the write path also merges, which changes
only what fills ranks 2--\readerTopK{}: contamination@5 \pairReadContam{}
against \pairRwContam{}. The retain-everything member is a \emph{control}, not
a system we propose --- its purpose is to hold rank 1 fixed while
contamination varies, which no comparison between two independently designed
policies can do.

current@1 cannot tell them apart, and current@5 prefers the
contaminated one, which retains every version and so never lacks the current
one. The reader is not indifferent: \pairPriRw{} against \pairPriRead{} on
\readerPriName{} ($p\,\pairPriP$, sign-flip permutation), a gap of
\pairPriDelta{} points, and \pairSecRw{} against \pairSecRead{} on
\readerSecName{} ($p\,\pairSecP$, \pairSecDelta{} points); with rank 1
identical on every query, that difference is attributable to ranks
2--\readerTopK{} and to nothing else. The contaminated member posts the
highest current@1 of any policy in the track --- above the oracle's
\readerOracleRet{}, since resolving the query is a better way to rank one hit
than perfect retention is --- together with the worst reader accuracy of any
keying policy on both readers, and for \readerPriName{} below even store-all
cosine (\readerCosineCorrect{}), whose current@1 is less than half of it (for
\readerSecName{} it stays above store-all, \pairSecRead{} against
\readerSecCosine{}). A benchmark reporting current@1 would rank
these two as tied for first, and one reporting current@5 would put the pair's
loser alone at the top. The retrieval-side measure of context quality that
separates them correctly is contamination@5 --- the metric this
paper adds, and this pair is the cleanest argument for it.

\textbf{Surfacing dates closes the pair.} With the stored year shown in the
context, reranking's reader moves to \readerRerankDates{}, the
constructed-context penalty of Appendix~\ref{app:readerx} largely disappears (\doseCurrentFirstDates{}
against \doseCurrentFirst{}), and the pair closes: the retain-everything
member goes from \pairPriRead{} to \pairPriReadDates{} while the symmetric
member stays at \pairPriRwDates{}, \pairPriNChangedDates{} individual answers
changing and cancelling\ifdefined\pairSecReadDates{}; for \readerSecName{}
the dated pair reads \pairSecReadDates{} against \pairSecRwDates{}, a gap of
\pairSecDeltaDates{} points\fi{}. Given a trustworthy year on every
statement, the reader resolves the contradiction itself. The claim above is untouched --- it
is about what current@$k$ can see, and the pair is identical at rank 1 on
every query while \pairPriDelta{} points apart without dates --- but
retention should be read as one of two ways to resolve the contradiction, in
the store or in the prompt, and the second is nearly free. Retention is the
one to prefer when the clock cannot be trusted (the dated prompt shares
reranking's assumption, and Appendix~\ref{sec:tshuffle} shows what each
family loses when its temporal signal is corrupted --- retention reads
arrival order, and none is exempt), when the store returns no validity timestamp per memory
(Mem0 does not; Graphiti does, and its default read path discards it,
\S\ref{sec:sysmain}), and under a slot budget, where the versions a
retain-everything store needs are the first evicted (the pair holds
\corrSlotsRead{} against \corrSlotsSym{} slots; Appendix~\ref{sec:controlled}).

\textbf{What this justifies.} The census of every policy pair against every
reader, without surfaced dates (Appendix~\ref{app:readerx}), says the same
thing in aggregate:
restricted to policies a practitioner might deploy, current@1 still inverts
\concordCurInvertStrict{} of \concordCurTotalStrict{} comparisons and
\textbf{contamination@5 inverts \concordContamInvertStrict{}}. The
inferential weight is carried by the controlled pair, at $p\,\pairPriP$ on
\pairNQueries{} paired queries.

Reversing the presentation order of the identical top-$k$ (\readerPriName{} only) moves reranking's value against the store-all baseline from \rerankVsBaseFwd{} to \rerankVsBaseRev{} points, while retention changes zero individual answers: one statement per fact leaves nothing to reorder. That ablation, the constructed-context probe that establishes the mechanism causally on both readers, and the full census are in Appendix~\ref{app:readerx}.

\subsection{Deployed systems, scored three ways}
\label{sec:sysmain}

\begin{table*}[t]
\centering\footnotesize
\setlength{\tabcolsep}{4pt}
\begin{tabular}{lcccccccc}
\toprule
& \multicolumn{4}{c}{anchored} & \multicolumn{4}{c}{anchor-free} \\
\cmidrule(lr){2-5}\cmidrule(lr){6-9}
system & cur@1 & by text & cont.@5 & reader & cur@1 & by text & cont.@5 & reader \\
\midrule
Mem0 2.0.11 & $0.180$ & $0.480$ & $0.947$ & $0.627$ & $0.120$ & $0.153$ & $0.820$ & $0.580$ \\
Mem0 0.1.118 & $0.913$ & $0.573$ & $0.593$ & $0.720$ & $0.653$ & $0.207$ & $0.747$ & $0.500$ \\
Graphiti, as returned & $0.347$ & $0.347$ & $0.947$ & $0.593$ & $0.180$ & $0.180$ & $0.820$ & $0.433$ \\
Graphiti, valid edges first & $0.833$ & $0.833$ & $0.120$ & $0.893$ & $0.640$ & $0.640$ & $0.153$ & $0.773$ \\
LangMem, qwen2.5:14b & $0.773$ & $0.773$ & $0.173$ & $0.833$ & $0.340$ & $0.413$ & $0.213$ & $0.600$ \\
LangMem, glm-4.6 & $0.800$ & $0.147$ & $0.753$ & $0.613$ & $0.627$ & $0.160$ & $0.640$ & $0.413$ \\
ERASE, qwen2.5:14b & $0.853$ & $0.860$ & $0.100$ & $0.867$ & $0.733$ & $0.707$ & $0.160$ & $0.807$ \\
ERASE, glm-4.6 & $0.953$ & $0.953$ & $0.060$ & $0.980$ & $0.713$ & $0.713$ & $0.240$ & $0.813$ \\
\midrule
Cosine (store-all) & $0.460$ & $0.460$ & $0.973$ & $0.573$ & $0.140$ & $0.140$ & $0.927$ & $0.427$ \\
+ time-rerank & $0.920$ & $0.920$ & $0.973$ & $0.300$ & $0.787$ & $0.787$ & $0.927$ & $0.220$ \\
Symmetric key & $1.000$ & $1.000$ & $0.007$ & $1.000$ & $0.920$ & $0.920$ & $0.027$ & $0.920$ \\
Overwrite: oracle & $0.987$ & $0.987$ & $0.007$ & $1.000$ & $0.887$ & $0.887$ & $0.000$ & $0.953$ \\
\bottomrule
\end{tabular}

\caption{Three deployed systems and ERASE, reimplemented, under the rungs'
harness, streams and encoder (test subsample, $n{=}\sysNFacts{}$ facts, $p{=}0.5$). \emph{cur@1}
attributes the top memory to its source statement, \emph{by text} scores what
it says, \emph{cont.@5} is contamination@5 by text, and \emph{reader} is
\sysrdAdjReader{} answering from the system's own top-5. The Mem0 and
Graphiti rows have qwen2.5:14b in the loop. The lower block is the ladder on
the same streams. Per-system tables, gates and the second adjudicator are in
Appendix~\ref{sec:systems}.}
\label{tab:deployed}
\end{table*}

We place four mechanisms on this surface (Table~\ref{tab:deployed}): three
deployed systems and our reimplementation of a published fourth. Mem0 adjudicates the incoming memory, Graphiti
invalidates the edge an arrival contradicts, LangMem updates in place, and
ERASE edits the entries already stored. Each runs under the rungs' harness,
streams and encoder, and is scored three ways: by attributing returned
memories to source statements, by what they say, and by what a reader answers
from them.

\textbf{Mem0.} Attribution and text disagree only on the deployed system; on
the rungs, which return the corpus verbatim, the two agree to within
\attrAdjControlMaxGapZero{} points. Attribution overstates
\probedAdjVersion{}, the release whose ingestion adjudicates, by
\textAdjCurDropZero{} points with qwen2.5:14b in the loop (an UPDATE relabels
a memory without rewriting it, so a current id sits on superseded text; with
glm-4.6 it rewrites the text and the two scorings agree,
Appendix~\ref{sec:systems}) and breaks outright on \probedCurVersion{}
(\attrCurIdWrongPctZero\% of memories carry an id naming a different entity
than their text). What no scoring convention changes is that the ADD/UPDATE
adjudication is absent from \probedCurVersion{}'s ingestion path: its event
histogram, a two-statement probe and the library source say so independently
of any metric. Neither release returns a validity timestamp, so the
dated-prompt alternative of \S\ref{sec:reader} is unavailable to a deployment
reading Mem0's output; applying contribution (3) post hoc to either release's
contexts recovers most of what they lose at no model call, and keeping the
oldest memory per entity instead of the freshest prices the write path itself
at \sysrdAdjSysOldestZero{} for \probedAdjVersion{} against
\sysrdCosineRagOldestZero{} for store-all. Every \mbox{(release, vector
store)} pair is hand-probed first (\sysNGatesPassed{}/\sysNGatesProbed{}
pass).

\textbf{Graphiti} is the mirror image: its invalidation fires on the write
path (\gphInvalidationsZero{} of \gphNSupersessionsZero{} supersessions) and
its default search ignores it, so it scores \gphCurZero{} as installed, below
store-all, and \gphVFCurZero{} with its own validity flags respected; the
reader moves from \gphRdAsIsZero{} to \gphRdVFZero{}.

\textbf{LangMem} applies \lgmQwenUpdatesZero{} updates and keeps
\lgmQwenSlotsZero{} slots with qwen2.5:14b (current@1
\lgmQwenCurZero{} by id and by text). With glm-4.6 it consolidates into
chronologies: the memory a query lands on is current by id
(\lgmGlmCurZero{}) and hedged by text (\lgmGlmHedgedZero{}), and the reader
answers \lgmGlmRdAsIsZero{} from it against \lgmQwenRdAsIsZero{} from the
smaller model's store.

\textbf{ERASE} \citep{li2024erase} edits the store rather than the arrival:
the \ersQwenRetrieveK{} stored facts nearest an incoming statement are
retrieved, and one model call keeps, rewrites or deletes each before the
statement is inserted. No release pins it, so we reimplemented the loop from
the paper under the same probe gate. With glm-4.6 deciding it reaches
current@1 \ersGlmCurZero{} by id and by text at contamination@5
\ersGlmContamTextZero{}, and \ersGlmRdAsIsZero{} at the reader: the best
number any of the four posts here, within \ersGlmGapToOracleZero{}
points of the write-path oracle (\ersGlmGapQueriesZero{} of \sysNFacts{}
queries) with no alias table and no identifiers. With
qwen2.5:14b it reaches \ersQwenCurZero{}, and for \ersQwenGoneZero{} of facts
no version survives anywhere in the returned depth: asked whether a
neighbouring fact is still accurate, the model deleted facts about entities
the arrival did not mention (\ersQwenDeletesZero{} deletions and
\ersQwenRewritesZero{} rewrites against \ersQwenKeepsZero{} keeps).
Anchor-free, the larger model falls to \ersGlmCurOne{}: an arrival written
with a short form does not retrieve the full-name version it supersedes among
its \ersQwenRetrieveK{} neighbours, so the edit never reaches the stale
entry. The mechanism edits only what its retriever finds, which is the
anchor dependence of \S\ref{sec:anchor} on the write path, and the mechanism
that can remove staleness is the one that can remove the fact.

Across the four the pattern is shared and the failure never is: Mem0 lost its
adjudication between releases, Graphiti keeps its verdicts and discards them
on read, LangMem applies them and either keeps everything or rewrites it into
a chronology, depending on the model, and ERASE applies them to the store
and, under qwen2.5:14b, to facts the arrival never mentioned. No single
scoring convention describes all four.

\section{Conclusion}

StaleBench isolates a failure every long-lived agent memory will meet,
measures it on real updating facts with labels no system under test can touch,
and prices each repair. Metadata heuristics fix the top-1 symptom when
timestamps are clean, except on the facts that change most, and leave a
self-contradicting context whose cost at the reader is set by prompt
conventions --- presentation order, whether the year is shown --- that they
do not control; retention removes the contradiction wherever same-fact keying
works and ingestion is in order, each family reading one temporal signal and
blind to the other (Appendix~\ref{sec:tshuffle}). Agent memory attempts
same-fact detection, a problem entity resolution addressed decades ago, with
weaker mechanisms and pays tens of points of freshness for it; running the
resolver on the query too is free and, where the inventory covers the queried
subjects, closes most of that gap (\S\ref{sec:anchor}). The
controlled pair --- identical at rank 1 on every query, \pairSecDelta{} and
\pairPriDelta{} points apart at the reader without surfaced dates --- is the
strongest evidence that current@$k$ alone cannot measure freshness
(\S\ref{sec:reader}).

\label{end:main}

\section*{Limitations}
\label{sec:limits}

\textbf{Saturation and headroom.} \texttt{anchored} and \texttt{full\_corpus}
are saturated on current@1 (a blind rule scores \blindFull{} on the latter);
\texttt{anchor\_free} is not: the strongest blind rule reaches
\blindAnchorFree{} against an oracle at \aliasOracleCurOne{}. The
entity-linker rung is handed the alias inventory the stream drew from, so its
residual \aliasEntityDefOne{} points bound a real linker's gap from below. The
best write-path rung without an alias table is the fuzzy linker, whose deficit of
\aliasFuzzyDefOne{} points roughly doubles on the realistic corpus
(\fullFuzzyDefOne{}, Table~\ref{tab:fullcorpus}). The reader metric has a
ceiling of \readerCeiling{}, because three test facts nest a superseded value
inside the gold value.

\textbf{Scope of the data.} The headline track queries single-valued facts
that change (\nChanging{} of \nFactKeys{} keys). Appendix~\ref{sec:scale} lifts that
restriction and reports set recall beside top-1, since a one-slot policy keeps
\scaleSetRecOracleZero{} of simultaneously valid values against
\scaleSetRecRwZero{} for recency-weighted store-all. Validity is yearly, statements are
templated, and the anchor-free axis uses mechanically derived short forms, so
conversational reference --- pronouns, nicknames, discourse coreference --- is
outside what this instrument measures; Appendix~\ref{app:natural} narrows the
gap with real aliases for the \natCovSubjectsPct\% of test subjects that have
one. We generate the axis because we need it controlled; TempEL
\citep{zaporojets2022tempel} would make it naturalistic. English only; nine
relations; \nTest{} test facts is enough for tight CIs but not for
per-relation claims.

\textbf{Systems track.} This is the narrowest evidence in the paper: four
mechanisms, one of them our reimplementation, on \sysNFacts{} facts,
with a single reader. For Mem0, \probedAdjVersion{} passes both halves of the probe on only one
of three backends, so we
offer a cross-backend check only for \probedCurVersion{}
(\probedCurLo{}--\probedCurHi{} over \probedCurNBackends{} stores) and an
independent repeat for \probedAdjVersion{}, which reproduced it exactly. The
second adjudicator, glm-4.6, moves Mem0's numbers at the extraction stage, and
that comparison holds only under an extraction prompt that admits third-person
facts. That \attrCurIdWrongPctZero\% of \probedCurVersion{}'s memories carry
an id naming a different entity than their text is a measurement without a
mechanism: a six-statement probe of the same release attached metadata
correctly.

\textbf{Temporal assumptions and the reader.} Facts arrive as they become true
and gold is the final value, so most-recent-equals-current holds by
construction in the clean condition. Appendix~\ref{sec:tshuffle} measures
corrupted clocks, corrupted arrival order and late stale reports. Ambiguous
updates, where it is unclear whether a statement supersedes an earlier one,
and corrections with their own semantics, such as a retraction that names an
earlier statement as wrong, are out of scope. The reader track uses
\readerNModels{} models with one prompt format, so its claims are comparative
across contexts under a fixed prompt, not absolute accuracy.
Appendix~\ref{app:limits-ext} gives the full discussion.

\section*{Ethical Considerations}

\textbf{Data and personal information.} The facts are public biographical and
organizational relations about public figures and organizations, taken from
Wikidata via TempLAMA \citep{dhingra2022templama} and released under CC0. They
name real people, which is intrinsic to the task: the failure concerns the
current value of a fact about a named entity, so pseudonymizing subjects would
remove the phenomenon under study. We add nothing beyond what Wikidata
publishes. Party affiliation, a special category of personal data under some
regimes, is among the nine relations; these are publicly recorded affiliations
of office-holders and candidates, and we neither infer nor annotate any.
Aliases are mechanical shortenings of source names or public short forms from
Wikidata and Wikipedia. The release is a frozen 2010--2020 window. Statements
are template-verbalized and no user-authored text is included, so the risk of
offensive content is low.

\textbf{Risks.} A benchmark that makes freshness measurable can be over-read
as certifying it; our central result argues against that reading. Retention
policies \emph{delete} superseded statements, and where the history is itself
the record (audit, compliance, contested claims about a person) destructive
overwrite is the wrong mechanism and a false merge is unrecoverable. We
therefore evaluate invalidation alongside overwrite and report its footprint
cost (Appendix~\ref{sec:keying}). We see no dual-use concern.

\textbf{Compute.} Every number outside the systems track reproduces on a
laptop CPU in under an hour. The \llmCacheEntries{} model calls behind the
LLM tracks are short completions, cached and shipped with the release
(Appendix~\ref{app:repro}).
Appendix~\ref{app:limits-ext} gives the full statement.

\section*{Acknowledgments}
We thank the ARR reviewers and the area chair for comments that shaped this
paper.

\clearpage
\bibliography{refs}

\clearpage
\appendix

\section{Positioning table}
\label{app:position}

Table~\ref{tab:position} positions the benchmark against each adjacent line of work.

\begin{table*}[t]
\centering
\caption{Positioning: what each line of work measures, and what it cannot see.}
\label{tab:position}
\small
\setlength{\tabcolsep}{4pt}
\begin{tabular}{p{0.30\linewidth}p{0.30\linewidth}p{0.32\linewidth}}
\toprule
line of work & asks & does not model \\
\midrule
TempLAMA; FreshQA; RealTime QA; StreamingQA; DyKnow; GrowOVER
\citep{dhingra2022templama,vu2023freshllms,kasai2023realtimeqa,liska2022streamingqa,mousavi2024dyknow,ko2024growover}
& is the \emph{model/corpus} current with the world? & an agent-owned store;
the retention decision; budgets \\
knowledge editing \citep{meng2022rome,meng2023memit} & can we rewrite
\emph{weights}? & the store entirely \\
knowledge conflict \citep{chen2022rich,jin2024tugofwar,wu2024clasheval,xu2024knowledge}
& how does a \emph{reader} behave given conflicting context? & which version
\emph{reaches} the context, and at what memory cost \\
entity linking, coreference, record linkage
\citep{wu2020blink,joshi2019coref,zaporojets2022tempel,fellegi1969theory,li2020ditto}
& are these two mentions/records the same entity? & the \emph{price} of
getting it wrong in a downstream retention decision \\
LoCoMo; LongMemEval \citep{maharana2024locomo,wu2025longmemeval} & broad
long-term memory ability & staleness in isolation (multi-session gold dilutes
it) \\
memory systems \citep{packer2023memgpt,chhikara2025mem0,rasmussen2025zep,yan2025memoryr1,li2024erase,gutierrez2024hipporag}
& ship freshness \emph{mechanisms} & an isolating measurement; oracle-priced
keying \\
\textbf{StaleBench} & does the store surface its own most recent knowledge,
what does each fix cost, and what does it cost the reader? & naturalistic
dialogue (see Limitations); multi-valued facts only as set recall
(Appendix~\ref{sec:scale}) \\
\bottomrule
\end{tabular}
\end{table*}

\section{Benchmark details}
\label{app:benchdetails}

One property of contamination@k must be stated plainly because it constrains
how the metric may be read: it is \emph{not} monotone in context quality. A
top-$k$ holding only stale versions scores $0$, the same as a top-$k$ holding
only the current one, because nothing in it contradicts anything else. A policy
can therefore improve contamination by deleting the answer. We accordingly
never report contamination@k without current@k beside it, and a policy that
wins on contamination while losing on current@k is winning by omission, not by
resolution --- the two together separate current-only, contaminated and
current-absent contexts.

\subsection{Controlled generator}

Each of $E$ entities carries a value that updates with per-step probability
$p_u$; an event embedding mixes a unit key direction $k_e$ (the entity) with a
value direction: $x = \mathrm{norm}(\tilde k_e + g\,v_{e,\mathrm{ver}})$,
where $g$ (\texttt{value\_gap}) sets how visibly an update moves the
embedding, and $\tilde k_e = \mathrm{norm}(k_e + j\,\varepsilon)$ jitters the
key per event with $j$ (\texttt{key\_jitter}) --- the controlled analog of
surface variation. Queries point at the clean key only (value-agnostic by
construction). Labels come from the generator's update schedule. The generator
is a \emph{diagnostic} for mapping preconditions; headline claims live on real
data.

\subsection{Alias inventory}

Alias candidates that are bare organizational type words
(\emph{FC}, \emph{Line}, \emph{Ministry}) are rejected, since no writer uses
them as standalone references and keeping them would make the track an unfair
strawman rather than a measurement.

Three properties keep the axis interpretable. The \emph{first} statement about
each fact always carries the full name and every query always uses it, so no
fact becomes unidentifiable in principle --- what changes is the cost of
identifying it. Aliases that collide across subjects (\emph{Harris}) are
deliberately \emph{kept} (\aliasAmbigPct\% of \aliasNForms{} forms, touching
\aliasSubjAmbigPct\% of subjects), because entity-resolution ambiguity is part
of what we are pricing. And the inventory is public: it ships with the
dataset, so a keyer can be handed it for free and the residual gap attributed
to resolution rather than to missing information. \aliasCoverage\% of subjects
admit at least one alias; at $a{=}1$, \aliasNEventsOne{} of
\aliasNEventsTotal{} test statements are anchor-free.

\subsection{Corpus size and inventory}

A store containing only the facts it will be asked about is an unrealistically
easy store. The \texttt{full\_corpus} configuration adds every TempLAMA fact
key excluded from the queried set (those multi-valued in some year) as
\emph{distractor} statements: \fullNDistract{} extra statements, a
\fullNEvents-statement stream. Distractors are never queried and need no
labels, but they populate the neighbourhood that same-fact detection must
avoid merging across. This configuration also separates two claims that the
small corpus conflates, by giving keyers either an \emph{open-world}
inventory (the \fullNInvOpen{} queried subjects only, so most of the corpus is
out-of-inventory) or a \emph{closed-world} one (all \fullNInvClosed{} subjects
in the stream).

\subsection{Guardrails}

Four design rules keep the benchmark from favouring its own baselines: currency
labels are source-defined, never derived from retrieval scores; every
hyperparameter of every policy ($\tau$, $k$, $\lambda$, decay) is tuned on the
val split only, by current@1, and re-tuned separately in every regime, so the
store-all fixes are given their best shot everywhere; reader-track
hyperparameters are inherited from that retrieval tuning and never tuned
against reader accuracy; and results always report where the retention fixes
\emph{fail}.

\section{Read-path and symmetric keying: additional analysis}
\label{app:symx}

\begin{figure*}[t]
\centering
\includegraphics[width=0.6\textwidth]{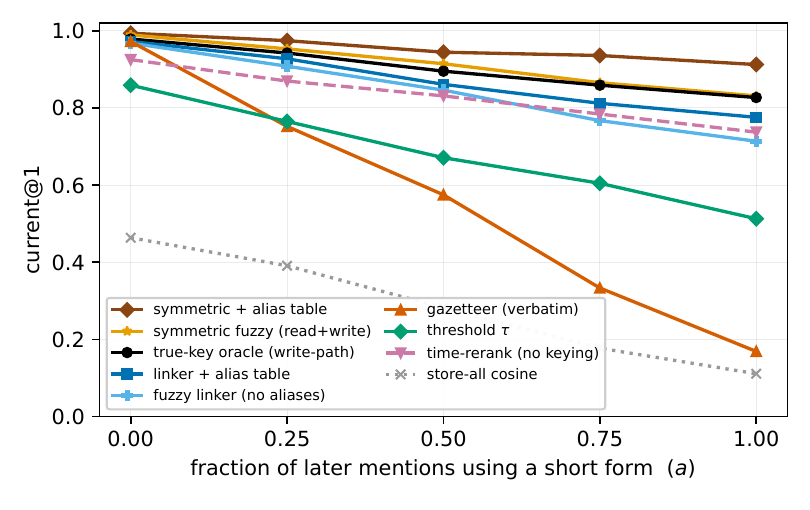}
\caption{The keying ladder as the lexical anchor is withdrawn. The
inventory-based keying rungs are indistinguishable at $a{=}0$, and at
$a{=}1$ the gazetteer falls \aliasLexDefOne{} points below the oracle; the entity linker (handed the alias inventory) tracks the oracle,
while the two mechanisms deployed memory systems use do not. The oracle keys
perfectly but still \emph{retrieves} by similarity, so the two rungs that
resolve the query as well are not bounded by it --- and the upper one, which
resolves symmetrically with a complete alias table, is close to flat across the
whole axis (\aliasEntityRWCurZero{} to \aliasEntityRWCurOne{}). Withdrawing
the anchor costs almost nothing to a system that resolves entities on both
paths.}
\label{fig:ladder}
\end{figure*}

Neither the entity linker's \aliasEntityDefOne{}-point deficit nor the fuzzy
linker's \aliasFuzzyDefOne{} (\S\ref{sec:anchor}) should be read as
\emph{the} residual, because both are conditioned on a second variable the
ladder holds fixed: how complete the inventory is. The alias rung's is exhaustive, correct, free, and generated by
the same function that inserted the aliases, so it has no out-of-vocabulary mass
at all. The fuzzy rung is handed no alias table --- but its inventory is still
exactly the \invSmallN{} queried subjects, which is to say the index of the
answer key. Holding the corpus fixed and enlarging that inventory to
\invBigN{} entries costs it \invSmallDef{} $\to$ \invBigDef{} points, and on
the realistic corpus its deficit is \fullFuzzyDefOne{}
(\S\ref{sec:fullcorpus}).

The object is therefore a surface, not a scalar. Same-fact detection costs
roughly \aliasLexDefOne{} points with exact matching, \aliasReconDefOne{} with
embedding thresholds, and \invSmallDef{}--\invBigDef{} with ordinary fuzzy
entity resolution depending on how much of the world the inventory covers ---
and a system that knows exactly which entities it will be asked about is at the
optimistic corner of that surface. Inventory completeness is a priced axis of
this benchmark, not a caveat on one number.

The comparison \S\ref{sec:anchor} draws --- ordinary resolution on both paths
against a complete alias table on the write path alone --- changes one thing
and confounds two, the path and the resolver, so we run the control that
separates them: the same complete alias table, applied on \emph{both} paths.
It reaches \aliasEntityRWCurOne{}, and the two terms come apart cleanly. Making
resolution symmetric is worth \symGainFuzzy{} points to the fuzzy resolver and
\symGainEntity{} to the alias-equipped one (both $p\,\aliasPRwFuzzyOne$),
\emph{and} resolver quality is still worth \resolverGainSym{} points once both
are symmetric ($p\,\aliasPEntityRwFuzzyRwOne$). So the asymmetry is not a
substitute for better entity resolution and better entity resolution is not a
substitute for fixing the asymmetry: they are separable, both large, and only
one of them is free. Part of what this section prices as the cost of imperfect
same-fact detection is the cost of resolving entities when writing and not when
reading --- recoverable at no cost in resolver quality --- and the rest is the
resolver, which the field should keep working on. None of this is new machinery. Linking a query to a
knowledge-base entity and then retrieving by key rather than by embedding is a
long-standing pattern in entity-oriented and KB question answering; what is new
here is not the mechanism but its \emph{price} in this setting --- that agent
memory does not do it, that doing it costs nothing it does not already have,
and that it is worth more than a complete alias table on the write path. This
is also not query expansion. Read-side rewriting --- HyDE and its
descendants \citep{gao2023hyde} --- improves the query's \emph{embedding} so
that similarity search retrieves better, and stays inside the similarity
paradigm. The read-path rung does the opposite: it stops using similarity for
this query, resolves the query to a store key, and orders that key's contents by
recency. The two are complementary --- a better query embedding would help the
similarity fallback these rungs still use when resolution fails --- but the
mechanism being priced here is resolution, not expansion, and it is cheap
precisely because it does not involve a generative model.

One scope condition belongs here rather than in the limitations, because it
cuts against the rungs we are advocating: queries in this benchmark always name
their subject in full, so the read side is the \emph{easy} direction for a
resolver. That is what makes the comparison informative rather than
flattering --- even where resolution is easiest, no system we know of does it
--- but a deployment whose queries are themselves elliptical (``and who does he
play for now?'') would pay on the read path too, and we do not measure that.

These rungs do not escape
the inventory axis, either: they resolve the query with the same matcher over
the same \invSmallN{}-entry inventory as the fuzzy rung, so they sit at the
same optimistic corner of the surface above, and a deployment whose inventory
does not already contain the queried subject pays there on both paths rather
than one. That is measured rather than assumed, and it is the one place where
the paper's headline \emph{surface} and its headline \emph{fix} would
otherwise sit on different policies. Run along the same inventory axis, the
symmetric rung's anchor-free deficit goes \invSmallDefRW{} $\to$
\invBigDefRW{} points as the inventory grows \invSmallN{} $\to$ \invBigN{},
against \invSmallDef{} $\to$ \invBigDef{} for the write-path rung: the
advantage is between \invRWAdvantageMin{} and \invRWAdvantageMax{} points of
current@1 at \emph{every} inventory size, and both rungs degrade at
essentially the same rate. Symmetric keying is therefore uniformly better
without being a way out of the inventory problem --- it pays on two paths
where the ladder paid on one, which is why its degradation is if anything
slightly steeper. The same holds on the realistic corpus with an open-world
inventory, the hardest configuration here: \fullFuzzyRWOpenOne{} anchor-free
for a deficit of \fullFuzzyRWDefOne{} points against the write-path rung's
\fullFuzzyDefOne{} (\S\ref{sec:fullcorpus}).

\paragraph{What this does to our oracle.} It also bounds our own metric, in our
disfavour. \texttt{keep-latest-oracle} has perfect same-fact detection but still
\emph{retrieves} by similarity, so it is a write-path oracle, not an end-to-end
one: at $a{=}1$ it misses \aliasOracleMissOne{} of queries outright because the
aliased query does not embed near its one retained slot. The symmetric rung,
with no oracle knowledge, is statistically indistinguishable from it on
current@1 (\aliasFuzzyRWCurOne{} vs.\ \aliasOracleCurOne{},
$p\,\aliasPRwOracleOne$ --- a failure to reject, not a demonstration of
equivalence) by trading misses for stale answers --- its stale rate is \aliasFuzzyRWStaleOne{}
where the oracle's is zero. So the keying deficit reported throughout this
section is the distance to the best \emph{write-path} policy, and is a valid
comparison within that family, which is the family deployed systems belong to.
It is not a floor on the task. At the \S\ref{sec:scale} scale the ordering
reverts --- the oracle leads by \scaleFuzzyRWDefOne{} points --- which is the
same trade resolving the other way, and is why we report the deficit against a
write-path oracle rather than claiming the task is solved.

Finally, and against our own framing: in the anchor-free regime the cheap
metadata fix is competitive on current@1 \emph{within the write-path family}. Time-rerank reaches
\aliasRerankCurOne{} against \aliasEntityCurOne{} for the entity linker,
because reranking needs no keying at all --- though it no longer holds against
the read-path rungs of this section (\aliasFuzzyRWCurOne{} and
\aliasEntityRWCurOne{}), which is one more reason the asymmetry was worth
finding. Retention's advantage there is not
top-1 freshness but contamination (\aliasEntityContamOne{} against
\aliasRerankContamOne{}) and footprint --- and \S\ref{sec:reader} shows
contamination is what the reader pays for when no date is shown.

\begin{table*}[t]
\centering
\caption{Where each policy's errors go in the anchor-free regime ($a{=}1$,
test, $p{=}0.5$). contamination@5 is shown with current@5 because it is not
monotone (\S\ref{sec:bench}): the entity linker's near-zero contamination is
partly the absence of any version of the fact, not only the absence of
conflict.}
\label{tab:aliasfail}
\footnotesize
\setlength{\tabcolsep}{4pt}
\begin{tabular}{lcccccc}
\toprule
policy ($a{=}1$) & cur@1 $\uparrow$ & stale@1 $\downarrow$ & miss@1 $\downarrow$ & cur@5 $\uparrow$ & cont.@5 $\downarrow$ & slots $\downarrow$ \\
\midrule
BM25 (store-all) & $0.062$ & $0.925$ & $0.013$ & $0.895$ & $0.893$ & $995$ \\
BM25 + time-rerank & $0.726$ & $0.113$ & $0.160$ & $0.895$ & $0.893$ & $995$ \\
Cosine (store-all) & $0.111$ & $0.868$ & $0.021$ & $0.891$ & $0.882$ & $995$ \\
+ time-rerank & $0.737$ & $0.105$ & $0.158$ & $0.891$ & $0.882$ & $995$ \\
+ recency-weight & $0.419$ & $0.199$ & $0.382$ & $0.650$ & $0.404$ & $995$ \\
Overwrite: threshold $\tau$ & $0.513$ & $0.297$ & $0.190$ & $0.771$ & $0.261$ & $561$ \\
Overwrite: gazetteer (exact) & $0.169$ & $0.808$ & $0.024$ & $0.893$ & $0.784$ & $936$ \\
Overwrite: fuzzy linker (no alias table) & $0.714$ & $0.120$ & $0.167$ & $0.801$ & $0.032$ & $467$ \\
Overwrite: linker + alias table & $0.776$ & $0.053$ & $0.171$ & $0.865$ & $0.024$ & $469$ \\
Read-path key (no overwrite) & $0.831$ & $0.124$ & $0.045$ & $0.987$ & $0.983$ & $995$ \\
Symmetric key (read+write) & $0.831$ & $0.122$ & $0.047$ & $0.868$ & $0.034$ & $467$ \\
Symmetric key + alias table & $0.912$ & $0.058$ & $0.030$ & $0.932$ & $0.024$ & $469$ \\
Overwrite: true-key oracle & $0.827$ & $0.000$ & $0.173$ & $0.925$ & $0.000$ & $468$ \\
\bottomrule
\end{tabular}

\end{table*}

Table~\ref{tab:aliasfail} decomposes where the anchor-free errors go, and it
qualifies the previous paragraph in the benchmark's own disfavour. The
keying-deficit subtraction assumes the retrieval penalty of aliasing is
separable from the keying penalty, and that assumption holds for the entity
linker --- its miss rate is close to the oracle's, so nearly all of its deficit
is stale answers, i.e.\ genuine keying failure. It does \emph{not} hold for the
gazetteer, whose loss is almost entirely stale, while the oracle's is entirely
miss: those are different failures and subtracting one from the other is
informative only about magnitude. Relatedly, the entity linker's low
contamination at $a{=}1$ is not purely a win: its current@5 is below store-all's,
so part of the ``no conflict in the context'' is ``no answer in the context''.

\section{The repair is concentrated where facts change least}
\label{sec:strat}

\begin{table*}[t]
\centering
\begin{minipage}[t]{0.48\linewidth}
\centering
\caption{current@1 by per-fact version count (test, $p{=}0.5$). The cheap fix
collapses on the facts that changed most --- exactly the motivating
population.}
\label{tab:strat}
\small
\begin{tabular}{lcc}
\toprule
policy & 2 ver.\ ($n{=}423$) & $\ge$3 ver.\ ($n{=}45$) \\
\midrule
BM25 & $0.423$ & $0.178$ \\
BM25+rerank & $0.986$ & $0.467$ \\
Cosine & $0.482$ & $0.289$ \\
+ time-rerank & $0.967$ & $0.533$ \\
+ recency-wt & $0.794$ & $0.689$ \\
Reconsolid. & $0.868$ & $0.778$ \\
Lexical key & $0.974$ & $0.956$ \\
Oracle & $0.981$ & $0.956$ \\
\bottomrule
\end{tabular}

\end{minipage}\hfill
\begin{minipage}[t]{0.48\linewidth}
\centering
\caption{A late stale report: for a fraction $q$ of the \lateNEligible{} test
facts the final version arrives \emph{before} its predecessor, timestamps
intact ($p{=}0.5$; current@1, mean of \corrSeeds{} draws). Overwrite loses its
margin in proportion to $q$ with contamination@5 fixed at \lateLexContamMax{}:
one confidently stale memory, not a contradiction.}
\label{tab:latereport}
\footnotesize
\setlength{\tabcolsep}{3pt}
\begin{tabular}{lcccc}
\toprule
policy & $q{=}0$ & $q{=}0.1$ & $q{=}0.25$ & $q{=}0.5$ \\
\midrule
+ time-rerank & $0.925$ & $0.925$ & $0.925$ & $0.925$ \\
Overwrite: gaz. & $0.972$ & $0.875$ & $0.729$ & $0.486$ \\
Overwrite: oracle & $0.979$ & $0.880$ & $0.734$ & $0.491$ \\
Symmetric key & $0.989$ & $0.890$ & $0.740$ & $0.493$ \\
If newer: gaz. & $0.972$ & $0.972$ & $0.972$ & $0.972$ \\
\bottomrule
\end{tabular}

\end{minipage}
\end{table*}

\begin{table*}[t]
\centering
\caption{The two temporal signals, corrupted separately and together (test,
$p{=}0.5$; current@1, mean over \corrSeeds{} permutation seeds; hyperparameters
are the val-tuned values of the clean regime, since a deployment cannot
retune for a corruption it does not know it has). \emph{clock permuted}: timestamps permuted across
statements, arrival order intact --- unreliable logging, imports, clock skew.
\emph{arrival permuted}: arrival order permuted, each statement keeping its
own timestamp --- delayed or out-of-order ingestion. Every time-aware or retention policy reads one
signal and is blind to the other; with both corrupted nothing recovers. The
per-seed spread is at most \corrSeedSpreadMax{} points in any cell.}
\label{tab:tshuffle}
\small
\setlength{\tabcolsep}{5pt}
\begin{tabular}{lcccc}
\toprule
policy & clean & clock permuted & arrival permuted & both \\
\midrule
Cosine (store-all) & $0.464$ & $0.464$ & $0.464$ & $0.464$ \\
+ time-rerank & $0.925$ & $0.484$ & $0.925$ & $0.449$ \\
+ recency-wt & $0.784$ & $0.451$ & $0.784$ & $0.447$ \\
Overwrite: threshold & $0.859$ & $0.859$ & $0.452$ & $0.452$ \\
Overwrite: gazetteer & $0.972$ & $0.972$ & $0.472$ & $0.472$ \\
Invalidate: gazetteer & $0.972$ & $0.972$ & $0.472$ & $0.472$ \\
Overwrite: oracle & $0.979$ & $0.979$ & $0.472$ & $0.472$ \\
Read-path key (retain all) & $0.989$ & $0.491$ & $0.989$ & $0.452$ \\
Symmetric key & $0.989$ & $0.989$ & $0.477$ & $0.477$ \\
Overwrite if newer: gazetteer & $0.972$ & $0.488$ & $0.972$ & $0.450$ \\
Overwrite if newer: oracle & $0.979$ & $0.490$ & $0.979$ & $0.451$ \\
\bottomrule
\end{tabular}

\end{table*}

Table~\ref{tab:strat}: \stratNTwo{} of \nTest{} test facts have exactly two
versions; on these, time-rerank scores \stratRerankTwo{}. On the \stratNMore{}
facts with $\ge$3 versions --- the population the introduction's argument is
about, where staleness pressure is highest --- time-rerank falls to
\stratRerankMore{} while the threshold keyer (\stratReconMore) \emph{overtakes}
it and the oracle stays at \stratOracleMore{}. The mechanism is structural:
rerank's $k$ must cover all versions of the queried fact, so the optimal $k$
tracks the per-fact version count --- knowledge that is itself the same-key
problem. The headline \realRerankCur{} is a mixture dominated by the easy
\stratPctTwo\%; the benchmark reports both strata so the mixture cannot mislead.

\section{Each fix reads one temporal signal}
\label{sec:tshuffle}

Table~\ref{tab:tshuffle} corrupts the two signals a policy can use to tell a
current version from a stale one. Permuting timestamps while keeping arrival
order collapses every time-aware fix to store-all level (time-rerank
\corrCleanRerank{} $\to$ \corrClockRerank{}; recency-weighted \corrCleanRw{}
$\to$ \corrClockRw{}) while overwrite and invalidation retention are unchanged
(threshold \corrClockRecon{}, gazetteer \corrClockLex{}, oracle
\corrClockOracle{}) --- by construction, since a keyer never reads $t$. It
would be easy to stop there and conclude that retention
``does not need the clock''. It does not; it needs the other signal. Overwrite
keeps whichever version arrived last, so permuting \emph{arrival order} while
every timestamp stays intact --- delayed or out-of-order ingestion ---
collapses the same rungs to the same floor (gazetteer \corrArrivalLex{},
oracle \corrArrivalOracle{}, invalidation \corrArrivalTkg{}; store-all cosine
is \corrArrivalCosine{}) while the time-aware fixes are untouched (time-rerank
\corrArrivalRerank{}). The read-path rung, which retains everything and sorts
the resolved key's slots by $t$, is the mirror image again: unchanged under
disordered arrival (\corrArrivalRead{}) and collapsed under a permuted clock
(\corrClockRead{}). A one-line variant that overwrites only when the incoming
timestamp is newer swaps overwrite's dependency (gazetteer
\corrArrivalLexNewer{} under disordered arrival, \corrClockLexNewer{} under a
permuted clock). With both signals corrupted nothing recovers, because the
information is gone. Which rung is best is therefore a property of which
signal a deployment can trust: with a clean clock and disordered arrival, the
best policy in the table is the most contaminated one.

Full permutation is an extreme, so Table~\ref{tab:latereport} doses the
realistic form: for a fraction $q$ of the \lateNEligible{} test facts the
final version arrives \emph{before} its predecessor --- a stale report landing
late. Overwrite retention loses its margin over chance in proportion to $q$
(gazetteer \corrCleanLex{} $\to$ \lateLexTen{} / \lateLexTwentyFive{} /
\lateLexFifty{} at $q = 0.1 / 0.25 / 0.5$; oracle \lateOracleTen{} /
\lateOracleTwentyFive{} / \lateOracleFifty{}), while time-rerank
(\lateRerankFifty{} at $q{=}0.5$) and the newer-only overwrite
(\lateLexNewerFifty{}) do not move. One detail matters for
\S\ref{sec:reader}: under this corruption overwrite's contamination@5 stays at
\lateLexContamMax{}. It does not hand the reader a contradiction; it hands the
reader a single, confidently stale memory, which the constructed-context probe
scores at \doseStaleOnly{}. Invalidation protects against that only if it
invalidates by timestamp rather than by arrival. The division of labour is
measured rather than assumed, and it is symmetric: metadata heuristics repair
freshness exactly when time metadata is trustworthy; retention repairs it
exactly when ingestion is in order.

\section{Paraphrase: the same-key tax}
\label{sec:para}

\begin{table*}[t]
\centering
\caption{Surface variation stresses similarity keying: current@1
{\scriptsize[95\% CI]} and contamination@5 on the test split as the
paraphrase probability $p$ grows. The oracle's keying is unaffected by
construction; the gazetteer keyer (values in the text) is unaffected
\emph{while subjects stay verbatim}.}
\label{tab:paraphrase}
\small
\setlength{\tabcolsep}{4pt}
\begin{tabular}{lcccccc}
\toprule
& \multicolumn{2}{c}{$p=0$} & \multicolumn{2}{c}{$p=0.5$} & \multicolumn{2}{c}{$p=1$} \\
\cmidrule(lr){2-3}\cmidrule(lr){4-5}\cmidrule(lr){6-7}
policy & current@1 & cont.@5 & current@1 & cont.@5 & current@1 & cont.@5 \\
\midrule
Cosine (store-all) & $0.447$ {\scriptsize[0.402,0.491]} & $1.00$ & $0.464$ {\scriptsize[0.419,0.506]} & $0.97$ & $0.468$ {\scriptsize[0.423,0.513]} & $0.98$ \\
Cosine + time-rerank & $0.959$ {\scriptsize[0.940,0.976]} & $1.00$ & $0.925$ {\scriptsize[0.900,0.947]} & $0.97$ & $0.942$ {\scriptsize[0.921,0.964]} & $0.98$ \\
Reconsolidation & $0.887$ {\scriptsize[0.857,0.915]} & $0.09$ & $0.859$ {\scriptsize[0.827,0.889]} & $0.14$ & $0.829$ {\scriptsize[0.795,0.863]} & $0.16$ \\
Keep-latest oracle & $0.998$ {\scriptsize[0.994,1.000]} & $0.00$ & $0.979$ {\scriptsize[0.966,0.991]} & $0.00$ & $0.970$ {\scriptsize[0.953,0.985]} & $0.00$ \\
\bottomrule
\end{tabular}

\end{table*}

Table~\ref{tab:paraphrase}: as statements shift from canonical to paraphrased
templates, the threshold keyer loses \realReconCurPzero{} $\to$
\realReconCurPone{} while the oracle holds (\realOracleCurPzero{} $\to$
\realOracleCurPone), and the gap widens from \gapReconOraclePzero{} to
\gapReconOraclePone{} points. The gazetteer keyer tracks the oracle at every
$p$ (\realLexCurPzero{}, \realLexCurPhalf{}, \realLexCurPone{}) because
template-level paraphrase preserves the subject's surface form. Paraphrase
alone therefore stresses only one rung of the ladder. \S\ref{sec:anchor} removes
the anchor itself.

\section{Keying, measured without retrieval}
\label{sec:keying}

\begin{table*}[t]
\centering
\caption{Same-fact decisions scored directly as a clustering of statements
against the true fact partition, independent of any retriever. \emph{Merge
err.} counts slots holding statements about more than one fact;
\emph{\% split} is the fraction of facts scattered across several slots.}
\label{tab:keying}
\small
\setlength{\tabcolsep}{4pt}
\begin{tabular}{lcccc}
\toprule
& \multicolumn{2}{c}{anchored ($a{=}0$)} & \multicolumn{2}{c}{anchor-free ($a{=}1$)} \\
\cmidrule(lr){2-3}\cmidrule(lr){4-5}
keyer & merge err.\ & \% split & merge err.\ & \% split \\
\midrule
Threshold $\tau$ & $40$ & $22.6$ & $98$ & $55.3$ \\
Gazetteer (exact) & $2$ & $0.4$ & $3$ & $89.3$ \\
Fuzzy linker (no alias table) & $5$ & $0.4$ & $35$ & $13.2$ \\
Linker + alias table & $2$ & $0.4$ & $25$ & $6.6$ \\
True-key oracle & $0$ & $0.0$ & $0$ & $0.0$ \\
\bottomrule
\end{tabular}

\end{table*}

\begin{table*}[t]
\centering
\caption{Retention mechanism at fixed keyer ($p{=}0.5$):
invalidation matches overwrite's accuracy and contamination while keeping
every superseded statement recoverable.}
\label{tab:mechanism}
\small
\setlength{\tabcolsep}{4pt}
\begin{tabular}{lcccccc}
\toprule
& \multicolumn{3}{c}{overwrite (destructive)} & \multicolumn{3}{c}{temporal KG (invalidate)} \\
\cmidrule(lr){2-4}\cmidrule(lr){5-7}
keyer & current@1 & cont.@5 & slots & current@1 & cont.@5 & slots \\
\midrule
anchored, threshold $\tau$ & $0.859$ & $0.137$ & $513$ & $0.821$ & $0.126$ & $995$ \\
anchored, entity linker & $0.972$ & $0.000$ & $467$ & $0.972$ & $0.000$ & $995$ \\
anchored, true-key oracle & $0.979$ & $0.000$ & $468$ & $0.979$ & $0.000$ & $995$ \\
anchor-free, threshold $\tau$ & $0.513$ & $0.261$ & $561$ & $0.451$ & $0.237$ & $995$ \\
anchor-free, entity linker & $0.776$ & $0.024$ & $469$ & $0.776$ & $0.024$ & $995$ \\
anchor-free, true-key oracle & $0.827$ & $0.000$ & $468$ & $0.827$ & $0.000$ & $995$ \\
\bottomrule
\end{tabular}

\end{table*}

Retrieval metrics mix keying quality with ranking quality.
Table~\ref{tab:keying} separates them by scoring each keyer's output directly
as a clustering of statements against the true fact partition. The picture
matches \S\ref{sec:anchor} and sharpens it: withdrawing the anchor scatters
\auditLexSplitOne\% of facts across multiple slots for the gazetteer (from
\auditLexSplitZero\%) and \auditThreshSplitOne\% for the threshold keyer,
against \auditFuzzySplitOne\% for the fuzzy linker and
\auditEntitySplitOne\% for the alias-equipped one. Merge errors move
the other way --- the entity linker creates \auditEntityImpureOne{} impure
slots at $a{=}1$ against \auditEntityImpureZero{} at $a{=}0$ --- which is the
failure that destroys evidence irreversibly under overwrite. Split and merge
trade against each other, and no rung but the oracle escapes both.

Two properties of this metric should be stated so it is not over-read. A keyer
that refuses to link anything scores zero merge errors and near-total
splitting by construction, so the gazetteer's \auditLexSplitOne\% at $a{=}1$
is partly definitional --- it \emph{cannot} match an absent string. And the
two columns have different denominators (clusters and facts), so they order
keyers rather than trading off on a common scale; a full merge/split operating
curve over $\tau$ is left to the released harness.

\textbf{Mechanism.} Table~\ref{tab:mechanism} holds the keyer fixed and varies
what retention does. With a good keyer, invalidation is free in quality terms:
identical current@1 and contamination to overwrite, at \mechKgSlots{} slots
against \mechOwSlots{} because nothing is deleted. With the threshold keyer it
is slightly \emph{worse} (\mechKgTauCur{} vs \mechOwTauCur{}), since a false
merge invalidates a correct statement that then ranks below its own
replacement. The design implication for production systems is concrete:
tombstoning buys recoverability of false merges at roughly $2\times$ footprint
and no accuracy cost --- provided the keyer is good enough that invalidations
are usually right.

\section{A realistic corpus, and the cost of an incomplete inventory}
\label{sec:fullcorpus}

\begin{table*}[t]
\centering
\caption{current@1 on the \fullNEvents-statement corpus (\fullNDistract{}
distractor statements from fact keys that are never queried). Keyers hold
either an \emph{open-world} inventory (the \fullNInvOpen{} queried subjects)
or a \emph{closed-world} one (all \fullNInvClosed{} subjects in the stream).
Store-all policies use no inventory, so their columns coincide.}
\label{tab:fullcorpus}
\small
\setlength{\tabcolsep}{4pt}
\begin{tabular}{lcccc}
\toprule
& \multicolumn{2}{c}{anchored} & \multicolumn{2}{c}{anchor-free} \\
\cmidrule(lr){2-3}\cmidrule(lr){4-5}
policy & open-world & closed-world & open-world & closed-world \\
\midrule
Cosine (store-all) & $0.462$ & $0.462$ & $0.109$ & $0.109$ \\
+ time-rerank & $0.902$ & $0.902$ & $0.590$ & $0.590$ \\
Overwrite: threshold & $0.793$ & $0.793$ & $0.404$ & $0.404$ \\
Overwrite: gazetteer & $0.942$ & $0.947$ & $0.162$ & $0.165$ \\
Overwrite: fuzzy linker & $0.912$ & $0.904$ & $0.453$ & $0.436$ \\
Overwrite: linker + alias table & $0.902$ & $0.947$ & $0.585$ & $0.600$ \\
Symmetric key (read+write) & $0.947$ & $0.932$ & $0.624$ & $0.577$ \\
Overwrite: oracle & $0.966$ & $0.966$ & $0.690$ & $0.690$ \\
\bottomrule
\end{tabular}

\end{table*}

Scaling the store to \fullNEvents{} statements (Table~\ref{tab:fullcorpus})
leaves the store-all picture intact (cosine \fullCosineOpen{}, time-rerank
\fullRerankOpen{}) and makes the keying rungs harder in two ways.

First, keying is simply harder against \fullNDistract{} more neighbours: in
the anchor-free configuration the entity linker's deficit to the oracle roughly
doubles relative to the small corpus, which is why \S\ref{sec:anchor}'s
\aliasEntityDefOne{} points are the optimistic end of the range. Second,
inventory completeness becomes its own axis. An entity linker with an
\emph{open-world} inventory --- one that knows the facts it is asked about but
not the rest of the corpus --- scores \fullEntityOpen{} against
\fullEntityClosed{} closed-world, so incompleteness costs
\fullEntityInventoryCost{} points on its own, a cost any system deployed on
open-domain logs pays by default and one invisible at the scale of
\S\ref{sec:anchor}.

That second effect is modest, and it depends on the matcher. Matching
inventory forms as bare substrings inflates it several-fold, because that
links \substringFalseLink\% of the
\nDistractStatements{} distractor statements into a queried fact through forms
like ``Democratic'' or ``John''; the token-boundary mention matching used here
(\S\ref{sec:policies}) links \mentionFalseLink\%. Anything weaker measures the
matcher rather than the inventory, and it does so in the direction that
flatters this paper's thesis. Completing the inventory helps both rungs, by
\fullEntityInventoryCost{} points for the linker and
\fullLexInventoryCost{} for the gazetteer: with token-boundary matching a
larger inventory does not manufacture cross-fact merges, which is precisely
what a substring matcher does.

\section{Natural short forms}
\label{app:natural}

\begin{table*}[t]
\centering
\caption{The anchor-free ladder at $a{=}1$ with short forms drawn from
Wikidata aliases and English Wikipedia redirects instead of generated
mechanically (test, $p{=}0.5$; current@1, and the keying deficit in points
against the write-path oracle of the same configuration; every tunable
re-tuned on val, whose own inventory covers \natValCovSubjects{} of its
subjects). \emph{natural}: a subject with no real short form stays anchored
(\natCovSubjects{} of the \natNSubjects{} distinct subjects behind the \nTest{} test facts have one;
\natCovStatementsPct\% of alias-eligible statements are aliased). \emph{natural+fallback}: such a
subject takes the mechanical forms instead (\natFbCovStatementsPct\%), the
like-for-like comparison with Table~\ref{tab:alias}.}
\label{tab:natural}
\small
\setlength{\tabcolsep}{4pt}
\begin{tabular}{lcccccc}
\toprule
& \multicolumn{2}{c}{mechanical} & \multicolumn{2}{c}{natural} & \multicolumn{2}{c}{natural+fallback} \\
\cmidrule(lr){2-3}\cmidrule(lr){4-5}\cmidrule(lr){6-7}
policy & current@1 & deficit & current@1 & deficit & current@1 & deficit \\
\midrule
Overwrite: threshold & $0.513$ & $+31.4$ & $0.656$ & $+16.5$ & $0.506$ & $+22.0$ \\
Overwrite: gazetteer & $0.169$ & $+65.8$ & $0.498$ & $+32.3$ & $0.162$ & $+56.4$ \\
Overwrite: fuzzy linker & $0.714$ & $+11.3$ & $0.733$ & $+8.8$ & $0.573$ & $+15.4$ \\
Read-path key (retain all) & $0.831$ & $-0.4$ & $0.780$ & $+4.1$ & $0.679$ & $+4.7$ \\
Symmetric key & $0.831$ & $-0.4$ & $0.780$ & $+4.1$ & $0.679$ & $+4.7$ \\
Overwrite: linker + alias table & $0.776$ & $+5.1$ & $0.812$ & $+0.9$ & $0.701$ & $+2.6$ \\
Symmetric key + alias table & $0.912$ & $-8.5$ & $0.985$ & $-16.5$ & $0.959$ & $-23.3$ \\
Overwrite: oracle & $0.827$ & $+0.0$ & $0.821$ & $+0.0$ & $0.726$ & $+0.0$ \\
\bottomrule
\end{tabular}

\end{table*}

The mechanical short forms of \S\ref{sec:anchoraxis} are what make the anchor
settable, and they are also the axis's weakest point: every one is a substring
of the full name. Table~\ref{tab:natural} reruns the ladder with the short
forms people actually use --- \natNForms{} forms harvested from Wikidata
aliases and Wikipedia redirects for the test subjects (\emph{VR46}, \emph{Vale
Rossi}, misspellings and transliterations included), \natOverlapPct\% of which
the mechanical generator also produces, and \natNAmbiguousForms{} of which
name two subjects. Two things change and one does not.

Natural forms are harder for retrieval and easier for keying. A nickname
shares no token with the full-name query, so even the write-path oracle drops
(\aliasOracleCurOne{} mechanical, \natNatOracleCurOne{} natural,
\natNatFbOracleCurOne{} with fallback) and store-all cosine moves with it. But
handed the natural inventory, the entity linker's deficit shrinks from
\aliasEntityDefOne{} to \natNatEntityDefOne{} points, because natural forms
have no cross-subject collisions where the mechanical ones have
\aliasNAmbiguousForms{}. The rungs deployed systems use still pay tens of
points: under fallback the threshold keyer's deficit is \natNatFbReconDefOne{}
and the gazetteer's \natNatFbLexDefOne{} (\aliasReconDefOne{} and
\aliasLexDefOne{} mechanical); the milder \emph{natural} column is a coverage
effect, since \natUncoveredPct\% of subjects stay anchored there.

What does not survive is the headline of \S\ref{sec:anchor} in its strong
form. Ordinary resolution applied symmetrically no longer beats a complete
alias table applied asymmetrically: the symmetric fuzzy rung scores
\natNatFuzzyRWCurOne{} against the entity linker's \natNatEntityCurOne{} on
natural forms and \natNatFbFuzzyRWCurOne{} against \natNatFbEntityCurOne{}
with fallback, where the trigram linker leaves \natFbUnlinked{} statements
unlinked. The point estimates favour the table; neither difference is
significant (raw $p\,\natPRwAliasNat$ and $p\,\natPRwAliasNatFb$), so what
the natural forms establish is that the mechanical advantage is gone, not
that it has reversed. Trigram similarity cannot link \emph{VR46} to
\emph{Valentino Rossi}; an alias table can. What survives is the weaker and
more useful claim: resolving the query helps whatever
resolver is used. The symmetric rung beats its write-path-only twin on both
configurations ($p\,\natPRwFuzzyNat$ and $p\,\natPRwFuzzyNatFb$, Holm), and
the entity linker applied symmetrically reaches \natNatAliasRWCurOne{}
natural and \natNatFbAliasRWCurOne{} with fallback
($p\,\natPAliasRwAliasNat$ against its write-path-only form) --- the best
number on either configuration, \natNatAliasRWAboveOne{} points above the
write-path oracle. \S\ref{sec:anchor}'s statement is therefore scoped to the
mechanical axis, and both configurations ship. One circularity carries over
from the mechanical axis: the linker is handed the inventory the stream drew
from, so its number is the ceiling of a deployment whose alias table is
complete.

\section{Can a language model do the keying?}
\label{sec:llmkey}

\begin{table*}[t]
\centering
\caption{A language model as the same-fact keyer, against the mechanical rungs
(test, $p{=}0.5$). For each arriving statement the model is shown the top-%
\llmKeyNCand{} stored candidates by cosine and asked which, if any, describes
the same fact; it sees statement text only --- never an id, a key, or a
timestamp. \emph{\% split} is the retrieval-independent share of facts scattered
over more than one slot (\S\ref{sec:keying}).}
\label{tab:llmkey}
\small
\setlength{\tabcolsep}{4pt}
\begin{tabular}{lcccc}
\toprule
& \multicolumn{2}{c}{current@1} & \multicolumn{2}{c}{\% facts split} \\
\cmidrule(lr){2-3}\cmidrule(lr){4-5}
keyer & anchored & anchor-free & anchored & anchor-free \\
\midrule
threshold $\tau$ & $0.859$ & $0.513$ & $22.6$ & $55.3$ \\
gazetteer & $0.972$ & $0.169$ & $0.4$ & $89.3$ \\
linker + alias table & $0.972$ & $0.776$ & $0.4$ & $6.6$ \\
true-key oracle & $0.979$ & $0.827$ & $0.0$ & $0.0$ \\
LLM keyer (glm-4.6) & $0.846$ & $0.459$ & $27.6$ & $52.4$ \\
\bottomrule
\end{tabular}

\end{table*}

The keying ladder so far is mechanical, and its cheap rungs fail in the
anchor-free regime. The obvious next move --- the one production memory systems
now make, from Mem0's LLM-decided UPDATE \citep{chhikara2025mem0} to RL-learned
memory operations \citep{yan2025memoryr1} --- is to let a language model
adjudicate. This rung tests that directly (Table~\ref{tab:llmkey}), and the
answer is negative in a way worth reporting.

\textbf{The model underperforms every mechanical keyer when the anchor is
present, and lands mid-ladder when it is not.} Anchored, \llmKeyPriName{}
reaches \llmKeyPriCurZero{} against \realLexCur{} for the gazetteer and
\aliasEntityCurZero{} for the entity linker, leaving \llmKeyPriDefZero{} points
to the oracle --- worse than every mechanical rung including the embedding
threshold. Anchor-free it reaches \llmKeyPriCurOne{}, which beats the collapsed
gazetteer (\aliasLexCurOne{}) but sits below the embedding threshold
(\aliasReconCurOne{}) and far below the entity linker
(\aliasEntityCurOne{}). The clustering audit shows the failure mode: it
scatters \llmKeyPriSplitZero\% of facts across slots when anchored and
\llmKeyPriSplitOne\% when not, against \auditEntitySplitOne\% for the linker,
while creating very few impure slots (\llmKeyPriImpureOne{} at $a{=}1$). The
model is not confusing facts; it is declining to merge them.

\textbf{The comparison is not resource-matched, and the direction of the
mismatch matters.} The mechanical rungs are handed a canonical subject
inventory --- for the entity linker, an alias table as well. The model is
handed neither: it sees the arriving statement and the candidate statements as
text, and must decide same-fact from that alone. So this is not a claim that
language models are worse same-fact detectors than trigram matching under equal
information; it is a measurement of the configuration people actually deploy,
in which the LLM is asked to adjudicate without a gazetteer. An
inventory-equipped LLM keyer is an obvious and unrun experiment, and the
plausible result is that it closes some of the gap. What the measurement does
establish is that a language model asked for a one-shot same-fact decision
does not replace the inventory, which is the substitution the literature
proposes.

\textbf{This is not a retrieval failure.} The obvious objection is that the
right candidate was never shown. It usually was: for statements whose fact
already occupies a slot, that slot is among the \llmKeyNCand{} offered
candidates \llmKeyRecallZero{} of the time when anchored and
\llmKeyRecallOne{} of the time when not. The distance between that ceiling and
the model's linking rate is judgment, not recall.

\textbf{What this does and does not license.} It is one prompt, one candidate
set size, and \llmKeyNModels{} model, so it bounds LLM keying from below rather
than above --- a different prompt, more candidates, or a model fine-tuned for
the task would plausibly do better, and the released harness makes that a
one-line experiment. What it does establish is that delegating same-fact
detection to a general-purpose model is not automatically an improvement over a
gazetteer, and that it is not cheap: \llmKeyPriCalls{} model calls for
\llmKeyNEvents{} statements --- one per arriving statement, against zero for
every mechanical rung --- to land below the entity linker in both regimes. On this evidence it is entity
resolution machinery, not model scale, that closes the anchor-free gap of
\S\ref{sec:anchor}.

\ifdefined\sysNFacts \section{Where a shipped system lands on the ladder}
\label{sec:systems}

\begin{table*}[t]
\centering
\caption{Mem0 scored under the same harness as the idealized rungs, on the same
streams, with the same encoder (test subsample, $n{=}\sysNFacts{}$ facts,
$p{=}0.5$; anchored unless the column says otherwise; contamination@5 by
id). The rightmost column is the system's own decision
histogram over its \sysMemZeroAdds{} ingestion calls. The library version is
part of the result, not provenance: see the text.}
\label{tab:systems}
\small
\setlength{\tabcolsep}{4pt}
\begin{tabular}{llcccccc}
\toprule
policy / system & version & cur@1 (anch.) & cur@1 (anch.-free) & cur@5 & cont.@5 & slots & ADD\,/\,UPDATE \\
\midrule
Cosine (store-all) & \texttt{---} & $0.460$ & $0.140$ & $0.987$ & $0.973$ & $329$ & --- \\
+ time-rerank & \texttt{---} & $0.920$ & $0.787$ & $0.987$ & $0.973$ & $329$ & --- \\
Overwrite: threshold & \texttt{---} & $0.893$ & $0.633$ & $0.973$ & $0.147$ & $173$ & --- \\
Overwrite: gazetteer & \texttt{---} & $0.973$ & $0.207$ & $0.987$ & $0.000$ & $150$ & --- \\
Overwrite: fuzzy linker & \texttt{---} & $0.987$ & $0.820$ & $1.000$ & $0.000$ & $150$ & --- \\
Symmetric key (read+write) & \texttt{---} & $1.000$ & $0.920$ & $1.000$ & $0.000$ & $150$ & --- \\
\textbf{Mem0} (qwen2.5:14b) & \texttt{2.0.11} & $0.180$ & $0.120$ & $0.820$ & $0.720$ & $857$ & 857\,/\,0 \\
\textbf{Mem0} (qwen2.5:14b) & \texttt{0.1.118} & $0.913$ & $0.653$ & $0.987$ & $0.120$ & $238$ & 269\,/\,149 \\
Overwrite: true-key oracle & \texttt{---} & $0.987$ & $0.887$ & $1.000$ & $0.000$ & $150$ & --- \\
\bottomrule
\end{tabular}

\end{table*}

\begin{table*}[t]
\centering
\caption{Every \mbox{(release, vector store)} pair we probed, and whether its
numbers are usable. \emph{writes}: the system can record its own ADD/UPDATE
decisions. \emph{retrieval}: a hand-scored query returns the right fact first
from what the store holds. Two pairs fail, in opposite halves, and both produce
plausible-looking metrics that measure the backend rather than the system ---
which is why the selection of runs quoted in this section is made by this
table and not by which result file was written last.}
\label{tab:gates}
\small
\setlength{\tabcolsep}{5pt}
\begin{tabular}{llcccc}
\toprule
mem0 release & vector store & writes & retrieval & returned & usable \\
\midrule
\texttt{0.1.118} & \texttt{chroma} & \checkmark & \checkmark & $6$ & \checkmark \\
\texttt{0.1.118} & \texttt{faiss} & \checkmark & $\times$ & $2$ & $\times$ \\
\texttt{0.1.118} & \texttt{qdrant} & $\times$ & \checkmark & $6$ & $\times$ \\
\texttt{2.0.11} & \texttt{chroma} & \checkmark & \checkmark & $14$ & \checkmark \\
\texttt{2.0.11} & \texttt{faiss} & \checkmark & \checkmark & $14$ & \checkmark \\
\texttt{2.0.11} & \texttt{qdrant} & \checkmark & \checkmark & $14$ & \checkmark \\
\bottomrule
\end{tabular}

\end{table*}

\begin{table*}[t]
\centering
\caption{Each policy's own top-5 handed to \sysrdAdjReader{} (anchored,
$n{=}\sysNFacts{}$; rows carry the harness's policy ids --- \texttt{reconsolidation} is the
threshold keyer, \texttt{lexical-key} the gazetteer, \texttt{fuzzy-key} the
fuzzy linker, \texttt{fuzzy-rw-key} symmetric keying; the system's row is its own rewritten memories in its own
words, from the \probedAdjVersion{} run --- \probedCurVersion{} is in the
text). \emph{as retrieved} is what a deployment gets; the next two pairs apply
this paper's read-path fix post hoc to those same contexts, resolving the query
through the keyer of \S\ref{sec:anchor} and then keeping one memory per
resolved entity, at no ingestion and no model call. The two middle \emph{cur@1}
columns are identical wherever the readers beside them differ by up to
\sysrdCosineRagMergedZero{} against \sysrdCosineRagKeyedZero{}: the retrieval
metric cannot see the difference the reader is made of.
The last column keeps the \emph{oldest} memory per entity instead of the
freshest. Because facts arrive as they become true, freshest-is-current is free
here (\sysRecencyValid{}/\sysRecencyTotal{} facts); withdrawing it leaves only
what each write path had already removed, so the column reads near zero for
anything that stores every version and near one for anything that merges on
write. It is a write-path purity measure that uses no identifiers and no
timestamps in its favour, and it is where the deployed system is priced.}
\label{tab:sysreader}
\small
\setlength{\tabcolsep}{4pt}
\begin{tabular}{lccccccc}
\toprule
context source & \multicolumn{2}{c}{as retrieved} & \multicolumn{2}{c}{$+$ query resolved} & \multicolumn{2}{c}{$+$ merged, freshest} & oldest \\
 & cur@1 & reader & cur@1 & reader & cur@1 & reader & reader \\
\midrule
\texttt{cosine-rag} & $0.460$ & $0.573$ & $1.000$ & $0.293$ & $1.000$ & $1.000$ & $0.007$ \\
\texttt{time-rerank} & $0.920$ & $0.300$ & $1.000$ & $0.293$ & $1.000$ & $1.000$ & $0.007$ \\
\texttt{reconsolidation} & $0.893$ & $0.920$ & $0.973$ & $0.893$ & $0.973$ & $0.973$ & $0.820$ \\
\texttt{lexical-key} & $0.973$ & $0.987$ & $0.987$ & $0.987$ & $0.987$ & $0.987$ & $0.987$ \\
\texttt{fuzzy-key} & $0.987$ & $1.000$ & $1.000$ & $1.000$ & $1.000$ & $1.000$ & $1.000$ \\
\texttt{fuzzy-rw-key} & $1.000$ & $1.000$ & $1.000$ & $1.000$ & $1.000$ & $1.000$ & $1.000$ \\
\texttt{keep-latest-oracle} & $0.987$ & $1.000$ & $1.000$ & $1.000$ & $1.000$ & $1.000$ & $1.000$ \\
\texttt{Mem0 0.1.118} & $0.913$ & $0.720$ & $0.973$ & $0.573$ & $0.973$ & $0.900$ & $0.353$ \\
\bottomrule
\end{tabular}

\end{table*}

Everything above prices same-fact detection with rungs we built. The obvious
question is where the systems people deploy actually sit, and
\S\ref{sec:related} names them --- Mem0's UPDATE operation, Zep's edge invalidation,
MemGPT's self-editing memory --- as evidence that the field already ships the
fix. We ran Mem0 first, and the first thing we found was that the answer
depends on which release you install.

\textbf{The adjudication is not on the default ingestion path in the current
release.} Mem0's documented loop is exactly the decision this paper prices: an
LLM sees an incoming statement with the nearest stored memories and returns
ADD, UPDATE, DELETE or NOOP. In the \texttt{0.1.x} line that is what
\texttt{add()} does --- it calls \texttt{get\_update\_memory\_messages} and can
emit any of the four. In the \texttt{2.0.x} line \texttt{add()} instead runs a
phased extraction whose system prompt states that its sole operation is ADD,
and the update-adjudication prompts have no call site anywhere under
\texttt{mem0/memory/}; the ADD/UPDATE decision is reachable only through the
separate \texttt{update(memory\_id, \ldots)} API, which an ingestion pipeline
does not call. So on \texttt{2.0.11} an UPDATE count of zero is fixed before
any data is streamed, and reporting it as a property of our data would be a
misattribution. We report it as what it is: a property of the release.

That distinction is easy to miss and expensive to get wrong, so the harness
records \texttt{lib\_version} and an \texttt{adjudicates\_on\_add} flag derived
by inspecting the installed package, and we score both lines.

The consequence is visible in \miniNStatements{} statements. Stream ``Peter
Sagan plays for Bora-Hansgrohe.'' then ``Peter Sagan plays for
TotalEnergies.'', ask which team he plays for now, and hold everything else
fixed --- same wrapper, same model, same encoder. \miniAdjVersion{} returns
\miniAdjEvents{}, keeps \miniAdjSlots{} memory, and ranks
\miniNonRankCurrent{} nowhere: rank 1 is \miniAdjRankCurrent{}.
\miniNonVersion{} returns \miniNonEvents{}, keeps \miniNonSlots{}, and ranks
\miniNonRankCurrent{} first. That is the whole finding, before any benchmark is
involved, and it is why the rest of this section reports a version rather
than a system.

\textbf{On \probedCurVersion{}, the system is a store-all policy.} Its
\sysMemZeroAdds{} ingestion calls over \sysNStatements{} statements produce
\sysMemZeroAddEvents{} extraction results --- more results than statements,
because extraction splits a statement into several memories --- and every one of
them is an ADD (\sysMemZeroUpdateEvents{} UPDATE). The store ends at
\probedCurSlotsZero{} slots, and retrieval from it reaches current@1
\probedCurCurZero{} at contamination@5 \probedCurContamZero{}: below plain cosine
(\probedAdjRefCosine{}) over the identical statements with the identical
encoder. It is not failing to retrieve the fact --- current@5 is
\sysMemZeroCurFive{} --- but failing to rank it first, because extraction
rewrites each statement into derived memories and what the retriever scores is
no longer the sentence whose currency the source defined. A system can lose
freshness by paraphrasing its own inputs.

\textbf{On \probedAdjVersion{}, where the adjudication is on the
\texttt{add()} path, it fires.} The same streams, the same model, the same
encoder, the same harness: \probedAdjUpdate{} UPDATE and \probedAdjDelete{}
DELETE decisions alongside \probedAdjAdd{} ADD, and a store that ends at
\probedAdjSlotsZero{} slots rather than \probedCurSlotsZero{}. Withdraw the
lexical anchor and it makes \probedAdjUpdateAF{} UPDATE decisions instead of
\probedAdjUpdate{}: a production adjudicator is not exempt from the axis
\S\ref{sec:anchor} varies, and its store grows accordingly. Those counts come
from the system's own report and are the least ambiguous thing in this
section. We repeated the whole cell --- same release, same backend, same
model, fresh store --- and it reproduced \repeatAdjEventsIdentical{}ly: the
event histogram, the \repeatAdjSlotsA{} slots and current@1
\repeatAdjCurA{}/\repeatAdjCurB{} all landed on the same values, a difference
of \repeatAdjCurDelta{} points.

\textbf{How well it works, though, depends entirely on how you score it --- and
that turned out to be the finding.} Our convention attributes a returned memory
to the statement that most recently wrote it. On \probedAdjVersion{} that
convention reports current@1 \probedAdjCurZero{} at contamination@5
\probedAdjContamZero{}, which reads as a retention policy whose keying works.
Scoring the same contexts by \emph{what they say} --- does the surfaced memory
name the current value of the queried fact, or a superseded one --- reports
\textAdjCurZero{} at \textAdjContamZero{}. The two disagree by
\textAdjCurDropZero{} points on rank~1 and \textAdjContamRiseZero{} on
contamination, in the direction that flatters the system.

The control says the disagreement is not the scorer. On the same stream, the
\attrAdjNControlsZero{} idealized rungs --- which return the corpus verbatim
and so must agree --- differ by at most \attrAdjControlMaxGapZero{} points
between the two scorings, with \attrAdjControlCollapsedZero{} ambiguous
attributions between them. The deployed system differs by
\textAdjCurDropZero{}.

\textbf{The mechanism is not the merge we expected, and it runs in both
directions.} The natural guess is self-contradicting merges --- one memory naming
both values --- but no memory does: hedged@1 by text is
\textAdjHedgedZero{}. What happens on \probedAdjVersion{} is that an UPDATE
relabels a memory without rewriting it, so a current version's id comes to sit
on superseded text. \attrAdjCollapsedZero{} of \attrAdjNQueriesZero{} queries
surface two distinct memories collapsed onto one benchmark id, and
\attrAdjUpdatedStaleZero{} surface a memory the system marked as updated whose
text still reads as the superseded value. Rank~1 and contamination@5 both call
that clean, because both identify a version by an id and the ids now agree.

On \probedCurVersion{} the same convention fails the other way, and harder.
\attrCurIdWrongZero{} of \attrCurIdScoredZero{} surfaced memories ---
\attrCurIdWrongPctZero\% --- carry an id naming a different subject than their
own text, against \attrAdjIdWrongPctZero\% on \probedAdjVersion{}. The
consequence is stark: the id-level miss rate is \probedCurMissZero{} while the
text-level miss rate is \textCurMissZero{}. The system returns a memory about
the queried fact essentially every time, and our wrapper cannot say which fact
it is. Every id-level number for that release is a statement about our
bookkeeping. We do not claim a mechanism: a six-statement probe of the same
release attached metadata correctly, so whatever produces this does not appear
at that size, and what we report is measured over the run's own memories, where
subject-named-in-text against subject-of-id needs no model call to check.

\textbf{So the id-level headline does not survive.} By id,
current@1 between the two releases reads \probedAdjCurZero{} $\to$
\probedCurCurZero{}. By text it reads \textAdjCurZero{} $\to$
\textCurCurZero{}. A \probedGapPts-point collapse becomes a single-digit one,
and the gap was manufactured by a convention that is too kind to one release
and broken on the other. This is the instrument catching a claim its own
id-level numbers invite, the case \S\ref{sec:reader} makes in general.\ifdefined\adjTpGlmCurZero{} The size of the disagreement is a
property of the adjudicator model as well as of the convention: with glm-4.6
in the loop the two scorings agree (below, Table~\ref{tab:adjudicator}).\fi{}

\textbf{We verified that this is the system and not our backend}, because
that distinction decided the result in both directions. Every
\mbox{(release, vector store)} pair is hand-probed before its numbers are used:
one query, the embedder scored by hand, checked against what the store reports
holding and against whether the system can record its own decisions at all. Of
the pairs we tried, \sysNGatesPassed{} of \sysNGatesProbed{} pass
(Table~\ref{tab:gates}); \sysGateFailNote{}. \sysGateWriteCaveat{}. Both release
lines above are quoted from \sysMatchedStore{}, so the comparison is
same-backend and not an inference across backends.%
\ifdefined\probedCurSpread{} The measurement is also stable across backends:
\probedCurVersion{} spans \probedCurLo{}--\probedCurHi{} on current@1
over \probedCurNBackends{} of the stores that pass, a range of
\probedCurSpread{} points --- far smaller than any difference this section
turns on.\fi{}

\textbf{This does not overturn \S\ref{sec:llmkey}, and the difference is
instructive.} That section found a language model asked to adjudicate same-fact
in one shot to be worse than every mechanical rung; this one finds a
purpose-built adjudication loop, running a \emph{smaller} model, doing rather
well. Four things differ and we can separate none of them here: the loop (a
system that maintains its own candidate set, retries and can DELETE, against a
single classification call), the model, the query set
($\sysNFacts$ facts against \nTest{}), and the attribution convention above.
The reading we can support is the one \S\ref{sec:llmkey} already states --- that it
bounds LLM keying from below rather than above --- and \S\ref{sec:systems} is
the empirical witness that the bound is loose. What survives both sections is
narrower: a language model asked for a one-shot same-fact decision does not
replace an entity inventory, and anchored, the cheapest mechanical rung that
has one still wins.

Every comparison in this section is paired by query on a single stream, so
the sign-flip test used throughout \S\ref{sec:anchor} and \S\ref{sec:reader}
applies unchanged; retrieval-side comparisons are Holm-corrected over one
family of \sysTestFamily{} covering both scorings, of which
\sysTestSurviving{} survive at $\alpha=0.05$, on $n=\sysTestN{}$ queries.

\textbf{Neither scoring is the end-task number, so we asked the reader.}
Table~\ref{tab:sysreader} hands each policy's own top-5 --- for the system, its
own rewritten memories in its own words --- to \sysrdAdjReader{}, and scores
the answer against the source-defined current value. Dates are never surfaced
in this track: the system returns no validity timestamp, only the store's own
created and updated times, and after an UPDATE a memory keeps its original
writer's metadata --- so the with-dates arm that closes the controlled pair in
\S\ref{sec:reader} is not available to a deployment reading this system's
output.
On \probedAdjVersion{} anchored, the id metric says \probedAdjCurZero{}, the
text metric says \textAdjCurZero{}, and the reader answers
\sysrdAdjSysAsIsZero{} correctly: above rank-1 text accuracy because the
current value is usually somewhere in the five (\textAdjCurKZero{}), below the
id metric because it is often not first. Both retrieval conventions are
proxies, and they err in opposite directions.

The releases separate far less at the reader than at any retrieval metric.
As deployed, \probedAdjVersion{} answers \sysrdAdjSysAsIsZero{} against
\probedCurVersion{}'s \sysrdCurSysAsIsZero{} anchored --- and anchor-free the
order \emph{reverses}, \sysrdAdjSysAsIsOne{} against
\sysrdCurSysAsIsOne{}. The release without the adjudication is not the disaster
the id metric reported.

\textbf{And our own fix, applied to a shipped system, is worth more than the
adjudication is.} \S\ref{sec:anchor} argues that resolving the query through
the same keyer as the stream costs nothing and buys double digits. Nobody had
applied it to a deployed system, so we did --- post hoc, over the contexts the
system already produced, with no change to its write path, no second ingestion
and no model call. Resolving the query and then merging same-entity memories,
freshest kept, lifts \probedAdjVersion{} from \sysrdAdjSysAsIsZero{} to
\sysrdAdjSysMergedZero{} anchored ($+\sysrdAdjSysGainZero{}$ points,
$p=\sysrdPAdjMergedZero{}$) and from \sysrdAdjSysAsIsOne{} to
\sysrdAdjSysMergedOne{} anchor-free ($+\sysrdAdjSysGainOne{}$). On
\probedCurVersion{} --- the store-all release --- it reaches
\sysrdCurSysMergedZero{}, above what the superseding release achieves unaided,
because a store that keeps everything still holds the current value and only
needs to be read correctly.

\emph{That last number is not an accuracy claim, and this benchmark is why.}
Facts here arrive as they become true, so the final version of a fact is a
current one for \sysRecencyValid{} of \sysRecencyTotal{} facts in this
subsample: on the clean condition a recency tie-break is correct for free, the
metric ceiling is $1.000$, and the keep-latest oracle of \S\ref{sec:anchor}
scores \probedAdjRefOracle{} on the same stream. The merged arm is fuzzy keying
plus keep-latest applied to a system's output, so it inherits that gift and
lands where the oracle already sits. How much survives corrupted timestamps is
\S\ref{sec:tshuffle}'s regime, which we do not run this arm through.

\textbf{Running the merge backwards prices the write path, and that is the
number to keep.} Hold the entity resolution fixed and keep the \emph{oldest}
memory per entity instead of the freshest. Everything recency was giving away
is now withdrawn, and what remains is whatever the system's own write path
already did. Plain cosine collapses to \sysrdCosineRagOldestZero{}, as it must:
it stores every version, so the oldest is the first. \probedCurVersion{}
collapses with it, to \sysrdCurSysOldestZero{} --- indistinguishable from
store-all, which is the reader-side confirmation that its ingestion path
performs no supersession at all, arrived at without reference to any
identifier. \probedAdjVersion{} does not collapse: it holds
\sysrdAdjSysOldestZero{} anchored and \sysrdAdjSysOldestOne{} anchor-free,
because where its adjudicator has already merged a fact to one memory the
oldest and the freshest are the same memory and it is the right one. That gap
--- \sysrdAdjSysOldestZero{} against \sysrdCurSysOldestZero{} --- is what the
ADD/UPDATE loop is worth once the read path is held identical and the free
recency signal is taken away. It is the cleanest positive measurement of a
deployed supersession mechanism in this paper, it is much smaller than the
id-level metric advertised, and it degrades on the paper's own axis when the
anchor is withdrawn.

That fix also decomposes into two halves a retrieval metric cannot tell apart,
which is the cleanest statement of this paper's thesis. Resolving the
query alone raises the system's current@1 to \sysrdAdjSysKeyedRetrZero{} and
\emph{drops} the reader to \sysrdAdjSysKeyedZero{}
($p=\sysrdPAdjKeyedZero{}$), because promoting every version of the queried
fact fills the five slots with the fact contradicting itself. Adding the merge
holds current@1 at \sysrdAdjSysMergedRetrZero{} and takes the reader to
\sysrdAdjSysMergedZero{}. The store-all control is starker still: cosine's
contexts score current@1 \sysrdCosineRagKeyedRetrZero{} under both arms and
read \sysrdCosineRagKeyedZero{} against \sysrdCosineRagMergedZero{}. Timestamp
reranking makes the same point unaided --- current@1
\sysrdTimeRerankAsIsRetrZero{}, reader \sysrdTimeRerankAsIsZero{}. Reader-side
comparisons are Holm-corrected over a family of \sysrdAdjFamily{}, of which
\sysrdAdjSurviving{} survive.

\textbf{The ladder still beats it, on the same stream, with no model at all.}
Table~\ref{tab:systems} scores our own rungs on the identical
\sysNFacts-fact stream, which matters because this subsample is easier for
them than the full test split and a cross-table comparison would understate
the gap. Symmetric keying --- a canonical subject list, character-trigram
matching, no alias table, no learning, no LLM call --- reaches
\ladderSymCurZero{} anchored and \ladderSymCurOne{} anchor-free at
contamination@5 \ladderSymContamZero{}/\ladderSymContamOne{} in
\ladderSymSlotsZero{} slots. Scored by text, the system reaches
\textAdjCurZero{}/\textAdjCurOne{} at
\textAdjContamZero{}/\textAdjContamOne{} in
\probedAdjSlotsZero{}/\probedAdjSlotsOne{} slots, for
\sysIngestHoursTotal{} hours of local LLM inference across the two quoted runs
(one per release, two configurations each). The expensive general mechanism is beaten by the cheap specific
one, and the same cheap mechanism bolted onto the expensive one afterwards is
what rescues it.

\textbf{So the mechanism the literature cites left the default path, and it
matters less than the id-level numbers say.} Between two releases of one
library, with no change to the documented interface, the adjudication moved off
\texttt{add()} and extraction began splitting one statement into several
memories. What that costs depends on where you measure. The store grows
\probedAdjSlotsZero{} $\to$ \probedCurSlotsZero{} slots and contamination by
text rises \textAdjContamZero{} $\to$ \textCurContamZero{}: those are
consequences of storing everything and they are not in dispute. Rank-1
accuracy is a different story --- \probedAdjCurZero{} $\to$
\probedCurCurZero{} by id, \textAdjCurZero{} $\to$ \textCurCurZero{} by text,
and \sysrdAdjSysAsIsZero{} $\to$ \sysrdCurSysAsIsZero{} at the reader, with
the order reversing anchor-free. The first of those three is the number a
leaderboard would report, and it is the least reliable.

That is the argument. A benchmark that
scores rewriting memory systems by attributing returned memories to source
records will misrank them, and the failure is not conservative: it flattered
one release, broke on another, and inflated a nine-point difference into a
\probedGapPts-point one. Nothing in the id-level numbers announced this. It
took scoring the same contexts by their content, and then putting a reader at
the end, to see it --- which is the case this paper makes for both, now made
against itself.

The finding is not that Mem0 is poorly built; it is a widely used system and
the \texttt{0.1.x} behaviour is the one its documentation describes. Nor is it
that the release without adjudication is fine: it holds
\probedCurSlotsZero{} memories for \sysNStatements{} statements and hands a
reader a context naming both a current and a superseded value
\textCurContamZero{} of the time. It is that the mechanism the literature
cites as the answer to stale retrieval is absent from the path a practitioner
installs, that its measured benefit where it does run is far smaller than a
retrieval metric suggests, and that improvements of \sysrdAdjSysGainZero{}
and \sysrdCurSysGainZero{} points were available to the two releases from a
wrapper with no model in it.

\ifdefined\gphCurZero
\begin{table*}[t]
\centering
\caption{The second deployed system. Graphiti (graphiti-core \gphLib{},
FalkorDB, qwen2.5:14b adjudicator, MiniLM encoder) on the identical
\sysNFacts{}-fact streams, scored by id, by text and at the reader
(\sysrdAdjReader{}, own top-5), by three read paths over the same stored
graph: the library's default search (\emph{as returned}), the same hits with
non-invalidated edges ranked first (the temporal-KG rung's read path), and
invalidated edges dropped. Ladder rows are the same stream; contamination@5
is by id here and by text in Table~\ref{tab:deployed}. An empty reader
cell means that arm was not run.}
\label{tab:graphiti}
\footnotesize
\setlength{\tabcolsep}{4pt}
\begin{tabular}{lcccccccc}
\toprule
& \multicolumn{4}{c}{anchored} & \multicolumn{4}{c}{anchor-free} \\
\cmidrule(lr){2-5}\cmidrule(lr){6-9}
policy & cur@1 & by text & cont.@5 & reader & cur@1 & by text & cont.@5 & reader \\
\midrule
Graphiti, as returned & $0.347$ & $0.347$ & $0.940$ & $0.593$ & $0.180$ & $0.180$ & $0.820$ & $0.433$ \\
Graphiti, valid edges first & $0.833$ & $0.833$ & $0.100$ & $0.893$ & $0.640$ & $0.640$ & $0.133$ & $0.773$ \\
Graphiti, valid edges only & $0.833$ & $0.833$ & $0.100$ & --- & $0.640$ & $0.640$ & $0.133$ & --- \\
Cosine (store-all) & $0.460$ & $0.460$ & $0.973$ & $0.573$ & $0.140$ & $0.140$ & $0.927$ & $0.427$ \\
+ time-rerank & $0.920$ & $0.920$ & $0.973$ & $0.300$ & $0.787$ & $0.787$ & $0.927$ & $0.220$ \\
Symmetric key & $1.000$ & $1.000$ & $0.000$ & $1.000$ & $0.920$ & $0.920$ & $0.027$ & $0.920$ \\
Overwrite: oracle & $0.987$ & $0.987$ & $0.000$ & $1.000$ & $0.887$ & $0.887$ & $0.000$ & $0.953$ \\
\bottomrule
\end{tabular}

\end{table*}

\textbf{A second system, and the other mechanism.} Graphiti
\citep{rasmussen2025zep}, the open-source engine behind Zep, is the
invalidation design our temporal-KG rung idealizes: a model extracts entities
and relation edges from each statement, and an edge a later statement
contradicts is marked \texttt{invalid\_at} rather than deleted. We run
graphiti-core \gphLib{} on FalkorDB with the same adjudicator model as Mem0
(qwen2.5:14b, through the library's JSON-mode client) and the same MiniLM
encoder as the dense baselines, on the identical \sysNFacts{}-fact streams,
under the same probe gate. On the two-statement probe invalidation fires and
the current version ranks first. On a six-statement probe every write is
stored and invalidation fires, and the current version is \emph{not} first as
returned: the library's default search returns invalidated edges beside valid
ones and breaks ties among versions of one fact by insertion order. The
library does expose a validity filter on search; the default search does not
apply it, and a deployment that calls \texttt{search()} as documented gets
the unfiltered ranking. The probe already contains the finding;
Table~\ref{tab:graphiti} measures it.

The write path works. Over the \sysNStatements{} anchored statements the
system makes \gphInvalidationsZero{} invalidations against
\gphNSupersessionsZero{} supersessions in the stream and ends with
\gphActiveZero{} active edges of \gphSlotsZero{} (\gphNoEdgeZero{} statements
yield no edge), at \gphCallsPerStmtZero{} model calls per statement and
\gphIngestHoursZero{} hours of local inference: the mechanism Mem0's current
release has lost is present here, on the default ingestion path. The read
path ignores it. As returned, the system reaches current@1 \gphCurZero{} by
id and \gphCurTextZero{} by text --- the two agree, because Graphiti's edge
text is a faithful rewrite of the statement --- at contamination@5
\gphContamZero{}, below plain cosine over the same statements
(\probedAdjRefCosine{}), and it ranks an invalidated edge above a valid one
for \gphInvalidFirstZero{} of \gphNQueries{} queries. Ranking the system's
own non-invalidated edges first --- the read path of the temporal-KG rung,
applied post hoc to what the system returned, with no model call --- lifts it
to \gphVFCurZero{} at contamination@5 \gphVFContamZero{}
($p\,\gphPVFCurZero$): most of the gap to the write-path rungs of
Table~\ref{tab:systems} on current@1, and nearly all of it on contamination.\ifdefined\gphCurOne{} Anchor-free the pattern holds and the axis reaches
the system's own entity resolution: \gphCurOne{} as returned against
\gphVFCurOne{} valid-first, \gphInvalidationsOne{} invalidations against
\gphNSupersessionsOne{} supersessions, and \gphNodesOne{} entity nodes
created for the same \sysNFacts{} subjects against \gphNodesZero{} anchored,
because a short form the extractor fails to link becomes a new entity and
\gphNoEdgeOne{} statements yield no edge at all (\gphNoEdgeZero{}
anchored).\fi\ifdefined\gphRdAsIsZero{} At the reader, \sysrdAdjReader{}
answers \gphRdAsIsZero{} from the system's own top-5 and \gphRdVFZero{} from
the valid-first top-5 (\gphRdAsIsOne{} and \gphRdVFOne{} anchor-free), against
\sysrdAdjSysAsIsZero{} for Mem0's adjudicating release as deployed; the
paper's read-path fix, resolving the query and merging same-entity memories,
takes it to \gphRdMergedZero{}. The oldest-per-entity control that priced
Mem0's write path reads \gphRdOldestZero{} here, as it must: invalidation
deletes nothing, so withdrawing the recency signal leaves the oldest version
of every fact, and what the mechanism is worth is read on the valid-first arm
instead.\fi{}

The finding is the mirror image of Mem0's. There the adjudication had left
the ingestion path of the current release; here it is present, fires, and is
discarded by the default read path, so the system scored as installed sits
below store-all and the system scored with its own verdicts respected closes
most of the gap to the retention rungs on current@1 and nearly all of it on
contamination. Both are the same lesson for a leaderboard: the
number depends on which code path the practitioner gets, and a benchmark must
say which one it scored.
\fi{}

\ifdefined\adjDefaultGlmCurZero
\begin{table*}[t]
\centering
\caption{A second adjudicator model. Mem0 \probedAdjVersion{} (chroma) under
its stock, user-centric extraction prompt and under a prompt that admits
facts about any named entity, each driven by qwen2.5:14b and by glm-4.6;
Graphiti with both models, as returned and with valid edges first. Same
\sysNFacts{}-fact streams as Table~\ref{tab:systems}; contamination@5 by
id. \emph{slots} for
Graphiti is active/total edges; the last column counts UPDATE+DELETE
decisions (Mem0) or invalidations (Graphiti).}
\label{tab:adjudicator}
\footnotesize
\setlength{\tabcolsep}{3pt}
\begin{tabular}{lcccccccccc}
\toprule
& \multicolumn{5}{c}{anchored} & \multicolumn{5}{c}{anchor-free} \\
\cmidrule(lr){2-6}\cmidrule(lr){7-11}
system, model, prompt & cur@1 & text & cont.@5 & slots & ops & cur@1 & text & cont.@5 & slots & ops \\
\midrule
Mem0, qwen, default & $0.913$ & $0.573$ & $0.120$ & $238$ & 149+31 & $0.653$ & $0.207$ & $0.320$ & $286$ & 114+21 \\
Mem0, glm, default & $0.067$ & $0.060$ & $0.000$ & $20$ & 2+1 & $0.153$ & $0.173$ & $0.007$ & $36$ & 7+1 \\
Mem0, qwen, third-person & $0.960$ & $0.587$ & $0.100$ & $251$ & 155+23 & $0.627$ & $0.200$ & $0.393$ & $303$ & 110+12 \\
Mem0, glm, third-person & $0.793$ & $0.793$ & $0.007$ & $138$ & 142+19 & $0.627$ & $0.627$ & $0.080$ & $150$ & 113+23 \\
Graphiti, qwen, as returned & $0.347$ & $0.347$ & $0.940$ & $164/321$ & 157 & $0.180$ & $0.180$ & $0.820$ & $169/306$ & 137 \\
Graphiti, qwen, valid first & $0.833$ & $0.833$ & $0.100$ & $164/321$ & 157 & $0.640$ & $0.640$ & $0.133$ & $169/306$ & 137 \\
Graphiti, glm, as returned & $0.347$ & $0.353$ & $0.953$ & $158/320$ & 162 & --- & --- & --- & --- & --- \\
Graphiti, glm, valid first & $0.860$ & $0.867$ & $0.073$ & $158/320$ & 162 & --- & --- & --- & --- & --- \\
\bottomrule
\end{tabular}

\end{table*}

\textbf{A second adjudicator model.} The runs above share one model, so
``this system does not supersede'' and ``the model we gave it does not'' were
not separated. Table~\ref{tab:adjudicator} separates them, and the first
thing it separates is a stage before the adjudication. Mem0's stock
extraction prompt asks for facts about \emph{the user}; glm-4.6 reads it
literally and returns no facts for \adjDefaultGlmSilentZero{} of
\sysNStatements{} third-person statements, so the store ends at
\adjDefaultGlmSlotsZero{} slots with \adjDefaultGlmUpdatesZero{} UPDATE
decisions and current@1 \adjDefaultGlmCurZero{}: as installed, the stronger
model empties the system. qwen2.5:14b extracts from the same prompt
(\sysMemZeroAdds{} statements, \adjDefaultQwenSilentZero{} silent). The
comparison the question needs therefore has to hold the extraction fixed. A
replacement prompt that admits facts about any named entity --- a documented
configuration option, not a change to the library --- leaves qwen2.5:14b
where it was (\adjTpQwenCurZero{} by id and \adjTpQwenCurTextZero{} by text
against \probedAdjCurZero{} and \textAdjCurZero{}; \adjTpQwenUpdatesZero{}
UPDATE decisions against \probedAdjUpdate{})\ifdefined\adjTpGlmCurZero{},
and under it glm-4.6 makes \adjTpGlmUpdatesZero{} UPDATE and
\adjTpGlmDeletesZero{} DELETE decisions, ends at \adjTpGlmSlotsZero{} slots
--- nearly one per fact --- and scores \adjTpGlmCurZero{} by id and
\adjTpGlmCurTextZero{} by text at contamination@5 \adjTpGlmContamZero{} by id
and \adjTpGlmContamTextZero{} by text (anchor-free \adjTpGlmCurOne{} /
\adjTpGlmCurTextOne{}). Two things follow. The model matters where the
adjudication runs: by text, the current value ranks first for
\adjTpGlmCurTextZero{} of queries against \adjTpQwenCurTextZero{} with the
smaller model, and the context contradicts itself \adjTpGlmContamTextZero{}
of the time against \adjTpQwenContamTextZero{}. And the attribution
disagreement this appendix made a finding of is itself model-dependent: the
larger model's UPDATE rewrites the memory's text to the current value, so id
and text agree, where the smaller model's UPDATE relabels a memory it does
not rewrite\ifdefined\adjTpGlmRdAsIsZero{}. At the reader the two arms
read \adjTpGlmRdAsIsZero{} (glm-4.6) against \adjTpQwenRdAsIsZero{}
(qwen2.5:14b) from their own top-5, and \adjTpGlmRdMergedZero{} against
\adjTpQwenRdMergedZero{} once the query is resolved and same-entity memories
merged: the paper's read-path fix recovers what the smaller adjudicator left
undone and finds nothing left to recover on the larger one\fi\fi{}.\ifdefined\gphGlmCurZero{} Graphiti's extraction
prompts are entity-centric, so the model swap reaches its adjudication
directly: with glm-4.6 it makes \gphGlmInvalidationsZero{} invalidations
against \gphInvalidationsZero{} with qwen2.5:14b, ends with
\gphGlmActiveZero{} active edges of \gphGlmSlotsZero{}, and scores
\gphGlmCurZero{} as returned and \gphGlmVFCurZero{} with valid edges first
(\gphCurZero{} and \gphVFCurZero{} with qwen2.5:14b), at
\gphGlmCallsPerStmtZero{} model calls per statement and \gphGlmHoursZero{}
hours of hosted inference --- the model changes almost nothing, because the
read path that discards the invalidations is the same. The anchor-free
configuration is not reported for this model: the library issues its
extraction calls concurrently, the hosted endpoint admits one request in
flight, and the run dropped statements on rate limits, which is a property
of our account rather than of the system.\fi{} What the swap establishes is narrower than a
verdict on either model: the number a deployment gets moves with the model
at the extraction stage before it moves at the adjudication, and a
leaderboard must pin the model as it pins the version.
\fi{}

\ifdefined\lgmQwenCurZero
\begin{table*}[t]
\centering
\caption{The third deployed system. LangMem \lgmQwenLib{} (LangChain's
long-term memory: one structured-output call per statement returns the
memories to insert or patch; inserts and updates on, deletes off, as
installed) on the identical \sysNFacts{}-fact streams with the benchmark's
MiniLM encoder as its store index, scored by id, by text and at the reader
(\sysrdAdjReader{}, own top-5). Ladder rows are the same stream.
contamination@5 is by text, since the library rewrites what it stores and
adds memories no statement wrote.}
\label{tab:langmem}
\footnotesize
\setlength{\tabcolsep}{4pt}
\begin{tabular}{lcccccccc}
\toprule
& \multicolumn{4}{c}{anchored} & \multicolumn{4}{c}{anchor-free} \\
\cmidrule(lr){2-5}\cmidrule(lr){6-9}
policy & cur@1 & by text & cont.@5 (text) & reader & cur@1 & by text & cont.@5 (text) & reader \\
\midrule
LangMem, qwen2.5:14b & $0.773$ & $0.773$ & $0.173$ & $0.833$ & $0.340$ & $0.413$ & $0.213$ & $0.600$ \\
LangMem, glm-4.6 & $0.800$ & $0.147$ & $0.753$ & $0.613$ & $0.627$ & $0.160$ & $0.640$ & $0.413$ \\
Cosine (store-all) & $0.460$ & $0.460$ & $0.973$ & $0.573$ & $0.140$ & $0.140$ & $0.927$ & $0.427$ \\
+ time-rerank & $0.920$ & $0.920$ & $0.973$ & $0.300$ & $0.787$ & $0.787$ & $0.927$ & $0.220$ \\
Symmetric key & $1.000$ & $1.000$ & $0.007$ & $1.000$ & $0.920$ & $0.920$ & $0.027$ & $0.920$ \\
Overwrite: oracle & $0.987$ & $0.987$ & $0.007$ & $1.000$ & $0.887$ & $0.887$ & $0.000$ & $0.953$ \\
\bottomrule
\end{tabular}

\end{table*}

\textbf{A third system, from the same mechanism family as Mem0.} LangMem
\citep{langmem2025} is LangChain's long-term memory: the model reads the
incoming statement beside the stored memories that resemble it and returns,
in one structured-output call, the memories to insert or patch --- Mem0's
ADD/UPDATE loop from a different vendor, with deletes off by default. If the
lesson of Mem0 is about the mechanism rather than the library, it should
show here, and what shows is a third way for the same mechanism to leave a
context stale (Table~\ref{tab:langmem}). On the probe every write is stored, the update fires, and the
current version ranks first. Over the \sysNStatements{} anchored statements
the system makes \lgmQwenInsertsZero{} inserts and \lgmQwenUpdatesZero{}
updates (\lgmQwenNoopZero{} statements change nothing) at
\lgmQwenCallsPerStmtZero{} model calls per statement and \lgmQwenHoursZero{}
hours, and ends at \lgmQwenSlotsZero{} slots: it rewrites the memories an
incoming statement resembles \emph{and} inserts a memory for the statement,
so a superseded value is overwritten in place while the footprint stays that
of store-all, and the store also holds memories no statement wrote (``the
user is likely interested in cycling'', a confidence annotation appended to a
fact). The library logs and skips \lgmQwenPatchFail{} patches the model
emitted that could not be applied. By id it reaches current@1
\lgmQwenCurZero{} at contamination@5 \lgmQwenContamZero{}; by text,
\lgmQwenCurTextZero{} at \lgmQwenContamTextZero{} --- the two agree, because
an update here rewrites the text --- with a miss rate of \lgmQwenMissZero{},
since the memory a query lands on is sometimes the library's own gloss on the
fact rather than the fact\ifdefined\lgmQwenRdAsIsZero{}; the reader answers
\lgmQwenRdAsIsZero{} from its own top-5 and \lgmQwenRdMergedZero{} once the
query is resolved and same-entity memories merged\fi{}. Anchor-free the
mechanism stops helping where the read path cannot find the memory: current@1
is \lgmQwenCurOne{} by id and \lgmQwenCurTextOne{} by text, and
\lgmQwenMissOne{} of queries miss outright, because a memory written from a
short form (``Ebrard now occupies the office of \ldots'') is not what the
full-name query retrieves\ifdefined\lgmQwenRdAsIsOne{}; the reader answers
\lgmQwenRdAsIsOne{} as returned and \lgmQwenRdMergedOne{} with the read-path
fix, which is where a resolver on the query pays\fi{}.\ifdefined\lgmGlmCurZero{}
With glm-4.6 in the loop the same library does something else again. It
consolidates: \lgmGlmSlotsZero{} slots anchored, and the memory a query lands
on is no longer a statement but a biography --- ``Career: Honda
(2000--2003), Yamaha (2004--2010), Ducati (2011--2012), Yamaha (2013--2020),
Petronas Yamaha (2021)'' --- \lgmGlmTextLenZero{} characters long against
\lgmQwenTextLenZero{} with the smaller model, naming
\lgmGlmValuesNamedZero{} of the fact's values on average against
\lgmQwenValuesNamedZero{}. By id it is current for \lgmGlmCurZero{} of
queries; by text it is current for \lgmGlmCurTextZero{} and \emph{hedged}
for \lgmGlmHedgedZero{}, the category this paper's text-level scorer was
built for and that neither Mem0 nor Graphiti populated. And
\lgmGlmPostWindowZero{} of those top memories name a year after 2020 --- the
model has written what it knows into the store, including, for Sagan,
Team TotalEnergies (2022--2023), a team he did ride for after the window and
one the stream never mentioned. The larger model turns the store into a chronology
that is right about the world, hedged about the benchmark's question, and
partly not derived from the statements it was given\ifdefined\lgmGlmRdAsIsZero{}.
The reader pays for that: it answers \lgmGlmRdAsIsZero{} from these
memories against \lgmQwenRdAsIsZero{} from the smaller model's, and its
errors are not stale (\lgmGlmRdStaleZero{}) but \emph{neither} value
(\lgmGlmRdOtherZero{}) --- a chronology that runs past the window invites the
answer that is true today rather than the one the source defines. The
read-path fix, which lifts the smaller model's store to
\lgmQwenRdMergedZero{}, has nothing to merge here (\lgmGlmRdMergedZero{})\fi{}.\fi{}
\fi{}

\ifdefined\ersQwenCurZero
\begin{table*}[t]
\centering
\caption{The fourth mechanism: ERASE, rewrite-on-arrival \citep{li2024erase},
reimplemented from the paper because no release pins it. When a statement
arrives, the \ersQwenRetrieveK{} stored facts nearest to it are retrieved and
one model call decides, for each, KEEP, REWRITE or DELETE; the statement is
then inserted. Same \sysNFacts{}-fact streams, MiniLM encoder, reader
(\sysrdAdjReader{}, own top-5). contamination@5 is by text, since a rewrite
can keep the superseded value's string in a past-tense clause.}
\label{tab:erase}
\footnotesize
\setlength{\tabcolsep}{4pt}
\begin{tabular}{lcccccccc}
\toprule
& \multicolumn{4}{c}{anchored} & \multicolumn{4}{c}{anchor-free} \\
\cmidrule(lr){2-5}\cmidrule(lr){6-9}
policy & cur@1 & by text & cont.@5 (text) & reader & cur@1 & by text & cont.@5 (text) & reader \\
\midrule
ERASE, qwen2.5:14b & $0.853$ & $0.860$ & $0.100$ & $0.867$ & $0.733$ & $0.707$ & $0.160$ & $0.807$ \\
ERASE, glm-4.6 & $0.953$ & $0.953$ & $0.060$ & $0.980$ & $0.713$ & $0.713$ & $0.240$ & $0.813$ \\
Cosine (store-all) & $0.460$ & $0.460$ & $0.973$ & $0.573$ & $0.140$ & $0.140$ & $0.927$ & $0.427$ \\
+ time-rerank & $0.920$ & $0.920$ & $0.973$ & $0.300$ & $0.787$ & $0.787$ & $0.927$ & $0.220$ \\
Symmetric key & $1.000$ & $1.000$ & $0.007$ & $1.000$ & $0.920$ & $0.920$ & $0.027$ & $0.920$ \\
Overwrite: oracle & $0.987$ & $0.987$ & $0.007$ & $1.000$ & $0.887$ & $0.887$ & $0.000$ & $0.953$ \\
\bottomrule
\end{tabular}

\end{table*}

\textbf{A fourth mechanism: editing the store rather than the arrival.}
Mem0 and LangMem adjudicate the incoming memory; Graphiti invalidates the
edge it contradicts. ERASE \citep{li2024erase} edits the \emph{existing}
entries: the facts an arrival is related to are retrieved and a model keeps,
rewrites or deletes each of them before the arrival is added. No release
pins it, so we reimplemented its loop from the paper (Table~\ref{tab:erase})
under the same gate: on the six-statement probe every write is stored, the
edits fire, and the current version ranks first. Over the \sysNStatements{}
anchored statements the model makes \ersQwenRewritesZero{} rewrites and
\ersQwenDeletesZero{} deletions against \ersQwenKeepsZero{} keeps
(\ersQwenParseFailZero{} decisions unparseable and treated as keep), at
\ersQwenCallsPerStmtZero{} model calls per statement and
\ersQwenHoursZero{} hours, ending at \ersQwenSlotsZero{} slots. By id it
reaches current@1 \ersQwenCurZero{} at contamination@5 \ersQwenContamZero{};
by text, \ersQwenCurTextZero{} at \ersQwenContamTextZero{}, hedged
\ersQwenHedgedZero{}. Its failure is the one editing the store invites:
\ersQwenMissZero{} of queries miss, and for \ersQwenGoneZero{} of facts no
version survives anywhere in the returned depth --- the model, asked whether
a neighbouring fact is still accurate, deleted facts about other entities
that the arrival did not touch\ifdefined\ersQwenRdAsIsZero{}; the reader answers
\ersQwenRdAsIsZero{} from its own top-5 (stale \ersQwenRdStaleZero{}) and
\ersQwenRdMergedZero{} with the read-path fix\fi{}. Anchor-free:
\ersQwenCurOne{} by id, \ersQwenCurTextOne{} by text, \ersQwenMissOne{}
missed, \ersQwenSlotsOne{} slots\ifdefined\ersQwenRdAsIsOne{}, reader
\ersQwenRdAsIsOne{}\fi{}.\ifdefined\ersGlmCurZero{} With glm-4.6 deciding, the same loop makes
\ersGlmRewritesZero{} rewrites and \ersGlmDeletesZero{} deletions, loses
every version of only \ersGlmGoneZero{} of facts, and ends at
\ersGlmSlotsZero{} slots with current@1 \ersGlmCurZero{} by id and
\ersGlmCurTextZero{} by text at contamination@5 \ersGlmContamTextZero{}
\ifdefined\ersGlmRdAsIsZero{};
the reader answers \ersGlmRdAsIsZero{} from its own top-5\fi{}. That is the
best number any of the four mechanisms posts in this paper --- within
\ersGlmGapToOracleZero{} points of the write-path oracle on this stream
(\ersGlmGapQueriesZero{} of \sysNFacts{} queries), with
no alias table and no identifiers --- and it comes from the one mechanism
that edits the store rather than the arrival. The price is the deletions the
smaller model made wrongly: the mechanism that can remove staleness is the
one that can remove the fact.\ifdefined\ersGlmCurOne{} Anchor-free the same
model posts \ersGlmCurOne{} with stale \ersGlmStaleOne{} at
\ersGlmSlotsOne{} slots (\ersGlmDeletesOne{} deletions against
\ersGlmDeletesZero{} anchored)\ifdefined\ersGlmRdAsIsOne{}, reader
\ersGlmRdAsIsOne{}\fi{}: an arrival written with a short form does not
retrieve the full-name version it supersedes among its five neighbours, so
the edit never reaches the stale entry. The mechanism edits only what its
retriever finds, which is the anchor-free dependence of
\S\ref{sec:anchor} again, now on the write path.\fi{}\fi{}
\fi{}

Across the four the pattern is the same and the failure is never the same:
Mem0 lost the adjudication between releases, Graphiti keeps its verdicts and
discards them on read, LangMem applies them and either keeps everything or
rewrites it into a chronology, depending on the model, and ERASE applies them
to the store and, under qwen2.5:14b, to facts the arrival never mentioned. No
single scoring convention describes all four.

We ran four systems rather than more, and the choice was depth over breadth:
each is scored three ways, across releases or read paths, under a hand probe,
for \sysIngestHoursTotal{} hours of local inference (Mem0, one quoted run per
release, two configurations each; five further runs separated the system from
its backend), \gphIngestHoursTotal{} hours
(Graphiti), \lgmQwenHoursZero{} hours per
configuration (LangMem) and \ersQwenHoursZero{} hours
per configuration (ERASE). The release ships the harness and the probe protocol so that
a leaderboard, rather than this paper, is where systems accumulate.

We draw one methodological conclusion for this benchmark rather than for Mem0.
A leaderboard for agent memory must pin and publish the version of every system
it scores, because the mechanism under test can leave the default code path
between minor releases without any change to the documented interface. Our
release records it per run.
\fi

\section{Reader track: constructed contexts, census details, and presentation order}
\label{app:readerx}

\begin{table*}[t]
\centering
\caption{Constructed contexts: reader accuracy when contamination is
manipulated rather than observed (test, $n$ facts per condition), with and
without the stored year surfaced in the context.}
\label{tab:dose}
\small
\setlength{\tabcolsep}{5pt}
\begin{tabular}{lccccc}
\toprule
& & \multicolumn{2}{c}{no dates} & \multicolumn{2}{c}{dates} \\
\cmidrule(lr){3-4}\cmidrule(lr){5-6}
constructed context & $n$ & qwen2.5:14b & glm-4.6 & qwen2.5:14b & glm-4.6 \\
\midrule
no context (closed book) & $468$ & $0.088$ & $0.244$ & $0.088$ & $0.244$ \\
current only & $468$ & $0.991$ & $0.987$ & $0.953$ & $0.985$ \\
+1 stale, current first & $468$ & $0.314$ & $0.592$ & $0.925$ & $0.938$ \\
+1 stale, current last & $468$ & $0.850$ & $0.637$ & $0.970$ & $0.972$ \\
+2 stale, current middle & $42$ & $0.262$ & $0.405$ & $0.952$ & $0.976$ \\
stale only & $468$ & $0.000$ & $0.000$ & $0.002$ & $0.000$ \\
\bottomrule
\end{tabular}

\end{table*}

\begin{figure*}[t]
\centering
\includegraphics[width=0.49\linewidth]{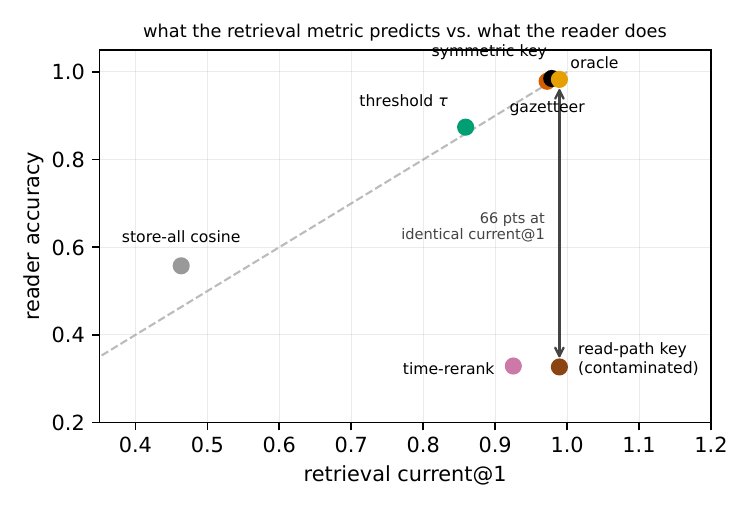}
\hfill
\includegraphics[width=0.49\linewidth]{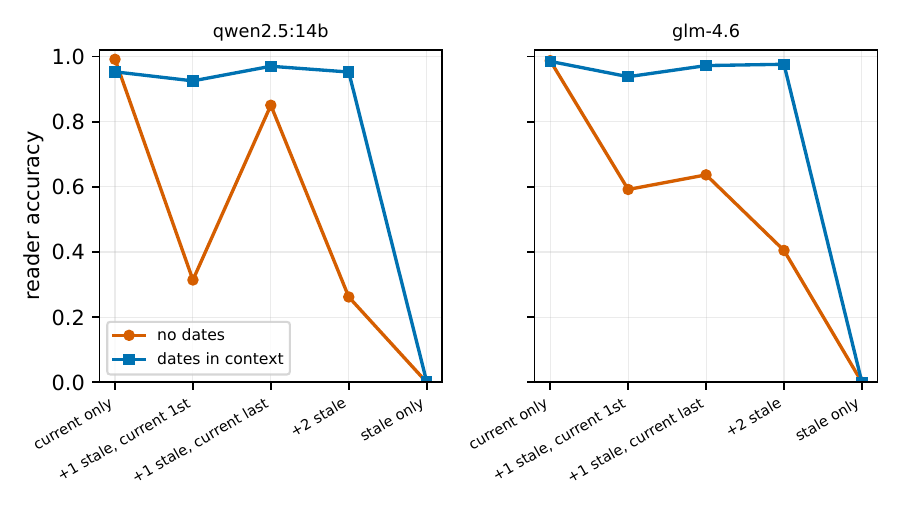}
\caption{Left: retrieval current@1 against reader accuracy for the primary
reader; the dashed line is what a retrieval metric would predict. The arrow is
the controlled pair --- two keyers that return the same statement at rank 1 on
every query of this configuration, so they share an $x$ exactly and the vertical distance is what
contamination of ranks 2--\readerTopK{} costs. Right:
constructed contexts, where contamination is manipulated rather than observed.}
\label{fig:reader}
\end{figure*}

Figure~\ref{fig:reader} plots the policy track (left) and the constructed
contexts of Table~\ref{tab:dose} (right). The two readers fail differently, and the decomposition matters. For
\readerPriName{} the gap is almost entirely the reader asserting a superseded
value: stale answers rise by \pairPriStaleShare{} points
and hedging by only \pairPriHedgeShare{}. For \readerSecName{} it splits
\pairSecStaleShare{} stale against \pairSecHedgeShare{} hedged --- a
meaningful part of its penalty is naming \emph{both} values rather than the
wrong one. Hedging is the better failure, and a benchmark that scored only
``contains the current value'' would miss the distinction entirely; we report
it because it makes the primary reader's result stronger and the secondary
reader's more qualified.

The obvious objection is that this is the position effect again: rank 1 is
fixed, so the stale versions all sit \emph{below} the current one, and
the constructed-context probe below shows that arrangement is exactly the one \readerPriName{}
handles worst. The second reader answers it. \readerSecName{}'s positional
penalty is \posSecGain{} points with a $[\posSecCiLo, \posSecCiHi]$ interval
and $p\,\posSecP$ --- we fail to detect a positional effect for it at all, which is not the same as establishing there is none, against
\posPriGain{} points at $p\,\posPriP$ for \readerPriName{} --- and it still
pays \pairSecDelta{} points on the pair at $p\,\pairSecP$. On a reader for
which we cannot measure a position effect, contamination alone still costs
almost thirty points.

The pair's \emph{direction} is a property of the design --- rank 1 is held
fixed, so only contamination can move the reader --- but its \emph{magnitude}
is not context-size-free. We hold the context at top-\readerTopK{} throughout,
matching the $k$ of contamination@5; a larger context admits more stale
versions and a smaller one fewer, so the gap should be read as the cost of
contamination at one context size we fixed in advance, not as a constant.

\textbf{The mechanism is positional, and the probe is causal.} In the policy
track contamination is observed rather than assigned, so that comparison cannot
separate ``the context contains a stale version'' from ``the context is ordered
differently''. Constructed contexts (Table~\ref{tab:dose}) can, and they split the harm into two identified parts.
From the current statement alone (\doseCurrentOnly{} correct),
\emph{prepending} one superseded version costs \contamDrop{} points (95\% CI
[\contamDropLo, \contamDropHi], $n{=}\contamNPaired$, sign-flip permutation
$p\,\contamDropP$) --- the cheapest arrangement contamination admits, since the
current statement still sits last. Putting the \emph{same two statements} in the
other order costs \posGain{} points more (\posCurrentFirst{} with the current
version first against \posCurrentLast{} with it last; 95\% CI [\posGainLo,
\posGainHi], $p\,\posP$), so the full cost of one stale neighbour ranges from
\contamDrop{} to \doseWorstDrop{} points depending only on where it sits. These
are not two independent additive effects --- ordering is a within-contamination
effect and cannot exist without it --- but conditional on contamination being
present, arrangement accounts for roughly four times the best-case harm. The
reader largely answers with whatever the context says last, which is exactly
the configuration timestamp reranking creates.
Reranking does not change which statements are retrieved, only their
order: its top-\readerTopK{} is plain cosine's top-\readerTopK{} for
\readerSameSetAll\% of queries, and on \readerSameSetReordered\% of them the
order actually differs --- promoting the current version to the front and
leaving stale versions adjacent to the question. A context holding only a stale version yields \doseStaleOnly{}. We do not
report a dose effect beyond one stale version: the two-stale condition exists
only for the \dosePlusTwoN{} facts with two distinct superseded values, and
comparing its \dosePlusTwo{} against a \nTest-fact condition would contrast
populations rather than doses.

\textbf{The census.} contamination@5 was introduced in \S\ref{sec:bench} on
the strength of prior work about readers, and the reader track supports the
direction of that argument: the policies that drive it to zero are the ones
whose gains transfer when no date is shown. Over every policy pair in each of the \concordNBlocks{}
reader policy tracks, current@1 orders \concordCurAgree{} of
\concordCurTotal{} pairs the way reader accuracy does and inverts
\concordCurInvert{}; contamination@5 agrees on \concordContamAgree{}, ties on
\concordContamTie{}, and inverts \concordContamInvert{}. That census includes
the retain-everything control, which we built to be pathological, and pairs
the reader itself separates by less than a point. Dropping both leaves
\concordCurTotalStrict{} comparisons between policies a practitioner might
deploy: current@1 still inverts \concordCurInvertStrict{} of them and
contamination@5 inverts \concordContamInvertStrict{}, at any gap threshold. An
exact paired test over the restricted census gives $p\,\concordExactP$ --- a
description of where the metrics disagree rather than a test.

contamination@5's census ties are real --- cosine and time-rerank share a
contamination@5 of \realCosineContam{} by construction (\S\ref{sec:policies})
while their reader accuracies are \readerPriCosine{} and \readerPriRerank{} ---
so \concordTieOnCurInvert{} of its \concordContamTie{} ties are comparisons
current@1 gets \emph{wrong}; the rest are comparisons current@1 also gets
right, and we do not claim it declines only the hard ones. A metric that
refuses to rank is weaker than one that ranks correctly, and stronger than one
that ranks backwards.

Three caveats. The \concordCurTotalStrict{} comparisons are
\concordCurPairsStrict{} distinct policy pairs scored against
\concordNBlocks{} readers, not \concordCurTotalStrict{} independent
observations. The current@1 inversions concentrate on two pairs, both
involving timestamp reranking --- which is the section's thesis restated, not
independent corroboration of it. What these results actually argue for is a position-aware successor --- the
rank of the \emph{last} stale version in the top-$k$ --- which the released
evaluator makes a one-function addition and which we propose rather than claim. The caveat belongs here rather than in the
limitations: a retention context differs from a store-all context in two ways
at once --- it holds one version of the queried fact instead of several, and
therefore has room for more statements about other facts. That is what these
policies genuinely produce, so the policy-track gap is a property of context
composition as a whole rather than of contamination alone; the
constructed-context probe is the isolated measurement, and it is the one the
causal claim rests on. That probe runs on \doseNModels{} readers
(\doseModelList{}), so the \emph{mechanism} is established causally on both
rather than on the one that exhibits the sign flip. Both agree on what matters
and disagree on what does not. A clean single-version context is answered
almost perfectly by either (\doseCurrentOnly{} and \doseSecCurrentOnly{}), and
adding a single stale version costs, in the worse order,
\doseDropMin{}--\doseDropMax{} points
depending on the reader --- the range is the disagreement, the direction is
not. Where they part company is \emph{position}: for \readerPriName{} the
penalty depends violently on where the current version sits
(\doseCurrentFirst{} first vs \doseStaleFirst{} last), while for
\readerSecName{} the two orders are close (\doseSecCurrentFirst{} vs
\doseSecStaleFirst{}). Contamination is what generalizes; the positional
sensitivity to it is a property of the reader, which is the asymmetry
\S\ref{sec:reader}'s order ablation exploits.\ifdefined\doseSecCurrentFirstDates{}
Surfacing the stored year removes most of the penalty on both: the worst
ordering recovers from \doseCurrentFirst{} to \doseCurrentFirstDates{} for
\readerPriName{} and from \doseSecCurrentFirst{} to
\doseSecCurrentFirstDates{} for \readerSecName{}.\fi{} Two weaker forms of the same
effect are established on both readers as well. Retrieval gains fail to
transfer under either; and within the threshold keyer --- the one policy that
produces a substantial mix of clean and contaminated contexts --- splitting its
own queries by whether its top-\readerTopK{} was contaminated gives
\condPriDirty{} against \condPriClean{} for \readerPriName{}
($n{=}\condPriNDirty$ vs $n{=}\condPriNClean$) and \condSecDirty{} against
\condSecClean{} for \readerSecName{}: a \condPriDelta{} and
\condSecDelta{}-point gap, same policy, same reader, same prompt. That is an
association rather than a manipulation, but it replicates.

\textbf{Reranking's sign is set by a convention it does not control.} The
contexts above are presented in descending relevance --- the common convention,
which places the top-ranked statement \emph{farthest} from the question and is
therefore, by the position result established above, the arrangement least
favourable to any policy whose mechanism is promoting the right statement to
rank 1. Reversing that presentation over the identical retrieved top-$k$ is
decisive, and the effect falls on exactly the policy that should show it:

\begin{center}\footnotesize\setlength{\tabcolsep}{3pt}
\begin{tabular}{lccc}
\toprule
policy & desc.\ & asc.\ & shift \\
\midrule
Cosine (store-all) & \readerPriCosine{} & \revOrderCosine{} &
  \revOrderCosineShift{} ($p\,\revOrderCosineP$) \\
+ time-rerank & \readerPriRerank{} & \revOrderRerank{} &
  \textbf{\revOrderRerankShift{}} ($p\,\revOrderRerankP$) \\
Overwrite: gazetteer & \readerPriLex{} & \revOrderLex{} &
  \revOrderLexShift{} ($p\,\revOrderLexP$) \\
Overwrite: oracle & \readerOracleCorrect{} & \revOrderOracle{} &
  \revOrderOracleShift{} ($p\,\revOrderOracleP$) \\
\bottomrule
\end{tabular}
\end{center}

The aggregate shifts understate what is happening, and the per-answer counts
are the informative statistic. Reordering changes \flipCosine{} of \nTest{}
individual answers for store-all cosine --- it is just as order-fragile as
anything else --- but the changes cancel, leaving \revOrderCosineShift{} points
($p\,\revOrderCosineP$). For time-reranking \flipRerank{} answers change and
they do \emph{not} cancel: \revOrderRerankShift{} points. The retention
policies change \flipLex{} answers. Not a null result at the aggregate --- zero
answers, on every query, because keeping one statement per fact leaves nothing
about the queried fact to reorder. Store-all retrieval is order-fragile at the
item level and retention is order-invariant at the item level, which is a
property of the store, not of the reader.

This qualifies the earlier paragraph in one direction and sharpens it in
another. The inversion is \emph{not} a general property of timestamp
reranking: relative to the store-all baseline it is worth \rerankVsBaseFwd{}
points under the common convention and \rerankVsBaseRev{} under the reversed
one, so a practitioner can change its sign by changing how a prompt is
assembled --- itself worth knowing, and not something current@1 can express.
What survives every convention is the gap it does not close: against a
\readerRetrievalDelta-point retrieval improvement its best case at the reader
is \rerankVsBaseRev{}, while retention is exactly invariant, its margin over the baseline moving from
\lexVsBaseFwd{} to \lexVsBaseRev{} only because the baseline itself moves. \textbf{For this reader, the retrieval metric overstates reranking under both
presentations tested and understates retention under neither} --- the order
ablation ran on \readerPriName{} only. It is the one experiment here we did
not run on the hosted reader, and the reason to want it there is
specific: \readerSecName{} shows a much weaker positional penalty in the
constructed-context probe (\doseSecCurrentFirst{} vs \doseSecStaleFirst{},
against \doseCurrentFirst{} vs \doseStaleFirst{}), so the sign flip this
ablation measures is plausibly smaller or absent for it. We report the
invariance of retention as the general claim and the sign flip as a
demonstration on one reader that it \emph{can} occur, not that it always does.

\section{Does it hold at scale, and with set-valued gold?}
\label{sec:scale}

\begin{table*}[t]
\centering
\caption{v1.1 scale track: the same ladder on every changing TempLAMA fact
rather than only the single-valued ones. \scaleNTest{} test facts
(\scaleNMulti{} of them multi-valued, mean \scaleMeanCurrentValues{} currently
valid values each) against \nTest{} in v1.0, scored identically --- a top-1 is
current when its value is valid in the final year --- plus \emph{set
recall@5}: the fraction of a fact's distinct currently valid values whose
statements appear in the top-5.}
\label{tab:scale}
\small
\setlength{\tabcolsep}{4pt}
\begin{tabular}{lcccccc}
\toprule
& \multicolumn{3}{c}{anchored} & \multicolumn{3}{c}{anchor-free} \\
\cmidrule(lr){2-4}\cmidrule(lr){5-7}
policy & current@1 & cont.@5 & set rec.@5 & current@1 & cont.@5 & set rec.@5 \\
\midrule
Cosine (store-all) & $0.470$ & $0.837$ & $0.896$ & $0.232$ & $0.681$ & $0.742$ \\
+ time-rerank & $0.721$ & $0.837$ & $0.896$ & $0.524$ & $0.681$ & $0.742$ \\
+ recency-wt & $0.699$ & $0.808$ & $0.940$ & $0.385$ & $0.579$ & $0.706$ \\
Overwrite: threshold & $0.777$ & $0.167$ & $0.791$ & $0.493$ & $0.266$ & $0.666$ \\
Overwrite: gazetteer & $0.916$ & $0.009$ & $0.851$ & $0.293$ & $0.543$ & $0.709$ \\
Overwrite: fuzzy linker & $0.900$ & $0.013$ & $0.838$ & $0.512$ & $0.051$ & $0.556$ \\
Overwrite: linker + alias table & $0.914$ & $0.009$ & $0.849$ & $0.576$ & $0.052$ & $0.640$ \\
Fuzzy linker, read+write & $0.932$ & $0.013$ & $0.846$ & $0.627$ & $0.054$ & $0.603$ \\
Overwrite: oracle & $0.955$ & $0.000$ & $0.890$ & $0.690$ & $0.000$ & $0.785$ \\
\bottomrule
\end{tabular}

\end{table*}

Every number so far comes from the \nChanging{} TempLAMA keys that are
single-valued in every year and change at least once, which is \nChanging{} of \nFactKeys{} (the v1.0
release). That restriction buys crisp top-1 scoring and invites the most
common objection a benchmark receives: the slice is small and possibly
unrepresentative. Keeping every changing key and scoring a top-1 as current
whenever its value is valid in the final year --- so a fact with two
simultaneously valid values simply has two current runs --- raises the
queried set to \scaleNTest{} test facts, of which \scaleNMulti{} are
genuinely multi-valued (the v1.1 scale track).

Table~\ref{tab:scale} repeats the ladder there, on \scaleNStatementsOne{}
statements, and adds the metric the single-valued track could not define. A
top-1 counts as current when its value is valid in the final year, so a fact
with several simultaneously valid values is satisfied by \emph{any} of them
--- the reading that favours retention, since a policy keeping one slot per
fact can discard a still-valid value unseen. \emph{Set recall@5} charges
that: the fraction of a fact's distinct currently valid values whose
statements appear in the top-5. It equals current@5 on single-valued facts
and prices one slot per fact on the rest. Anchored, the oracle's set recall is
\scaleSetRecOracleZero{} against \scaleSetRecRwZero{} for recency-weighted
store-all, and on the \scaleNMulti{} multi-valued facts
\scaleSetRecMultiOracleZero{} against \scaleSetRecMultiRwZero{}; every
retention rung sits below every store-all policy on this column while
beating all of them on current@1 and contamination. One slot per fact is the
right footprint for freshness and the wrong one for coverage, and the
benchmark prices the two separately.

The write-path keying ladder survives intact and one store-all comparison does not ---
a pre-specified comparison from Table~\ref{tab:real}, so it is stated
plainly. Store-all similarity still fails
(\scaleCosineCurZero{} anchored, \scaleCosineCurOne{} anchor-free); timestamp
reranking still repairs top-1 without touching contamination; and the keying
ladder keeps its shape --- anchored, the gazetteer and fuzzy rungs sit within a
few points of the oracle, while anchor-free the gazetteer collapses to
\scaleLexCurOne{} and fuzzy resolution holds at \scaleFuzzyCurOne{}.
Multi-valued gold makes the task harder for every policy in the same direction,
which is what it should do, and does not reorder the \emph{keying} ladder. The
v1.0 restriction bought scoring clarity, not a favourable result.

One ordering does reverse, and it is the cheap fix losing. Anchored, v1.0 has
timestamp reranking ahead of the threshold keyer (\realRerankCur{} against
\realReconCur{}); at this scale the threshold keyer leads,
\scaleReconCurZero{} $[\scaleReconCiLo, \scaleReconCiHi]$ against
\scaleRerankCurZero{} $[\scaleRerankCiLo, \scaleRerankCiHi]$ --- a sign
reversal with non-overlapping intervals. The mechanism is the one
\S\ref{sec:strat} already identifies: reranking only repairs a query when its
window $k$ spans every version of the fact, and multi-valued facts have more
versions to span. So the statement is not that everything survives --- it is
that what survives is the keying result, and the metadata heuristic's
advantage over the cheapest keyer is the thing that does not.

The symmetric rung of \S\ref{sec:anchor} holds at this scale, where the
comparison is better powered: it reaches \scaleFuzzyRWCurOne{} anchor-free
for a deficit of \scaleFuzzyRWDefOne{} points, against \scaleAliasDefOne{}
for the rung handed a complete alias table on the write path alone, and
\scaleFuzzyRWDefZero{} anchored --- on mechanical short forms;
Appendix~\ref{app:natural} shows where natural ones remove that advantage.

\section{Data statement}
\label{app:datastatement}

Following \citet{bender2018datastatements}, and covering the artifact
questions of the Responsible NLP checklist.

\textbf{Curation rationale.} The queried set is not a sample of language use;
it is a deliberately filtered set of \emph{facts that changed}. From
TempLAMA's \nFactKeys{} fact keys we keep those single-valued in every year
(so top-1 scoring is crisp) whose value changes at least once in
2010--2020 (\nChanging{} facts). Facts that never change are excluded from the
queried set precisely because they cannot exhibit the failure under study;
they re-enter as distractors in the \texttt{full\_corpus} configuration
(Appendix~\ref{app:benchdetails}). This filtering makes the resource an
instrument for one phenomenon rather than a sample of world knowledge, and
prevalence estimates should not be read off it.

\textbf{Language variety.} English only (\texttt{en-US}/\texttt{en-GB}
undifferentiated), in two registers, both authored by us rather than found:
declarative statement templates and interrogative query templates
(Appendix~\ref{app:templates}). Entity names appear as Wikidata's English
labels, which retain diacritics and non-English orthography for non-anglophone
entities (e.g.\ \emph{Juan Rom\'an Riquelme}). No text is drawn from
naturally occurring corpora, so no dialect or idiolect is represented.

\textbf{Speaker and annotator demographic.} There are no speakers and no
crowdworkers. Every statement is generated by template from a
subject--relation--object triple, and every currency label is computed from
Wikidata validity qualifiers by the released code
(\texttt{label\_source: wikidata\_validity\_year} in the key files). No human
judgment enters the labels, which is a design guarantee of the benchmark
rather than a resourcing decision: a label derived from human annotation could
not be claimed to be independent of the systems under test.

\textbf{Subject demographic.} The entities named are public figures and
organizations whose employment, office, party, team, or coaching relations
Wikidata records with temporal qualifiers. The distribution is inherited from
Wikidata's coverage and is therefore skewed toward anglophone, male, and
Global-North sports and politics; we did not rebalance it, and per-relation or
per-region claims should not be made from this sample
(see the Limitations section).

\textbf{Speech situation.} Synthetic and asynchronous: statements are
presented to a memory system as a time-ordered stream with year metadata, and
queries are posed after the stream ends. No interaction, no dialogue, no
intended audience.

\textbf{Text characteristics.} Nine relations, one canonical statement
template plus three paraphrase templates each, three query templates each
(Appendix~\ref{app:templates}). Statements are single sentences, typically
6--14 tokens. The anchor-free configurations replace the subject's surface
form in non-introducing statements with a mechanically derived short form, or
on the natural configurations with a Wikidata alias or a Wikipedia redirect
title (Appendix~\ref{app:natural}); the redirect-derived forms are Wikipedia
page titles, released with the inventory under CC BY-SA 4.0 rather than CC0.

\textbf{Provenance appendix.} Facts: Wikidata (CC0) via TempLAMA
\citep{dhingra2022templama}, which its authors distribute under the Apache
License 2.0. Templates, the mechanical alias inventory, evaluator, and baselines: ours,
released CC0 and Apache-2.0 respectively.

\section{Controlled regime map}
\label{sec:controlled}

\begin{table*}[t]
\centering
\caption{Controlled diagnostic, current@1 (mean {\tiny$\pm$95\% CI half-width}
over \ctrlSeeds{} seeds). Left: update visibility (\texttt{value\_gap}).
Right: surface variation (\texttt{key\_jitter}) at gap $0.4$. Every policy is
re-tuned per column on \ctrlTuneSeeds{} held-out generator seeds; the selected
values vary across columns; their ranges are given in Appendix~\ref{app:tuned}.}
\label{tab:controlled}
\small
\setlength{\tabcolsep}{3pt}
\begin{tabular}{lccccccc}
\toprule
& \multicolumn{3}{c}{\texttt{value\_gap}} & \multicolumn{4}{c}{\texttt{key\_jitter}} \\
\cmidrule(lr){2-4}\cmidrule(lr){5-8}
policy & $0.3$ & $0.5$ & $0.8$ & $0.0$ & $0.2$ & $0.4$ & $0.8$ \\
\midrule
Cosine (store-all) & $0.25$\,{\tiny$\pm$0.05} & $0.22$\,{\tiny$\pm$0.06} & $0.24$\,{\tiny$\pm$0.05} & $0.23$\,{\tiny$\pm$0.06} & $0.27$\,{\tiny$\pm$0.06} & $0.24$\,{\tiny$\pm$0.06} & $0.15$\,{\tiny$\pm$0.05} \\
+ time-rerank & $0.84$\,{\tiny$\pm$0.03} & $0.82$\,{\tiny$\pm$0.04} & $0.80$\,{\tiny$\pm$0.04} & $0.83$\,{\tiny$\pm$0.03} & $0.92$\,{\tiny$\pm$0.04} & $0.62$\,{\tiny$\pm$0.06} & $0.30$\,{\tiny$\pm$0.07} \\
+ recency-weight & $0.99$\,{\tiny$\pm$0.01} & $0.97$\,{\tiny$\pm$0.02} & $0.95$\,{\tiny$\pm$0.02} & $0.98$\,{\tiny$\pm$0.01} & $0.90$\,{\tiny$\pm$0.04} & $0.60$\,{\tiny$\pm$0.05} & $0.31$\,{\tiny$\pm$0.06} \\
Reconsolidation & $1.00$\,{\tiny$\pm$0.00} & $1.00$\,{\tiny$\pm$0.00} & $1.00$\,{\tiny$\pm$0.00} & $1.00$\,{\tiny$\pm$0.00} & $0.80$\,{\tiny$\pm$0.05} & $0.54$\,{\tiny$\pm$0.05} & $0.25$\,{\tiny$\pm$0.06} \\
Oracle & $1.00$\,{\tiny$\pm$0.00} & $1.00$\,{\tiny$\pm$0.00} & $1.00$\,{\tiny$\pm$0.00} & $1.00$\,{\tiny$\pm$0.00} & $0.99$\,{\tiny$\pm$0.01} & $0.70$\,{\tiny$\pm$0.05} & $0.30$\,{\tiny$\pm$0.05} \\
\bottomrule
\end{tabular}

\end{table*}

\begin{table*}[t]
\centering
\begin{minipage}[t]{0.48\linewidth}
\centering
\caption{Slot budget $B$ (gap $0.4$, $E{=}12$ entities): current@1. Store-all
fixes need the store; retention keeps one slot per fact and is budget-immune
once $B \ge E$.}
\label{tab:budget}
\small
\begin{tabular}{lccc}
\toprule
policy & $B{=}12$ & $B{=}20$ & $B{=}40$ \\
\midrule
Recency (FIFO) & $0.60$ & $0.75$ & $0.79$ \\
Cosine + time-rerank & $0.47$ & $0.65$ & $0.82$ \\
Reconsolidation & $1.00$ & $1.00$ & $1.00$ \\
Keep-latest oracle & $1.00$ & $1.00$ & $1.00$ \\
\bottomrule
\end{tabular}

\end{minipage}\hfill
\begin{minipage}[t]{0.48\linewidth}
\centering
\caption{$\tau$ sensitivity (current@1). Clean: perfect across
$\tau\in[0.3,0.75]$, false-merging below and false-splitting at $\tau{=}0.9$.
Under jitter $0.4$ the optimum moves to $\tau{=}\ctrlBestTauJitterFour$ and
buys \ctrlTauDefaultJitterFourCur{} $\to$ \ctrlBestTauJitterFourCur{} --- real,
and still short of the oracle.}
\label{tab:tau}
\small
\begin{tabular}{lcc}
\toprule
$\tau$ & clean & jitter $0.4$ \\
\midrule
$0.1$ & $0.73$ & $0.50$ \\
$0.2$ & $0.95$ & $0.54$ \\
$0.3$ & $1.00$ & $0.42$ \\
$0.45$ & $1.00$ & $0.25$ \\
$0.6$ & $1.00$ & $0.24$ \\
$0.75$ & $1.00$ & $0.24$ \\
$0.9$ & $0.24$ & $0.24$ \\
\bottomrule
\end{tabular}

\end{minipage}
\end{table*}

The generator (Table~\ref{tab:controlled}) varies three preconditions the real
data cannot vary independently. Every policy is re-tuned on \ctrlTuneSeeds{}
held-out generator seeds at every setting, as on the real track; fixing
$k$, $\lambda$ and $\tau$ across the whole
controlled track would understate three of the four conclusions below.
\textbf{Update visibility:} threshold keying stays perfect across the
\texttt{value\_gap} range (\ctrlReconGapHighCur{} even at gap $0.8$)
\emph{because $\tau$ is retuned}, from \ctrlTunedTauGapLow{} at gap $0.3$ down
to \ctrlTunedTauGapHigh{} at gap $0.8$. Held at a fixed $\tau{=}0.6$,
the same regime gives \ctrlFixedTauGapHighCur{}: the
knob does not break threshold keying, it \emph{moves its operating point}. That
is the weaker claim, and the useful one --- the required $\tau$ tracks a
property of the data the keyer is trying to infer, and is not knowable in
advance. \textbf{Surface variation:} \texttt{key\_jitter}
degrades similarity keying ($1.00 \to \ctrlReconJitterTwoCur{}$ at jitter
$0.2$, \ctrlReconJitterFourCur{} at $0.4$) faster than it degrades the oracle
(\ctrlOracleJitterTwoCur{}, \ctrlOracleJitterFourCur{}), leaving keying
deficits of \ctrlKeyDeficitJitterTwo{}, \ctrlKeyDeficitJitterFour{} and
\ctrlKeyDeficitJitterEight{} at jitter $0.2/0.4/0.8$. Threshold tuning matters
here, and a grid that stops too high hides it: the optimal $\tau$ \emph{moves down} as the key signal
degrades (\ctrlBestTauClean{} clean, \ctrlBestTauJitterFour{} at jitter
$0.4$), and at jitter $0.4$ that difference is worth
\ctrlTauDefaultJitterFourCur{} $\to$ \ctrlBestTauJitterFourCur{}
(Table~\ref{tab:tau}). What survives is the weaker and more useful statement:
a threshold keyer can be retuned to track a degrading key signal, but not back
to the oracle, and the required $\tau$ is not knowable without the labels the
keyer is trying to infer. At jitter $0.8$ the knob degrades query--item
matching itself (the oracle falls to \ctrlOracleJitterEightCur{}), so readings
there say little about keying. \textbf{Budget} (Table~\ref{tab:budget})\textbf{:} time-rerank inherits FIFO
eviction's misses (\ctrlRerankBudgetLowCur{} at $B{=}12$) because it needs the
stale copies it stores, while retention is budget-immune once one slot per fact
fits. Note finally that in the clean regime the strongest store-all fix is
recency-\emph{weighting} (\ctrlRwCleanCur{} against
\ctrlRerankJitterZeroCur{} for reranking) while on real data reranking wins:
which cheap fix is best is itself regime-dependent. One generator property is
worth stating because it shapes these numbers: with \texttt{key\_jitter}$=0$
every mention of the same (entity, version) has an \emph{identical} embedding,
so a reranker's top-$k$ fills with duplicates and its tuned $k$ is
correspondingly large (\ctrlTunedKJitterZero{}).

\section{Robustness checks}
\label{app:robust}

\textbf{Verbalization, crossed with the anchor.} \S\ref{sec:para} sweeps the
paraphrase probability $p$ with subjects verbatim, and finds it stresses only
the threshold rung. That leaves the paper's central measurement untested along
the same axis: \S\ref{sec:anchor} fixes $p{=}0.5$ and varies the anchor, so
``the anchor-free ladder is an artifact of how we phrase things'' is a live
objection that neither section alone answers. Here we cross the two, sweeping
$p$ over \paraNLevels{} levels ($p=$ \paraLevels{}; $p{=}0$ is every statement
on one template, $p{=}1$ every statement paraphrased) at both ends of the
anchor axis, re-tuning on val at each of the six cells.
Anchor-free, the write-path ladder's ordering is \paraOrderStable{}, and the
full ladder's is \paraOrderStableAll{} --- the same regime-dependent trade
\S\ref{sec:anchor} reports, appearing again as a function of verbalization
rather than of scale. Magnitudes move as they should: store-all cosine spans
\paraCosineSpread{} points across the range and the entity linker
\paraEntitySpread{}, since both depend on surface overlap. The exception is
mechanistically informative --- the symmetric rung moves \paraFuzzyRWSpread{}
points across the entire range, because resolving the query text to a key and
ranking that key by recency never asks how the sentence was phrased. Read-path
keying is verbalization-invariant close to by construction, and the sweep shows
it.

\textbf{Encoder.} With all-mpnet-base-v2 in place of MiniLM at $p{=}0.5$
(re-tuned on val), the ordering is unchanged: cosine \mpnetCosineCur{},
time-rerank \mpnetRerankCur{} (contamination \mpnetRerankContam),
recency-weighted \mpnetRwCur{}, threshold \mpnetReconCur{} (contamination
\mpnetReconContam), gazetteer \mpnetLexCur{}, oracle \mpnetOracleCur{}. One
margin is fragile and worth naming: the threshold keyer's lead over
recency-weighting shrinks from \mpnetReconLead{} points (MiniLM) to \mpnetReconLeadMp{} (mpnet).
Critically, \S\ref{sec:anchor}'s measurement is not an artifact of the primary
encoder either: under mpnet the gazetteer keyer's deficit to the oracle still
grows from \mpAliasLexDefZero{} points when anchored to \mpAliasLexDefOne{}
when anchor-free, and the entity linker's from \mpAliasEntityDefZero{} to
\mpAliasEntityDefOne{} --- the same collapse, the same surviving residue.
\textbf{Template realization.} Across three template-assignment seeds per $p$
(hyperparameters fixed from the seed-0 tuning), current@1 varies by at most
$\pm 0.02$ (threshold keyer \seedReconLoPhalf--\seedReconHiPhalf{} at
$p{=}0.5$, time-rerank \seedRerankLoPhalf--\seedRerankHiPhalf); no comparison
in Table~\ref{tab:real} is realization-dependent.
\textbf{Query wording.} Sampling the query from all three question templates
instead of the canonical one moves current@1 by $<0.01$ for every policy
(time-rerank \qpRerankCur{}, threshold \qpReconCur{}); the failure is a
property of value-agnosticity, not of one question phrasing.

\section{Analysis}
\label{sec:analysis}

\textbf{Why value-agnostic queries defeat similarity.} The query embeds the
entity; each version embeds entity $+$ value. The query--version similarity is
dominated by the shared entity component, so version order is decided by noise
--- and there are more stale versions than current ones whenever a fact has
changed more than once. current@1 for store-all cosine on real data
(\realCosineCur) is close to the reciprocal of the mean version count, as this
account predicts, and the stratification (Table~\ref{tab:strat}:
\stratCosineTwo{} vs \stratCosineMore{}) matches it stratum by stratum.

\textbf{What overwrite destroys.} Auditing the threshold keyer's overwrites at
$\tau{=}\tunedTau$, $p{=}0.5$: \auditOverwrites{} of \auditNEvents{} events overwrite a
slot; \auditCross{} of them (\auditCrossPct\% of the stream) overwrite a
\emph{different} fact --- irreversible false merges that leave \auditZero{}
facts with no surviving memory --- while \auditDup{} residual duplicate slots
are false splits that keep stale versions retrievable. This is why its
current@5 (\realReconCurFive) is \emph{below} store-all's
(\realCosineCurFive): destroyed evidence cannot be re-ranked back.
Table~\ref{tab:mechanism} prices the remedy: invalidation instead of deletion
costs footprint and nothing else.

\textbf{What the cheap fixes actually buy.} Timestamp reranking is two lines
of code for a $+\realRerankJump$ jump in current@1, and StaleBench prices its
boundaries: the context still carries the contradiction, the repair
concentrates on 2-version facts, the store keeps every version, eviction
undercuts it, it is hostage to timestamp quality --- and, most consequentially,
for one of our two readers it leaves the end task worse than no fix at all
(\S\ref{sec:reader}). A benchmark that reported only current@1 would have
counted it a clear gain.

\section{Reproducibility}
\label{app:repro}

All numbers are generated from result files produced by
\texttt{python -m stalebench.run} (controlled; \ctrlSeeds{} seeds),
\texttt{run\_real} (TempLAMA), \texttt{run\_alias} (anchor-free, corpus,
mechanism), \texttt{run\_reader}/\texttt{run\_llmkey} (LLM tracks),
\texttt{run\_scale} (set-valued gold), \texttt{run\_paraphrase}
(verbalization sweep), \texttt{run\_systems} (deployed systems), and
\texttt{diag\_pair} (the controlled pair's rank-1 agreement), and injected
into the manuscript by a script; no result is hand-transcribed. Reader calls
that failed outright (provider content filter, unrecoverable error) are counted
rather than dropped silently: \readerNFailed{} across the reported runs. The generator
refuses to emit a macro twice with conflicting values, and asserts the
premise the controlled pair depends on, so a stale or contradicted number
fails the build rather than reaching the page.
TempLAMA splits are the public GCS release of \citet{dhingra2022templama};
embeddings use publicly released sentence-transformers checkpoints
\citep{reimers2019sbert}. Controlled-track intervals are bootstrapped over generator seeds rather than
queries, and each seed contributes only 12 queries, so their per-seed
granularity is coarse; real-data intervals resample the \nTest{} test queries
directly. Bootstrap CIs use $10^4$ resamples;
paired tests are two-sided sign-flip permutation tests with $10^4$
permutations, Holm-corrected within a family pre-specified in code before test evaluation:
four comparisons for Table~\ref{tab:real}, and \aliasHolmFamily{} for the
anchor-free track, which grew as rungs were added. A larger
family is a harsher correction, so this is conservative, but the same raw $p$
is corrected differently in the two tracks and that is worth stating. Every LLM call is greedy-decoded and cached on disk keyed by (model, prompt,
max\_tokens); the cache ships with the release, so the reader and LLM-keyer
numbers regenerate offline with no API access. It also contains judgments from
runs that were started and abandoned --- a third reader whose provider quota was
exhausted mid-run, and a partial second LLM-keyer model --- which are shipped as
found rather than deleted, and are not reported anywhere in this paper.
Answers are capped at 32 tokens, which suits the short-answer scoring but would
penalise a verbose reader; that cap is a scope condition on the reader metric,
not a property of the benchmark.

\textbf{Compute.} Everything reported here runs on one laptop CPU apart from
the LLM tracks: the controlled sweeps, the TempLAMA tracks, and the alias and
corpus tracks together take under an hour, dominated by sentence-transformer
encoding, and are cached thereafter. The LLM tracks issue
\llmCacheEntries{} model calls, \llmCacheLocal{} served by a locally hosted
open-weights model and \llmCacheHosted{} by hosted models; the answers are
a few tokens each and every call is cached. No GPU is required to reproduce any
number in this paper.

The systems track (\S\ref{sec:systems}) is the exception to the cost profile
above. Streaming \sysNStatements{} statements through a wrapped memory system
costs \sysIngestHours{} hours of local LLM inference for Mem0's slowest
configuration and \sysIngestHoursTotal{} hours across its two quoted
runs\ifdefined\gphIngestHoursTotal{}, and \gphIngestHoursTotal{} hours for
Graphiti, whose extraction makes \gphCallsPerStmtZero{} model calls per
statement\fi{}, because each system issues its own extraction and
adjudication calls. Reproducing it additionally requires a pinned
\texttt{mem0ai} version, one of the vector backends in Table~\ref{tab:gates},
a pinned \texttt{graphiti-core} with a FalkorDB server, and a local Ollama
serving the adjudicator model --- which is why that track runs on a
\sysNFacts-fact subsample. On the anchored configuration the subsample reproduces store-all cosine,
time-rerank and the oracle of the full test
split to within \sysSubsampleMaxDev{} points; the keying rungs score higher
on it than on the full split (Appendix~\ref{sec:systems}).

One reproducibility hazard is worth recording, since it is easy to
introduce: ordering the
gazetteer with \texttt{sorted(set(...), key=len)} resolves ties by
\texttt{set} iteration order and therefore by the interpreter's per-process
string hash seed, and gazetteer results then drift by one query between runs.
Keyers impose a total order.

\section{Release}

The release directory contains the \relNConfigs{} configurations
(\relConfigList{}) as statement and query JSONL for both splits, the alias inventories (the derivable one, and the natural one per split that Appendix~\ref{app:natural} draws from), a datasheet, a manifest
with per-file row counts and MD5s, and \texttt{evaluate.py} --- a frozen
scorer that takes a submission of per-query rankings and reports the metrics
above. Consuming only the released files and scoring with that evaluator
reproduces the store-all rows of Table~\ref{tab:real} exactly
(\texttt{baseline.py} implements cosine and timestamp reranking, and
\texttt{trivial\_baselines.py} the two blind rules; the keying
rungs need the subject inventory and the oracle needs the answer key, so
neither is a baseline a submission could run). Facts derive from Wikidata (CC0) via TempLAMA, which its authors distribute under the Apache License 2.0; our derived splits, templates and mechanical alias inventory are released CC0, the natural alias inventory under CC BY-SA 4.0 (Appendix~\ref{app:datastatement}), and the evaluator and baselines under Apache 2.0. Redistribution of the derived data is consistent with both upstream licences. A
leaderboard built on this release is meant to host production memory systems (Mem0,
Graphiti/Zep, MemGPT-style agents) as policies under this harness. The full release directory, the harness and the result records are archived at \releasedoi{}; the repository is \releaseurl{}.

\section{Verbalization crossed with the anchor}
\label{app:paracross}

Table~\ref{tab:paracross} crosses the paraphrase probability $p$ with anchor
availability $a$, to check that the ladder's ordering does not depend on how
statements are verbalized.

\begin{table*}[t]
\centering
\caption{current@1 as verbalization diversity and anchor availability are
varied together (test split, re-tuned on val in every cell). $p$ is the
probability a statement uses a paraphrased template rather than the relation's
base one. The write-path ladder's ordering is \paraOrderStable{}; including
the symmetric rung it is \paraOrderStableAll{}. Magnitudes move --- store-all
cosine by \paraCosineSpread{} points across the range, the entity linker by
\paraEntitySpread{} --- while the symmetric rung moves \paraFuzzyRWSpread{},
because resolving the query to a key never asks how the sentence was phrased.}
\label{tab:paracross}
\footnotesize
\setlength{\tabcolsep}{4pt}
\begin{tabular}{lcccccc}
\toprule
& \multicolumn{3}{c}{anchored ($a{=}0$)} & \multicolumn{3}{c}{anchor-free ($a{=}1$)} \\
\cmidrule(lr){2-4}\cmidrule(lr){5-7}
policy & $p{=}0$ & $p{=}0.5$ & $p{=}1$ & $p{=}0$ & $p{=}0.5$ & $p{=}1$ \\
\midrule
Cosine (store-all) & $0.447$ & $0.464$ & $0.468$ & $0.085$ & $0.111$ & $0.128$ \\
+ time-rerank & $0.959$ & $0.925$ & $0.942$ & $0.801$ & $0.737$ & $0.744$ \\
Overwrite: threshold & $0.887$ & $0.859$ & $0.829$ & $0.549$ & $0.513$ & $0.526$ \\
Overwrite: gazetteer & $0.991$ & $0.972$ & $0.964$ & $0.143$ & $0.169$ & $0.177$ \\
Overwrite: fuzzy linker & $0.987$ & $0.968$ & $0.959$ & $0.754$ & $0.714$ & $0.690$ \\
Overwrite: linker + aliases & $0.991$ & $0.972$ & $0.964$ & $0.825$ & $0.776$ & $0.774$ \\
Symmetric key (read+write) & $0.989$ & $0.989$ & $0.987$ & $0.833$ & $0.831$ & $0.835$ \\
Overwrite: true-key oracle & $0.998$ & $0.979$ & $0.970$ & $0.880$ & $0.827$ & $0.808$ \\
\bottomrule
\end{tabular}

\end{table*}

\section{Tuned hyperparameters}
\label{app:tuned}

Every policy is re-tuned in every regime, on the val split (real tracks) or on
\ctrlTuneSeeds{} held-out generator seeds disjoint from the reported ones
(controlled track). The selected values vary substantially --- time-rerank's $k$
ranges over $\{2,\dots,20\}$ and the threshold keyer's $\tau$ over
$[0.1,0.6]$ --- which is the reason a single fixed setting misrepresents several
policies at once. The full table ships in
\texttt{results/stalebench\_results.json} under \texttt{tuned} and in each
real-track cell under \texttt{tuned}.

\section{Relation templates}
\label{app:templates}

Nine TempLAMA relations are verbalized with one canonical statement template
(the TempLAMA cloze with the value filled in) plus three paraphrase templates
each, and carry three value-agnostic question templates; the headline uses
the canonical question, and the query-wording check of
Appendix~\ref{app:robust} samples all three.
The full sets ship in \texttt{stalebench/templama.py}. Example (P286):
canonical ``\emph{\{v\} is the head coach of \{s\}.}''; paraphrases
``\emph{\{s\} is coached by \{v\}.}'', ``\emph{\{v\} runs the bench for
\{s\}.}'', ``\emph{The head-coaching job at \{s\} belongs to \{v\}.}'';
questions ``\emph{Who is currently the head coach of \{s\}?}'', ``\emph{Who
coaches \{s\} now?}'', ``\emph{What is the name of \{s\}'s current head
coach?}''.

\section{Limitations and ethical considerations in full}
\label{app:limits-ext}

The Limitations and Ethical Considerations sections of the main paper are
condensed from the text below.

\subsection{Limitations}

Three properties of this release bound what its numbers mean. First,
saturation, which \S\ref{sec:real} states once: \texttt{anchored} (and
\texttt{full\_corpus}, where the blind rule scores \blindFull{}) is saturated
on current@1 and \texttt{anchor\_free} is not. The strongest blind rule there
--- trigram mention linking on the read path over an inventory recovered
from the public question templates --- reaches \blindAnchorFree{} against an
oracle at \aliasOracleCurOne{}, at contamination@5 \blindAnchorFreeContam{};
it differs from the read-path rung of \S\ref{sec:anchor}
(\aliasFuzzyReadCurOne{} on the same statements) only in requiring no
token-boundary-aligned mention span, so careful mention detection is worth
most of that gap and the released number is a floor rather than a restatement
of our own result. Both rules ship as reference baselines. Second, the
entity-linker rung is handed the alias inventory the stream drew from --- the
generator's on the mechanical configurations, the harvested one on the natural
configurations of Appendix~\ref{app:natural} --- so it has no
out-of-vocabulary mass at all; its residual
\aliasEntityDefOne{} points are ambiguity within a known alias space, and the
gap a real linker pays is bounded below, not estimated, by that number. The best write-path rung without an
alias table is the fuzzy linker: it reaches
\aliasFuzzyCurOne{} for a deficit of \aliasFuzzyDefOne{} points, and that --- not
\aliasEntityDefOne{} --- is this paper's estimate of the residue. On the
realistic corpus it roughly doubles (\fullFuzzyDefOne{} points,
Table~\ref{tab:fullcorpus}), so the benchmark is not saturated where it asks to
be ranked. Third, the reader metric has a
ceiling below one: three test facts have a superseded value nested inside the
gold value (\emph{Democratic Party} inside \emph{Democratic Party of Korea}),
so answering exactly correctly scores as hedged; the attainable maximum is
\readerCeiling{}, which is where the oracle reader's \readerOracleCorrect{}
should be read against.

The headline track restricts the queried set to single-valued facts that
change (\nChanging{} of \nFactKeys{} keys) for crisp top-1 scoring; Appendix~\ref{sec:scale}
lifts that restriction to \scaleNTest{} test facts, \scaleNMulti{} of them
multi-valued, and the write-path keying ladder holds. A top-1 there counts as current
if its value is valid in the final year, so a fact with several simultaneously
valid values is satisfied by any one of them; the set recall@5 of
Appendix~\ref{sec:scale} charges what that hides --- a one-slot policy keeps
\scaleSetRecOracleZero{} of the valid values against \scaleSetRecRwZero{}
for recency-weighted store-all --- and we report the two metrics side by side rather than
claiming either alone. Validity is yearly and
statements are templated: the anchor-free track models aliasing with
mechanically derived short forms, which covers the common case in logs but not
nicknames, cross-lingual forms, or pronominal coreference resolvable only from
discourse. Conversational reference --- pronouns, nicknames, discourse
coreference --- is therefore outside what this instrument measures;
Appendix~\ref{app:natural} narrows the gap with real aliases for the
\natCovSubjectsPct\% of test subjects that have one, without closing it. TempEL \citep{zaporojets2022tempel} supplies real temporal mention
drift over Wikipedia snapshots and would make that axis naturalistic rather
than synthetic; we generate the axis instead because we need it
\emph{controlled} --- the point is to read a policy's cost as a function of
anchor availability, which requires setting that availability rather than
observing it --- and a naturalistic version of the same axis is the obvious next
instrument. The alias inventory we hand the entity linker is also cleaner than
one a deployed system would build. The systems track is the narrowest evidence in the paper and should be read
that way. Four mechanisms, one of them our reimplementation, on
\sysNFacts{} facts, and for Mem0 two
releases with, for \probedAdjVersion{}, a single backend --- the only
\mbox{(release, backend)} pair of the three we tried that passes both halves
of the probe. We report a
cross-backend stability check for \probedCurVersion{}
(\probedCurLo{}--\probedCurHi{} over \probedCurNBackends{} stores) and cannot
offer one for \probedAdjVersion{}; we do report an independent repeat of that
cell on \repeatAdjStore{}, which reproduced its event histogram, slot count and
current@1 exactly (\repeatAdjCurDelta{} points apart), so within one backend
the pipeline is reproducible at temperature zero even though its extraction is
model-driven. A second adjudicator model, glm-4.6, is
measured in Appendix~\ref{sec:systems} (Table~\ref{tab:adjudicator}); it
moves the numbers at the extraction stage before the adjudication, and the
comparison holds only under an extraction prompt that admits third-person
facts. Reader-side numbers use a single reader, unlike
\S\ref{sec:reader}, which uses two.

The attribution finding itself has a bound. That
\attrCurIdWrongPctZero\% of \probedCurVersion{}'s memories carry an id naming a
different entity than their text is measured over that run's own memories, and
a six-statement probe of the same release attached metadata correctly --- so
the effect does not appear at small scale and we offer no mechanism for it,
only the measurement. Nothing here is load-bearing for the version finding,
which rests on a dispatch path anyone can read, an event histogram and a
two-statement probe.

Because facts arrive as they become true
and gold is the final value, most-recent-equals-current holds by construction
in the clean condition; Appendix~\ref{sec:tshuffle} measures the
corrupted-clock and corrupted-arrival regimes and a late-stale-report dose,
but ambiguous updates --- a statement that may or may not supersede an
earlier one --- and corrections with their own semantics --- a retraction
that names a previous statement as wrong --- are out of scope. The reader track uses \readerNModels{} models with one prompt
format; prompt sensitivity is a known confound in reader studies, and our
claim is comparative across contexts under a fixed prompt, not absolute
accuracy. English only; nine relations; \nTest{} test facts is enough for
tight CIs but not for per-relation claims.

\subsection{Ethical considerations}

\textbf{Data provenance and personal information.} The benchmark's facts are
public biographical and organizational relations --- team membership, employer,
office held, head coach, chair, party, head of government, owner, school ---
about public figures and organizations, taken from Wikidata via TempLAMA
\citep{dhingra2022templama} and released under CC0. They name real people, and
that is intrinsic to the task rather than incidental: the failure we measure is
a retriever's inability to prefer the current value of a fact \emph{about a
named entity}, so pseudonymizing the subjects would remove the phenomenon under
study. We add no information beyond what Wikidata already publishes, collect
nothing from private sources, and include no contact details, identifiers,
images, or free text written by the individuals concerned. One category
deserves explicit mention: political party affiliation is among the nine
relations, and party membership is treated as a special category of personal
data under some regimes. These are the publicly recorded affiliations of
office-holders and candidates, already published under CC0 by Wikidata, and we
neither infer nor annotate any affiliation ourselves. The aliases on the
anchor-free track are mechanical shortenings (head token, leading token,
initialed form) of names already present in the source, or --- on the two
natural configurations --- public short forms taken from Wikidata's alias
field and English Wikipedia redirect titles; neither adds identifying
information beyond what those sources publish. The release is a frozen 2010--2020 window and is not updated, so
it does not track any individual going forward.

\textbf{Offensive content.} Statements are template-verbalized
subject--relation--object triples over nine factual relations; no free text is
model-generated in the released data and no user-authored text is included, so
the release carries little risk of offensive, harmful, or deceptive content.
The reader-track model outputs that ship in the call cache are short factual
answers to these same questions.

\textbf{Risks of the work.} The clearest risk is that a benchmark which makes
freshness measurable is over-read as certifying it. Our central result argues
against exactly that reading --- a single retrieval number can rank memory
policies backwards --- and we report where each fix fails for the same reason.
A second risk is specific to the remedy we price: retention policies
\emph{delete} superseded statements, and in deployments where the history is
itself the record (audit, compliance, provenance, contested claims about a
person), destructive overwrite is the wrong mechanism and a false merge is
unrecoverable. That is why we evaluate Zep-style invalidation alongside
overwrite and report its footprint cost (Appendix~\ref{sec:keying}): a system
that must not lose history has a measured alternative here. We see no dual-use
concern; the beneficiary of fresher agent memory is the user who would
otherwise receive a confidently outdated answer.

\textbf{Environmental impact.} Every number outside the systems track
reproduces on a single laptop CPU in under an hour, and no GPU is required for
any result in this paper. The LLM tracks issue \llmCacheEntries{} short
completions, \llmCacheLocal{} served by a locally hosted open-weights model
and \llmCacheHosted{} by hosted models, cached on
disk and shipped with the release so that reproduction requires no further
inference (Appendix~\ref{app:repro}).

\end{document}